%% file: main.tex
\documentclass[10pt,twocolumn,letterpaper]{article}

\usepackage[pagenumbers]{iccv} 
\usepackage[accsupp]{axessibility}

\newcommand{\real}{\mathbb{R}}

\newcommand{\grad}{\nabla}
\newcommand{\loss}{\mathcal{L}}

\usepackage{siunitx}

\usepackage{makecell}
\usepackage{amsmath}
\usepackage{caption}
\usepackage{varwidth}
\usepackage{wrapfig}
\usepackage{multirow}
\usepackage[T1]{fontenc}

\definecolor{iccvblue}{rgb}{0.21,0.49,0.74}
\usepackage[pagebackref,breaklinks,colorlinks,allcolors=iccvblue]{hyperref}

\def\paperID{12339} 
\def\confName{ICCV}
\def\confYear{2025}

\title{FLOAT: Generative Motion Latent Flow Matching for Audio-driven Talking Portrait}

\author{
Taekyung Ki$^{1\star}$ ~\quad~ Dongchan Min$^1$ ~\quad~ Gyeongsu Chae$^2$ \\
\normalsize{$^1$KAIST ~\quad~ $^2$DeepBrain AI Inc.} \\
{\tt\small \{taekyung.ki, alsehdcks95\}@kaist.ac.kr} ~~~ {\tt\small gc@deepbrain.io} \\
{\small \url{https://deepbrainai-research.github.io/float/}}
}

\begin{document}
\twocolumn[{
\maketitle
\begin{center}
    \vspace*{-1mm}
    \captionsetup{type=figure}
    \includegraphics[width=0.83\linewidth]{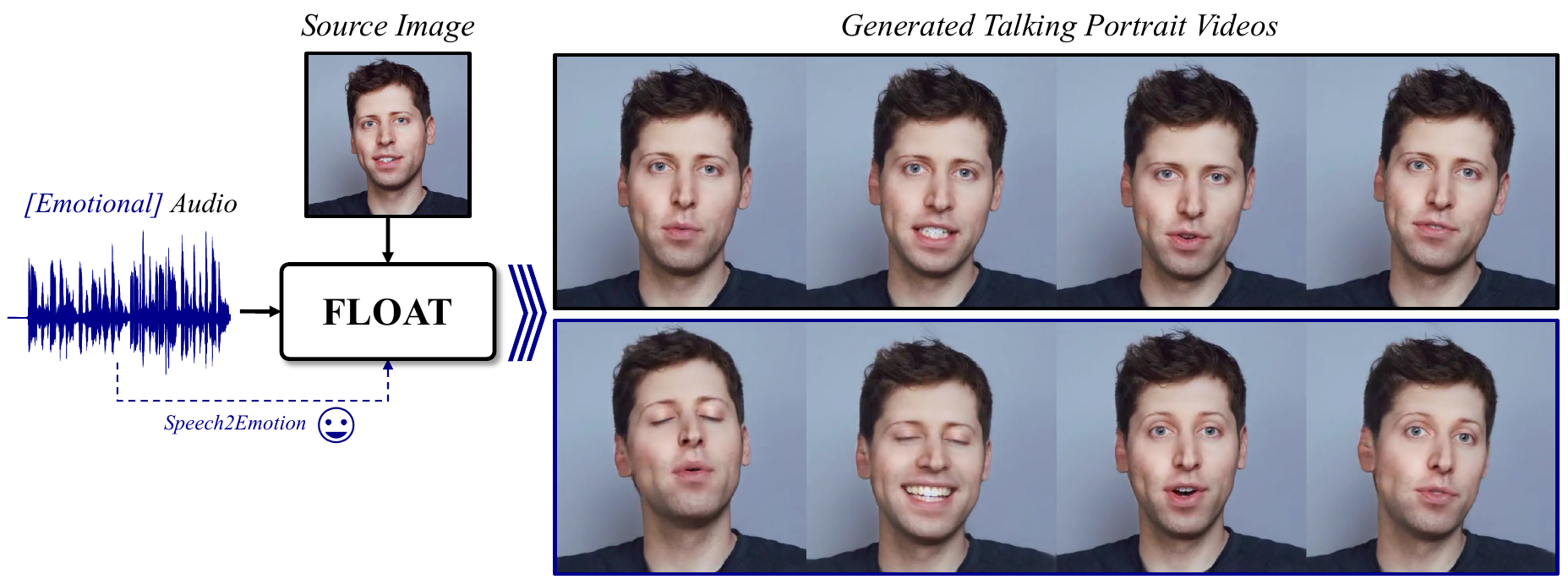}
    \vspace*{-2mm}
    \captionof{figure}{FLOAT can generate a talking portrait video from a single source image and audio where the talking motion is generated by the motion latent flow matching. It can enhance the emotion-related talking motion by leveraging speech-driven emotion labels, a natural way of emotion-aware motion control.}
\end{center}
}]

\def\thefootnote{$\star$}\footnotetext{This work was done during South Korea Mandatory Military Service at DeepBrain AI Inc.}\def\thefootnote{\arabic{footnote}}

\input{sec/0_abstract}    
\input{sec/1_intro}
\input{sec/2_related}
\input{sec/3_methods}
\input{sec/4_experiments}
\input{sec/5_conclusion}

{
    \small
    \bibliographystyle{ieeenat_fullname}
    \bibliography{main}
}
\input{sec/6_suppl.tex}

\clearpage

\begin{figure*}
    \centering
    \includegraphics[width=\linewidth]{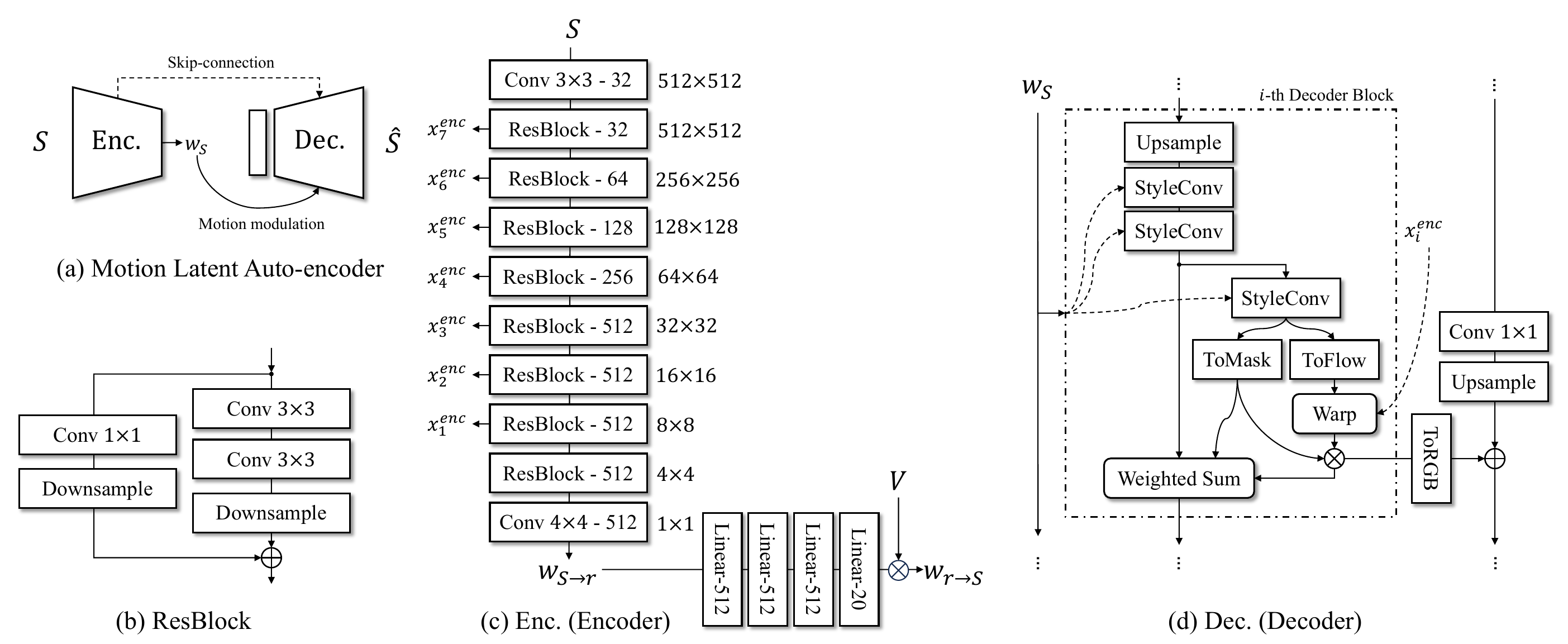}
    \caption{Detailed Model architecture of our motion latent auto-encoder. The notations are adopted from LIA \cite{lia} and StyleGAN2 \cite{stylegan2}.}
    \label{fig:phase1_architecture}
\end{figure*}

\begin{figure*}
    \centering
    \includegraphics[width=0.8\linewidth]{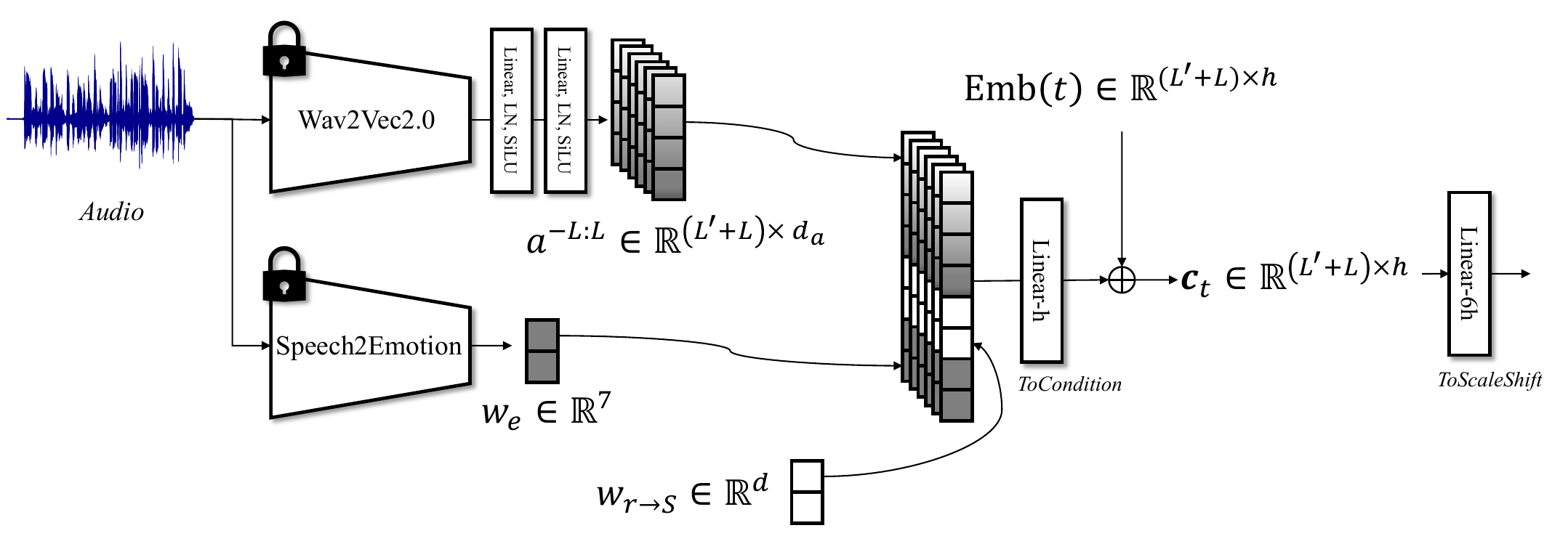}
    \caption{Detailed model architecture for constructing the driving conditions $\mathbf{c}_t \in \real^{(L' + L) \times h}$ in FLOAT.}
    \label{fig:phase2_architecture}
\end{figure*}

\clearpage

\begin{figure*}
    \centering
    \includegraphics[width=0.95\linewidth]{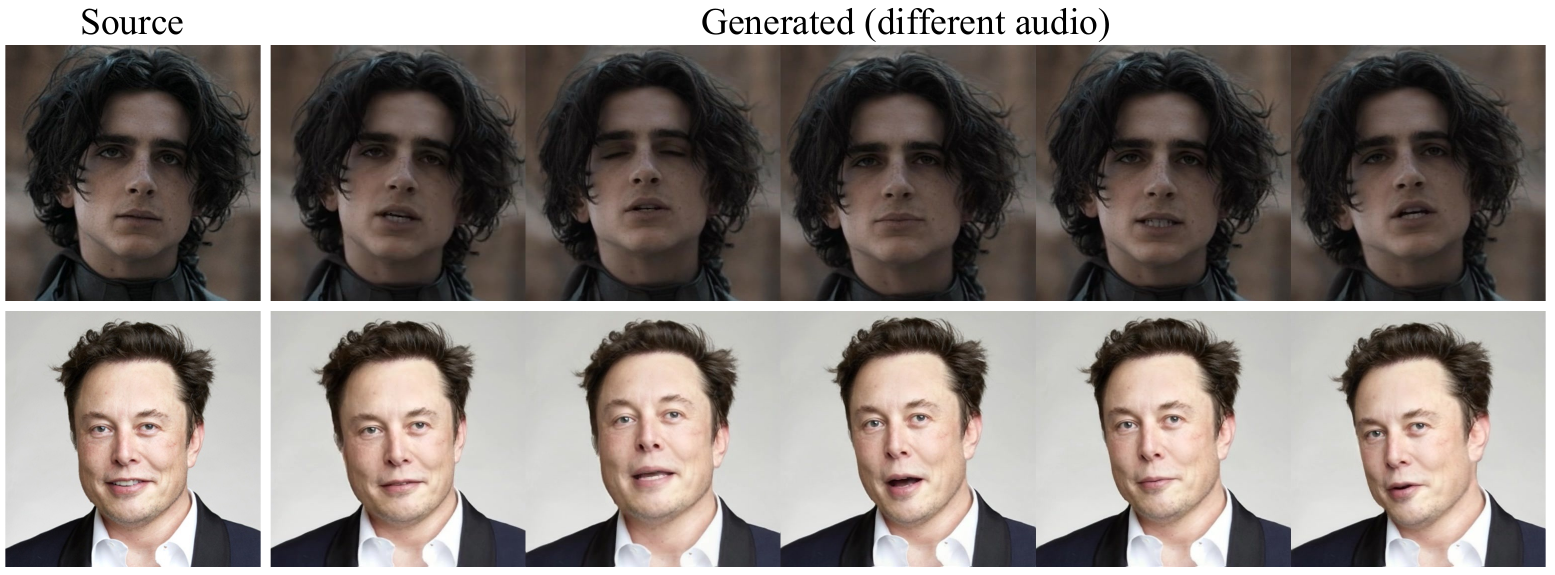}
    \vspace*{-2mm}
    \caption{Out-of-distribution results. The first row shows the result for \textit{Chinese} audio, and the second row shows the result for \textit{singing} audio. Please refer to supplementary video.}
    \label{fig:ood_supp}
\end{figure*}

\begin{figure*}
    \centering
    \includegraphics[width=0.95\linewidth]{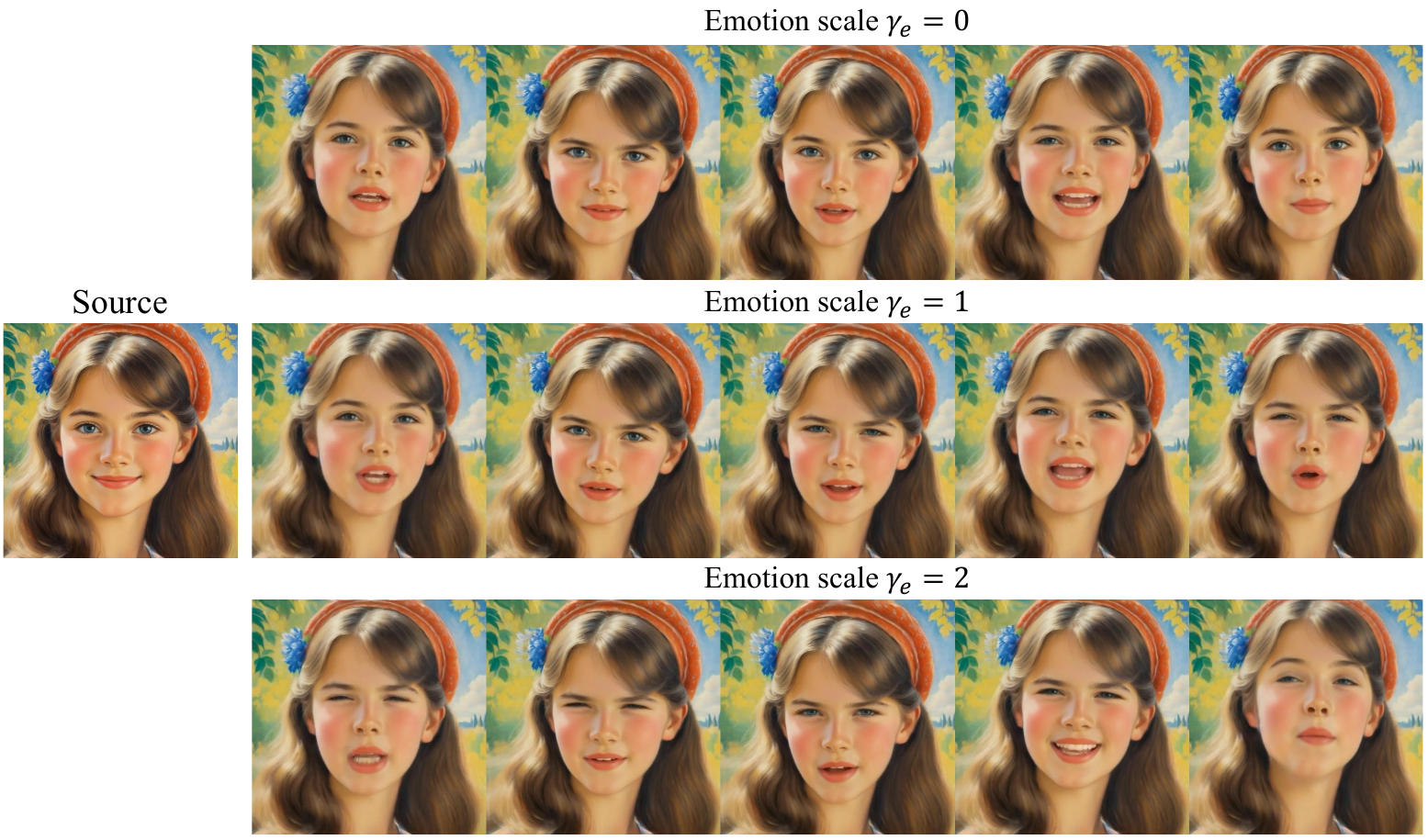}
    \vspace*{-2mm}
    \caption{Ablation on emotion guidance scale $\gamma_e$. The predicted speech-to-emotion label is \textit{disgust} of $99.99\%$. Please refer to supplementary video.}
    \label{fig:ecfg_scale_supp}
\end{figure*}

\clearpage
\clearpage

\begin{figure*}
    \centering
    \includegraphics[width=0.85\linewidth]{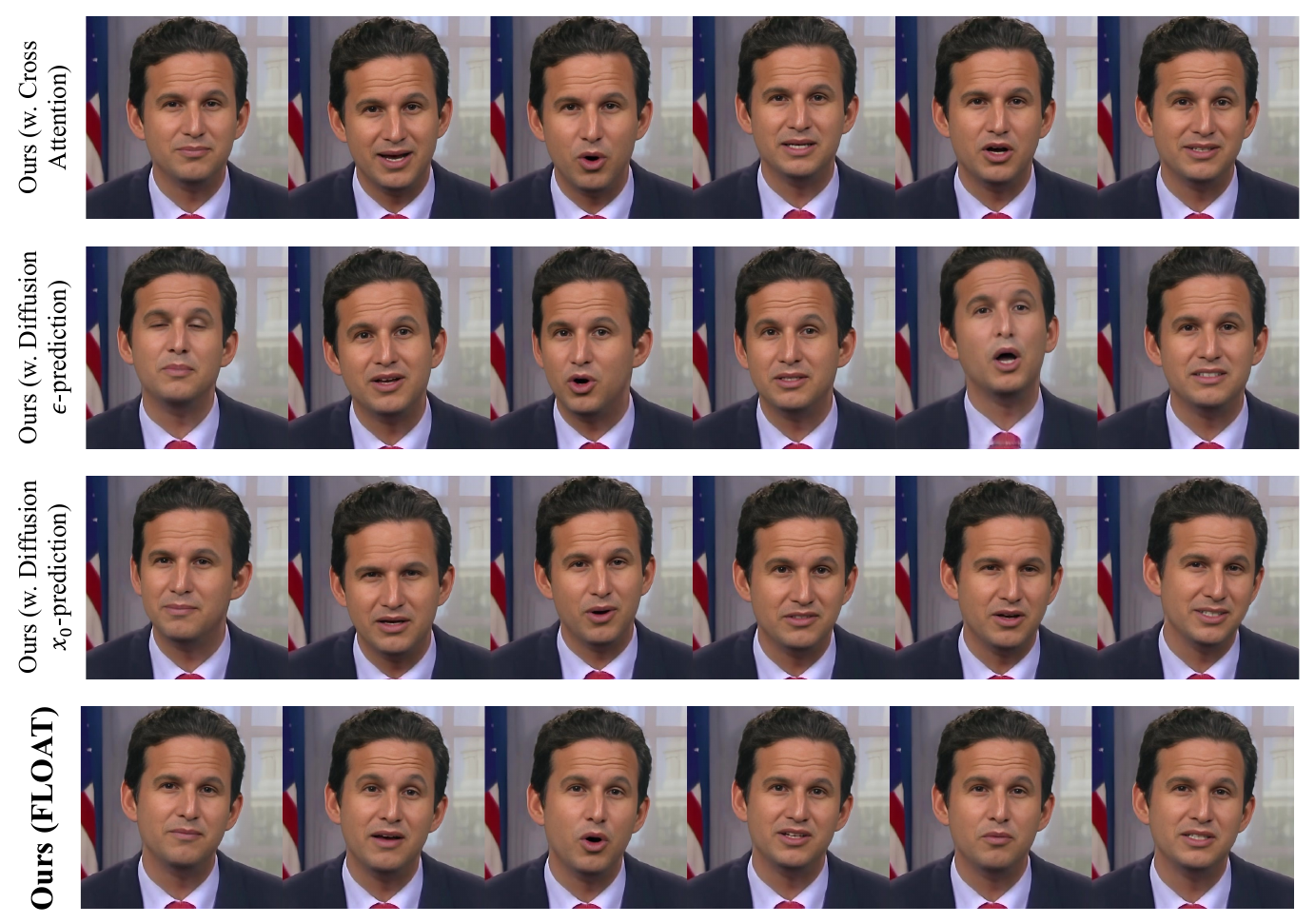}
    \vspace*{-2mm}
    \caption{Ablation results on frame-wise AdaLN and flow matching. Please refer to supplementary video.}
    \label{fig:ablation_diffusion_supp1}
\end{figure*}

\begin{figure*}
    \centering
    \includegraphics[width=0.85\linewidth]{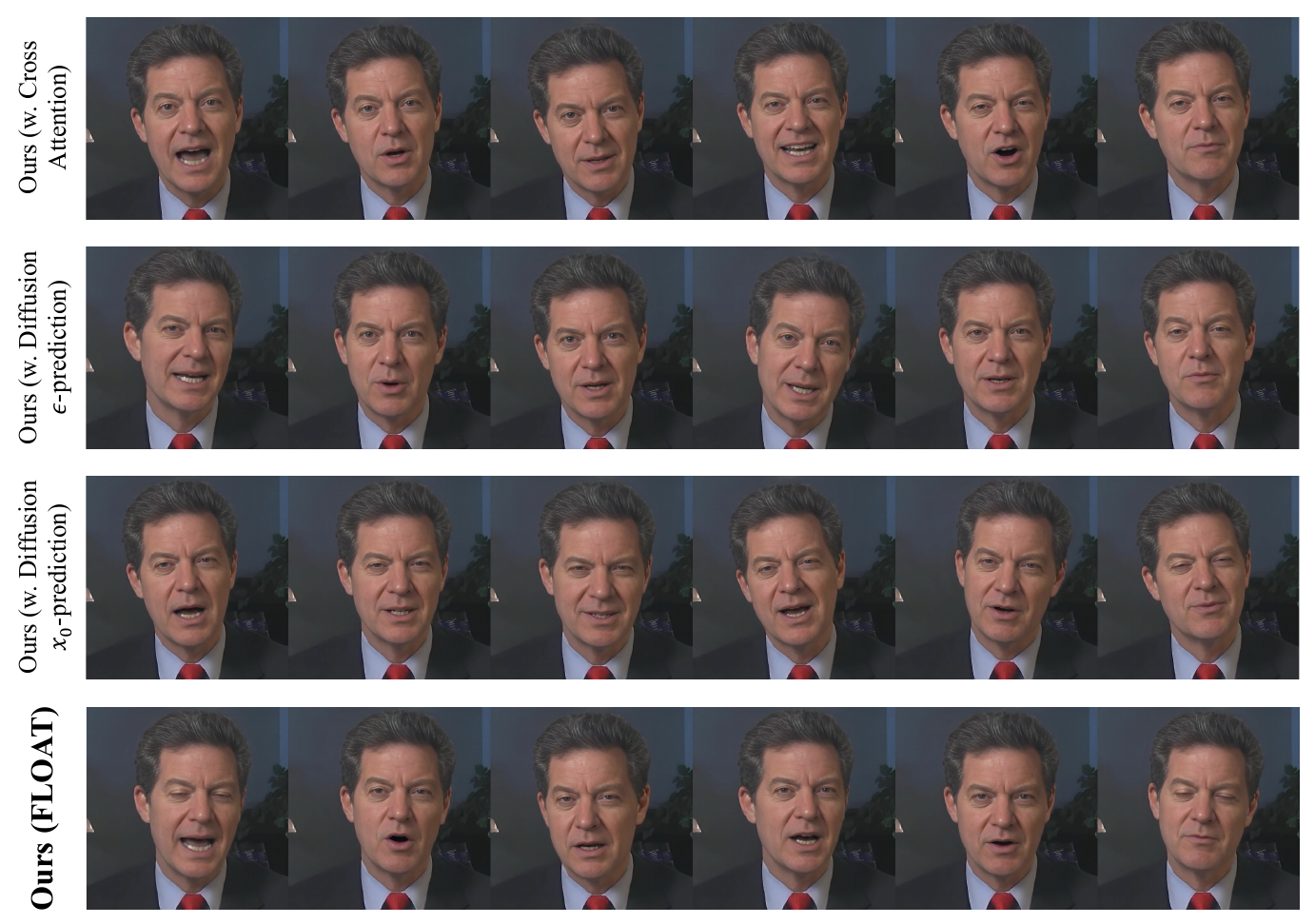}
    \vspace*{-2mm}
    \caption{Ablation results on frame-wise AdaLN and flow matching. Please refer to supplementary video.}
    \label{fig:ablation_diffusion_supp3}
\end{figure*}

\begin{figure*}
    \centering
    \includegraphics[width=0.85\linewidth]{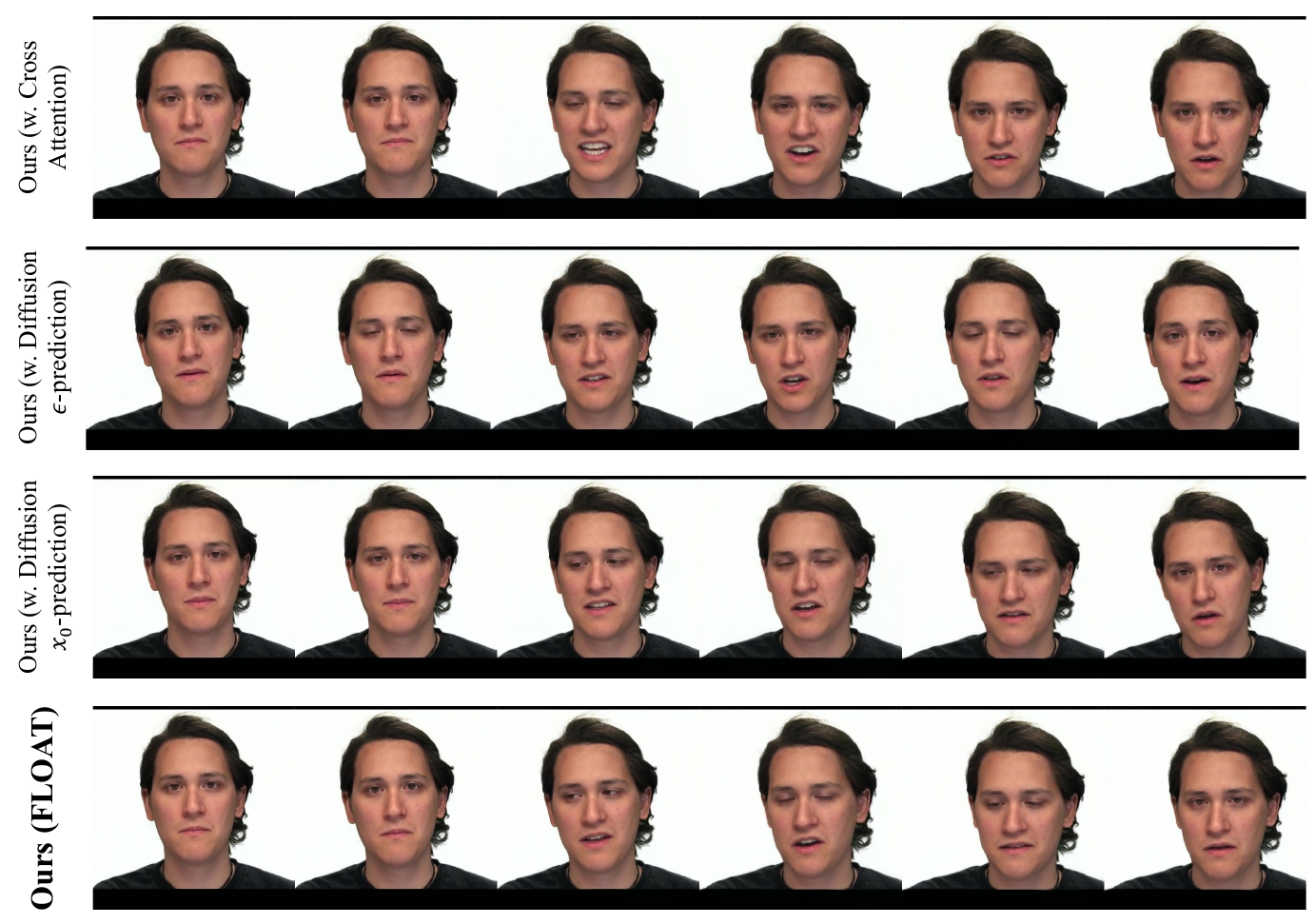}
    \vspace*{-2mm}
    \caption{Ablation results on frame-wise AdaLN and flow matching. Please refer to supplementary video.}
    \label{fig:ablation_diffusion_supp4}
\end{figure*}

\clearpage

\begin{figure*}
    \centering
    \includegraphics[width=0.95\linewidth]{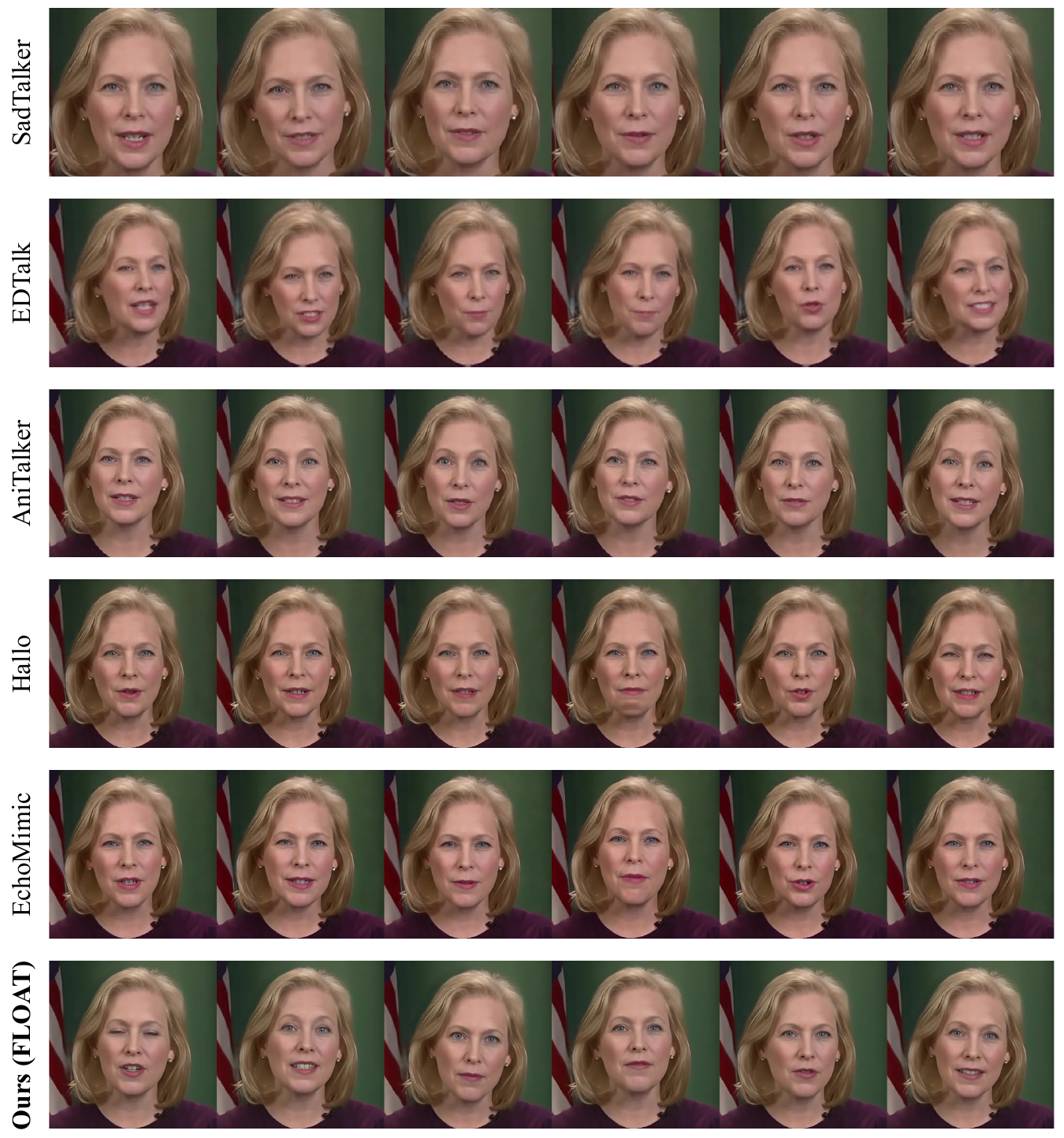}
    \vspace*{-3mm}
    \caption{Qualitative comparison results with state-of-the-art methods. Please refer to supplementary video.}
    \label{fig:sota_compare_supp1}
\end{figure*}
\clearpage

\begin{figure*}
    \centering
    \includegraphics[width=0.95\linewidth]{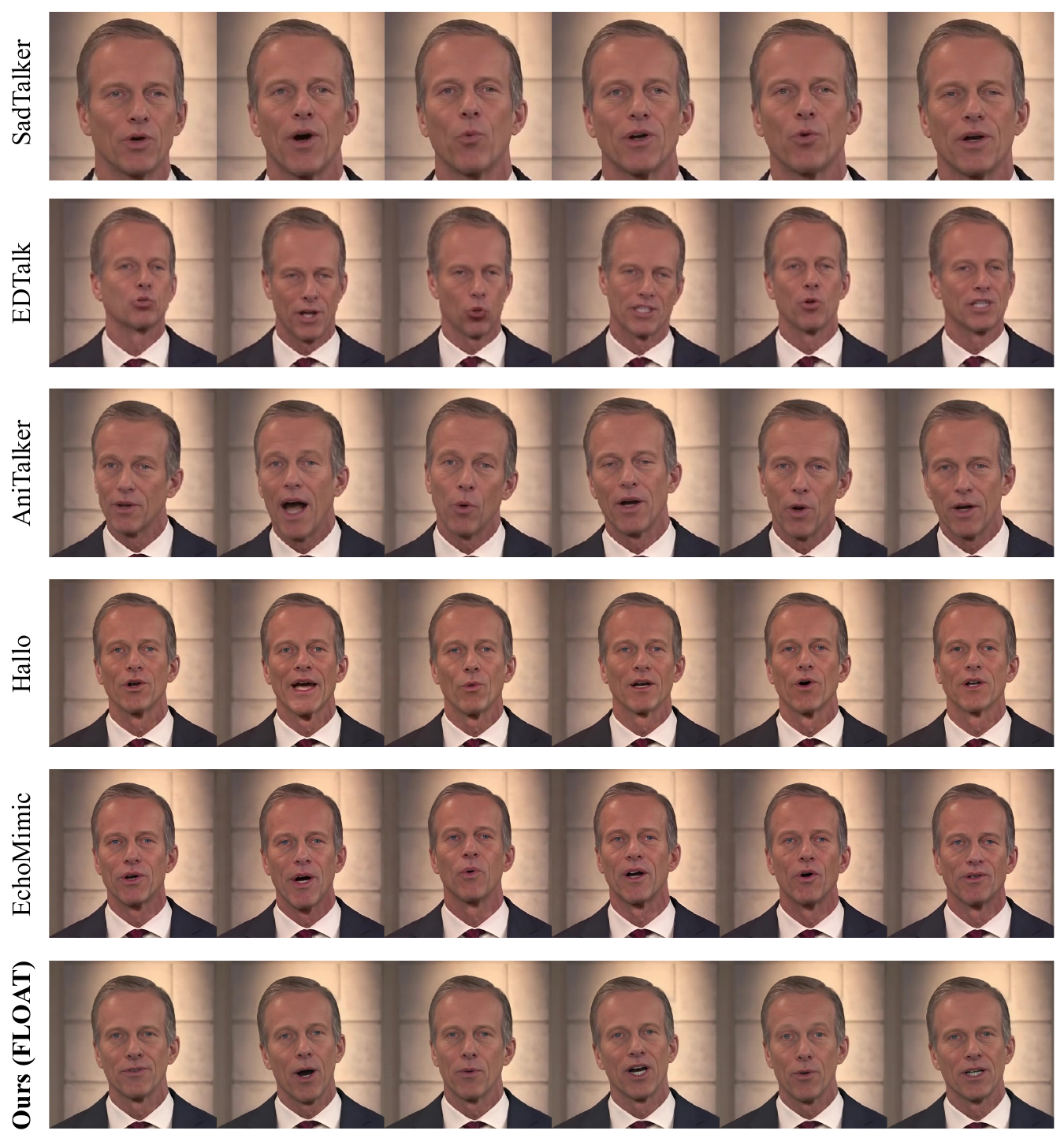}
    \vspace*{-3mm}
    \caption{Qualitative comparison results with state-of-the-art methods. Please refer to supplementary video.}
    \label{fig:sota_compare_supp2}
\end{figure*}
\clearpage

\begin{figure*}
    \centering
    \includegraphics[width=0.95\linewidth]{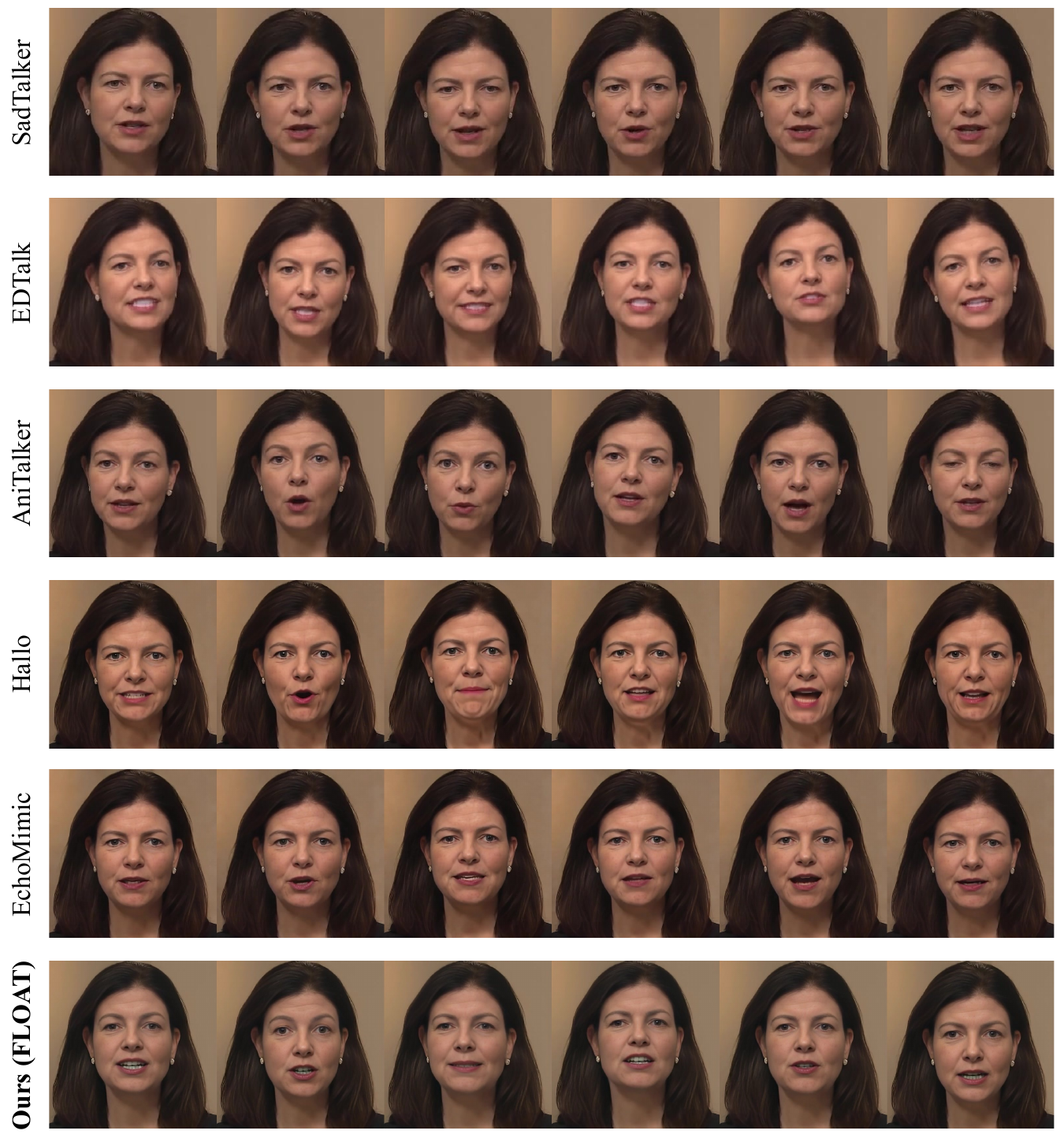}
    \vspace*{-3mm}
    \caption{Qualitative comparison results with state-of-the-art methods. Please refer to supplementary video.}
    \label{fig:sota_compare_supp3}
\end{figure*}
\clearpage

\end{document}

%% file: sec/0_abstract.tex
\begin{abstract}
With the rapid advancement of diffusion-based generative models, portrait image animation has achieved remarkable results. However, it still faces challenges in temporally consistent video generation and fast sampling due to its iterative sampling nature. This paper presents FLOAT, an audio-driven talking portrait video generation method based on flow matching generative model. Instead of a pixel-based latent space, we take advantage of a learned orthogonal motion latent space, enabling efficient generation and editing of temporally consistent motion. To achieve this, we introduce a transformer-based vector field predictor with an effective frame-wise conditioning mechanism. Additionally, our method supports speech-driven emotion enhancement, enabling a natural incorporation of expressive motions. Extensive experiments demonstrate that our method outperforms state-of-the-art audio-driven talking portrait methods in terms of visual quality, motion fidelity, and efficiency.
\end{abstract}

%% file: sec/1_intro.tex
\section{Introduction}
\label{sec:intro}
Animating a single image using a driving audio (\ie, audio-driven talking portrait generation) has gained significant attention in recent years for its great potential in avatar creation, video conferencing, virtual avatar chat, and user-friendly customer service. It aims to synthesize natural talking motion from audio signals, including accurate lip synchronization, rhythmical head movements, and fine-grained facial expressions. However, generating such motion solely from audio is extremely challenging due to its one-to-many correlation between audio and motion. In the earlier stage of this field, many works \cite{wav2lip, stylelipsync, stylesync, synctalkface, videoretalking, dinet} focus on generating accurate lip movements by relying on learned audio-lip alignment losses \cite{syncnet, styletalker}.

To comprehensively extend the range of motion, some works \cite{sadtalker, styletalker, edtalk} incorporate probabilistic generative models, such as VAE \cite{vae} and normalizing flow \cite{normalizing_flow}, turning the motion generation into probabilistic sampling. However, these models still lack expressiveness in generated motion due to the limited capacity of these generative models.

Recent talking portrait generation methods \cite{emo, aniportrait, echomimic, hallo, loopy, anitalker, dreamtalk, v_express, gaia, diffusedhead}, powered by diffusion-based generative models \cite{ddpm, sde}, successfully mitigate this expressiveness issue. EMO \cite{emo} introduces a promising approach to this field \cite{aniportrait, v_express, echomimic, hallo, loopy} by employing a strong pre-trained image diffusion model (\ie, StableDiffusion \cite{ldm}) and lifting it into video generation \cite{animate_anyone}. However, it still faces challenges in generating temporally coherent videos and achieving sampling efficiency, requiring tens of minutes for a few seconds of video. Moreover, they heavily rely on auxiliary facial prior, such as bounding boxes \cite{emo, hallo}, 2D landmarks and skeletons \cite{controlnet, echomimic, loopy}, or 3D meshes \cite{aniportrait}, which significantly restricts the diversity and the fidelity of head movements due to their strong spatial bias. 

In this paper, we present FLOAT, an audio-driven talking portrait video generation model based on flow matching generative model in a motion latent space. Flow matching \cite{cfm, rectflow} has emerged as a promising alternative to diffusion models due to its fast and high-quality sampling. By modeling talking motion within a learned motion latent space \cite{lia}, we can more efficiently sample temporally consistent motion latents. This is achieved by a simple yet effective transformer-based \cite{transformer} vector field predictor, inspired by DiT \cite{dit}. Since our motion latent space has orthogonal structure, our method can manipulate head motion of the generated video using its basis. Furthermore, our method supports natural emotion-aware motion enhancement driven by speech.
Our contributions are summarized as follows:
\begin{itemize}
    \item We present, \textbf{FLOAT}, \textbf{flo}w matching based \textbf{a}udio-driven \textbf{t}alking portrait generation model using a learned orthogonal motion latent space, enabling to generate talking portrait videos with reduced sampling steps.
    \item We introduce a simple yet effective transformer-based flow vector field predictor for temporally consistent motion latent sampling, which also enables the speech-driven emotional controls.
    \item Extensive experiments demonstrate that FLOAT achieves state-of-the-art performance compared to both diffusion- and non-diffusion-based methods. 
\end{itemize}

%% file: sec/2_related.tex
\section{Related Works}
\subsection{Diffusion Models and Flow Matching} 
\noindent \textbf{Diffusion Models~}~ Diffusion models or score-based generative models \cite{ldm, ddpm, ddim, iddpm, sde, beats} are generative models that gradually diffuse input signals into Gaussian noise and learn the denoising reverse process for the generative modeling. They have shown remarkable results in various generation tasks, such as unconditional image and video generation \cite{dit, sora, sd3}, text-to-image generation \cite{ldm, dalle2, imagen}, text-to-video generation \cite{sora, animatediff}, conditional image generation \cite{controlnet, animate_anyone}, and 3D human generation \cite{mdm, diffposetalk, diffusionavatars}.

\noindent \textbf{Accelerating Diffusion Models~}~ While diffusion models demonstrate superior performance, their iterative sampling nature still bottlenecks the efficient generation compared to VAEs \cite{vae}, normalizing flow \cite{normalizing_flow}, and GANs \cite{gan}. To overcome this limitation, several works have been developed to boost the sampling speed of the diffusion models. StableDiffusion (SD) \cite{ldm} partially mitigates this problem by moving the diffusion process from the pixel space to the spatial latent space, establishing itself as a pivotal framework among diffusion models. Another line of research has developed the sampling solvers \cite{dpm_solver, dpm_solver_pp} based on ordinary differential equations (ODEs). Meanwhile, model distillation \cite{distillation} has been introduced to transfer the knowledge of the learned diffusion models into a student model, enabling one (or a few) steps of generation \cite{consistency_model, latent_consistency_model, instaflow, diffusion2gan, fasterdiffusion}. However, these approaches involve substantial effort to create a well-trained diffusion model and suffer from training instability.

\noindent \textbf{Flow Matching~}~ Flow matching \cite{cfm, rectflow} stands out as an alternative to diffusion models for its high sampling speed and competitive sample quality compared to diffusion models \cite{cfm, lcfm, voicebox, boosting_flow_matching, moviegen}. It belongs to the family of flow-based generative models, which estimates a transformation (referred to as a \textit{flow}) between a prior distribution (\eg, Gaussian) and a target distribution. Unlike the normalizing flow \cite{normalizing_flow, realnvp} that directly estimates the noise-to-data transformation under specific architectural constraints (\eg, affine coupling), flow matching regresses the time-dependent vector field that \textit{generates} this flow by solving its corresponding ODEs \cite{node} with flexible architectures. One specific design of flow matching is an optimal transport (OT) based one, which transforms the data distribution along the straight path with constant velocity \cite{cfm}.

Our audio-driven talking portrait method employs flow matching to generate the natural talking motions. Thanks to the architectural flexibility of flow matching, we use transformer-encoder architecture \cite{transformer} to estimate the generating vector field, allowing us to take the video temporal consistency into account. 

\subsection{Audio-driven Portrait Animation}
Audio-driven portrait animation is the task of generating a realistic talking portrait video using a single portrait image and driving audio \cite{audio2head, makeittalk, pcavs, sadtalker, styletalker}. Since audio-to-motion relation is basically a one-to-many problem, several works utilize additional facial prior for driving conditions, \eg, 2D facial landmarks \cite{makeittalk, v_express, echomimic, loopy, gaia, aniportrait}, 3D prior \cite{styletalk, styleheat, sadtalker, videoretalking, dreamtalk}, or emotional labels \cite{vasa_1, eamm, emmn}.
In earlier stages, most works \cite{wav2lip, stylelipsync, stylesync, videoretalking} focused on generating accurate lip motion from audio by utilizing the lip-sync discriminator \cite{syncnet}. These approaches have advanced to generating audio-related head poses in a probabilistic way. For example, StyleTalker \cite{styletalker} uses normalizing flow \cite{normalizing_flow, realnvp} to generate the head motion from audio, while SadTalker \cite{sadtalker} uses audio-conditional variational inference \cite{vae} to learn the 3DMM coefficients \cite{3dmm}, bridging the intermediate representations of a pre-trained portrait animator \cite{osfv}. 

Meanwhile, several works \cite{mead, eamm, emmn, gmtalker} focus on an emotion-aware talking portrait generation. In particular, EAMM \cite{eamm} considers an emotion as the complementary displacement of facial motion, and learns these displacement from an emotion label extracted from the image.

Recent audio-driven talking portrait methods powered by diffusion models show remarkable results \cite{dreamtalk, emo, hallo, vasa_1,loopy, echomimic, v_express, aniportrait, anitalker}. Specifically, EMO \cite{emo} and subsequent extensions \cite{hallo, echomimic, v_express, aniportrait} utilize the pre-trained SD \cite{ldm} as their backbone to leverage generative prior trained on the large-scale image datasets. They introduce additional modules, \eg, ReferenceNet \cite{animate_anyone} and Temporal Transformer \cite{animatediff}, to preserve input identity and enhance the video temporal consistency, respectively. However, these modules introduces additional computational cost, requiring several minutes for a few seconds of video, and still suffer from video-level artifacts, such as noisy frames, and flickering.

VASA-1 \cite{vasa_1} addresses the sampling time issue by sampling motion latents \cite{megaportrait}, producing lifelike talking portraits. Our method takes advantage of this approach. However, unlike \cite{vasa_1}, our motion latent space has a strong linear orthogonal structure represented by a computable basis, enabling to manipulate the generated motion at the test-time without external driving signals. Based on this orthogonality, we employ OT-based flow matching for motion latent sampling along a straight line with reduced sampling steps.

%% file: sec/3_methods.tex
\section{Preliminaries: (Conditional) Flow Matching} \label{sec:cfm}
\begin{figure*}[t]
    \centering
    \includegraphics[width=0.98\linewidth]{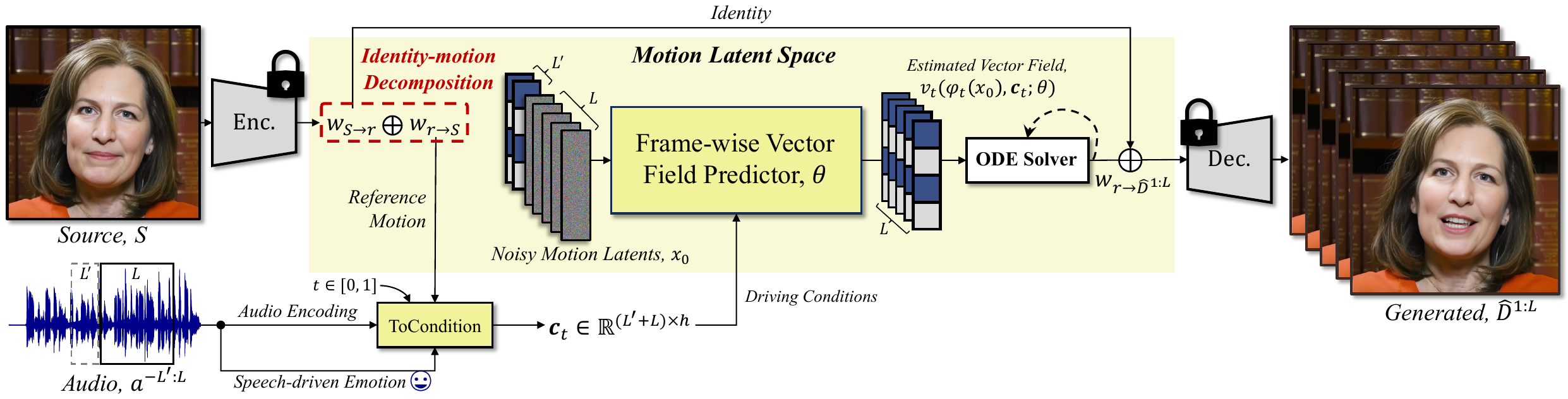}
    \vspace*{-1mm}
    \caption{Overview of FLOAT. We encode the source image $S\in\real^{3\times H \times W}$ into the latent with the explicit identity-motion decomposition $w_s = w_{s\to r} + w_{r \to s} \in \real^d$. Given audio segments $a^{-L':L} \in \real^{(L'+L) \times d_a}$ of the length $L' + L$ and the reference motion $w_{r\to s}$ $\in$ $\real^{d}$, and the speech-driven emotion label $w_e$ $\in$ $\real^7$, a flow matching transformer estimates the generating vector field $v_t(\varphi_t(x_0), \mathbf{c}_t; \theta) \in \real^{L \times d}$ from noisy motion latents, which is used to solve corresponding ODE and generates the motion latents $w_{r\to\hat{D}^{1:L}}$. Finally, the sequence of latents $w_{S \to \hat{D}^{1:L}} := (w_{S \to r} + w_{r \to \hat{D}^{l}})_{l=1}^{L}$ are decoded into the video $\hat{D}^{1:L} \in \real^{L \times 3 \times H \times W}$.}
    \label{fig:overview}
    \vspace*{-3mm}
\end{figure*}

Let $x$ $\in$ $\real^d$ be a data, $t$ $\in$ $[0, 1]$ be the time, and $q$ be a unknown target distribution. We can define a \textit{flow} as a time-dependent transformation $\varphi_t: [0, 1] \times \real^d \to \real^d$ that transforms a tractable prior distribution $p_0$ to the distribution $p_1 \approx q$. This flow $\varphi_t$ further introduces a \textit{probability flow path} $p_t : [0, 1] \times \real^d \to \real_{>0}$ and a \textit{generating vector field} $v_t: [0, 1] \times \real^d \to \real^d$ where $p_t$ is defined by the push-forwarding \begin{align}
    p_t (x) = p_0 (\varphi_t^{-1}(x)) \det\left|\frac{\partial\varphi_t^{-1}(x)}{\partial x} \right|,
\end{align}
and $v_t$ generates $\varphi_t$ by means of an ordinary differential equation (ODE) \cite{node}:
\begin{align}
    \frac{d}{dt}\varphi_t (x) = v_t (\varphi_t(x)) \quad \text{and} \quad \varphi_0(x) = x. \label{eq:ode}
\end{align}
Flow matching \cite{cfm} aims to estimate the target generating vector field $u_t$ with a neural network parameterized by $\theta$:
\begin{align}
    \loss_{\text{FM}}(\theta) := \| v_t(x;\theta) - u_t(x) \|_2^2 , \label{eq:fm}
\end{align}
 where $t$ $\sim$ $\mathcal{U}[0, 1]$ and $x$ $\sim$ $p_t(x)$.
However, the target generating vector field $u_t$ and the sample distribution $p_t$ are intractable. To address this issue, \cite{cfm} proposes a method for constructing a ``conditional" probability path $p_t(\cdot | x_1)$ as well as target ``conditional" vector field $u_t (\cdot | x_1)$ using a sample $x_1$ $\sim$ $q$ as a condition. And they prove that the following objective
\begin{align}
    \loss_{\text{CFM}}(\theta) := \|v_t(x;\theta) - u_t(x | x_1)\|_2^2, \label{eq:cfm}
\end{align}
where $t$ $\sim$ $\mathcal{U}[0, 1]$ and $x$ $\sim$ $p_t(x | x_1)$, is equivalent to \eqref{eq:fm} with respect to the gradient $\grad_{\theta}$.

One natural way of constructing $u_t(\cdot | x_1)$ is a ``straight line" that connects $x_0$ $\sim$ $p_0$ and $x_1$ $\sim$ $q$, drawing an \textit{optimal transport (OT)} path with constant velocity \cite{cfm}. Specifically, a linear time interpolation between $x_0$ and $x_1$ gives us the flow $x_t = \varphi_t(x) = (1 - t) x_0 + t x_1$, the conditional probability path $p_t (x | x_1)$ defined via the affine transformation $p_t (x | x_1) = \mathcal{N}(x | t x_1, (1-t)^2 I)$, and the target generating vector field
$u_t (x | x_1) = x_1 - x_0$. 
This specific choice turns the objective \eqref{eq:cfm} into
\begin{align}
        \loss_{\text{OT}}(\theta) := \|v_t( (1 - t)x_0 + tx_1 ;\theta) - ( x_1 - x_0) \|_2^2, \label{eq:cfm_final}
\end{align}
where $t$ $\sim$ $\mathcal{U}[0, 1]$, $x_0$ $\sim$ $p_0$, and $x_1$ $\sim$ $q$, all of which are tractable.

\noindent \textbf{Classifier-free Vector Field~~}~\cite{lcfm} formulates a classifier-free vector field (CFV) technique for flow matching, which enables class-conditional sampling more controllable manner without any extra classifier trained on noisy trajectory. Formally, CFV compute the modified vector field $\tilde{v}_t$ by  
\begin{align}
        \tilde{v}_t(x_t, c; \theta) \approx \gamma v_t(x_t, c; \theta) + (1 - \gamma) v_t(x_t, c=\emptyset; \theta), \label{eq:cfv}
\end{align}
where $\gamma$ denotes the guidance scale. $v_t(x_t, c = \emptyset; \theta)$ is the predicted vector field without a driving condition $c$. For more details, please refer to \cite{cfm, lcfm}.

\section{Method: Flow Matching for Audio-driven Talking Portrait} \label{sec:methods}
We provide an overview of FLOAT in \cref{fig:overview}. Given source image $S$ $\in$ $\real^{3\times H \times W}$, and a driving audio signal $a^{1:L}$ $\in$ $\real^{L \times d_a}$ of length $L$, our method generates a video
\begin{align}
    \hat{D}^{1:L} = (\hat{D}^l)_{l=1}^{L} \in \real^{L \times 3 \times H \times W}
\end{align}
of $L$ frames, featuring audio-synchronized talking head motions, including both verbal and non-verbal motions. Our method consists of two phases. First, we pre-train a motion auto-encoder, which provides us with the expressive and smooth motion latent space for the talking portraits (\cref{sec:phase1}).
Next, we employ OT-based flow matching \cite{cfm} to generate a sequence of motion latents with a transformer-based vector field predictor using the driving audio, which is decoded to the talking portrait videos (\cref{sec:phase2}). We also incorporate speech-driven emotions as the driving conditions, achieving automatic emotion-aware talking portrait generation without any extra user input for emotion.

\subsection{Motion Latent Auto-encoder} \label{sec:phase1}
Recent talking portrait methods utilize the VAE of StableDiffusion (SD) \cite{ldm} due to its rich semantic pixel-based latent space. However, they often struggle to generate temporally consistent frames when lifted to video generating tasks \cite{emo, hallo, echomimic, animate_anyone, champ}. Thus, our first goal for realistic talking portrait is to obtain \textit{good} motion latent space, capturing both global (\eg, head motion) and fine-grained local (\eg, facial expressions, mouth and pupil movement) dynamics.

Instead of VAE of SD, we employ LIA \cite{lia} as a base motion latent auto-encoder and pre-train it to encode images into motion latents. This is achieved by training the auto-encoder to reconstruct a driving image from a source image sampled from the same video clip, enforcing the encoder to implicitly capture both temporally adjacent and distant motions. Following \cite{lia}, we use a learned orthonormal basis that can decompose the motion along distinct orthogonal directions. Specifically, our motion auto-encoder encodes the source $S$ into the latent $w_{S} \in \real^{d}$ with following explicit decomposition: 
\begin{equation}
    w_{S} := w_{S \to r} + w_{r\to S},
\end{equation} 
where $w_{S \to r}$ $\in$ $\real^{d}$ is the identity latent and 
\begin{equation}
    w_{r\to S} = \sum_{m=1}^{M} \lambda_{m}(S) \cdot \mathbf{v}_m \in \real^d \label{eq:motion}
\end{equation}
is the motion latent with $\lambda (S)$ $:=$ $\left( \lambda_m (S)\right)_{m=1}^{M}$  $\in$ $\real^{M}$ being the source-dependent motion coefficients that span the learned source-agnostic motion basis $V := \{\mathbf{v}_m\}_{m=1}^{M} \subseteq \real^{d}$. In this space, $\lambda_m(S)$ is the intensity of the motion direction $\mathbf{v}_m$. As shown in \cref{fig:lambda_control}, our method enables motion editing of the sampled (generated) motion using only the basis $V$ and its orthogonality, as stated in \cref{eq:lambda}.

\begin{figure}[t]
    \centering
    \includegraphics[width=0.47\textwidth]{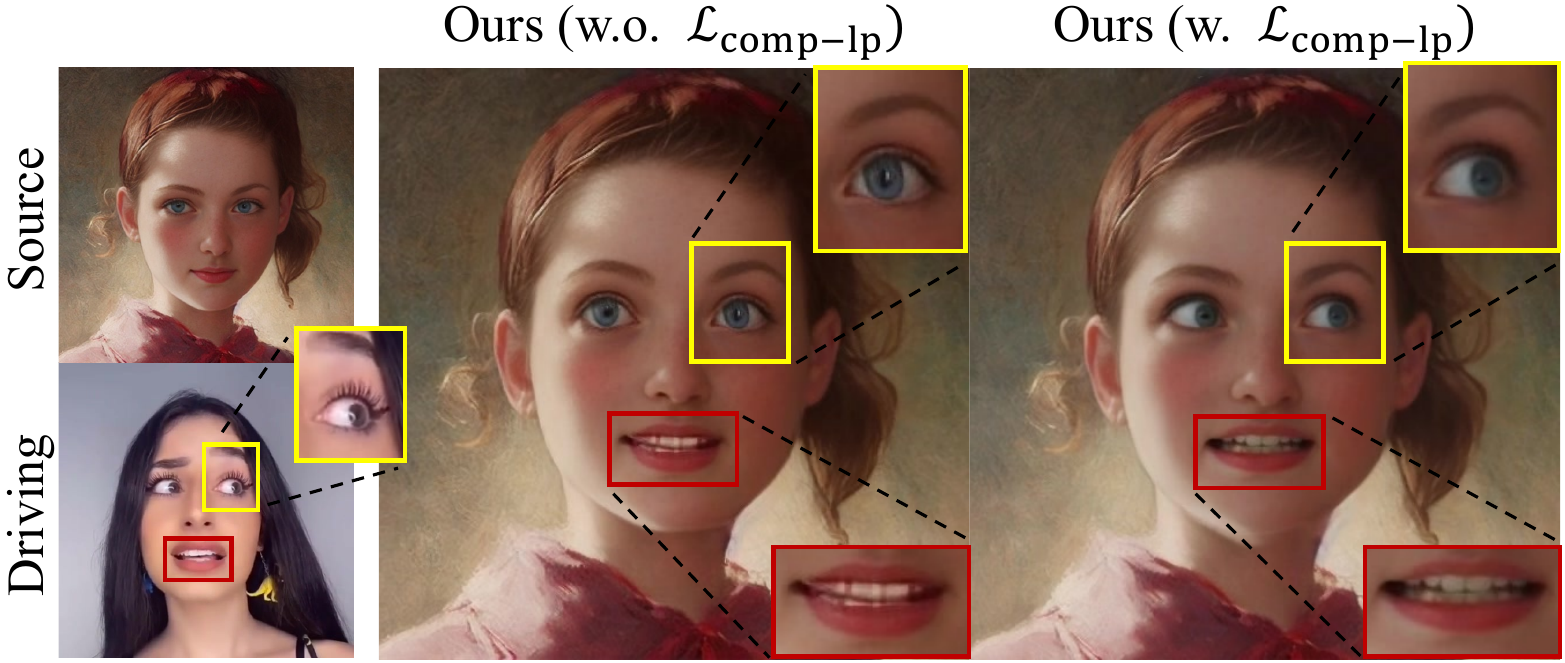}
    \vspace*{-2mm}
    \caption{Efficacy of $\loss_{\text{comp-lp}}$ for fine-grained motion and fidelity.}
    \vspace*{-2mm}
    \label{fig:wo_comp_lp_1}
\end{figure}

\noindent \textbf{Improving Fidelity of Facial Components:} {$\loss_{\text{comp-lp}}$~~} ~ The expressiveness of generated motions and the image fidelity are determined by the motion space and the motion auto-encoder. However, as resolution increases, fine details in small facial regions (\eg, teeth, eyeballs) often get buried in large-scale dynamics. To address this issue, we propose a \textit{facial component perceptual loss} $\loss_{\text{comp-lp}}$ using \cite{lpips, vgg} that significantly improves the image fidelity (\eg, teeth and eyes) as well as fine-grained motions (\eg, eyeball and eyebrows movements). As shown in \cref{fig:wo_comp_lp_1}, $\loss_{\text{comp-lp}}$ allows us to generate high-fidelity facial components and their fine-grained motions without relying on pre-trained foundation models, such as StableDiffusion \cite{ldm}.

\subsection{Flow Matching in Motion Latent Space} \label{sec:phase2}
\begin{figure}[t]
    \centering
    \includegraphics[width=0.44\textwidth]{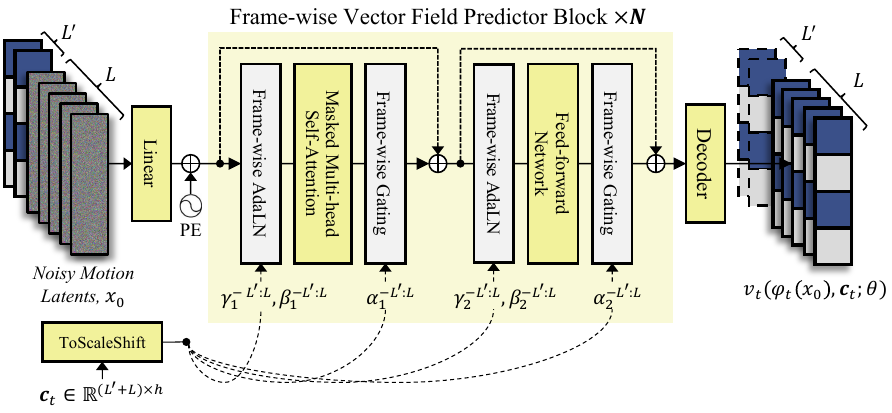}
    \vspace*{-3mm}
    \caption{Frame-wise vector field predictor block at inference.}
    \vspace*{-3mm}
    \label{fig:cfmt}
\end{figure}

Armed with this linear orthogonal space, we employ OT-based flow matching \cite{cfm, rectflow} for the motion sampling. Specifically, we predict a vector field $v_t(x_t, \mathbf{c}_t;\theta)$ $\in$ $\real^{L \times d}$ where $x_t$ is the sample at flow time $t \in [0, 1]$, and $\mathbf{c}_t$ $\in$ $\real^{L \times h}$ represents the driving conditions for $L$ consequent frames. This vector field generates the flow $\varphi_t: [0, 1] \times \real^{L \times d} \to \real^{L \times d}$ of $L$ frames by solving ODE (\cref{eq:ode}). As illustrated in \cref{fig:cfmt}, we build our vector field predictor upon the transformer encoder \cite{transformer} architecture. Specifically, we adopt DiT \cite{dit} architecture, but decouple frame-wise conditioning from time-axis attention mechanism, which enables us to model temporally consistent motion latents.

In DiT \cite{dit}, distinct semantic tokens are modulated by a single diffusion time step embedding and class embedding through adaptive layer normalization (AdaLN). In contrast, our vector field predictor modulates each $l$-th input latent with its corresponding $l$-th condition and then combines their temporal relations through a masked self-attention layer that attends to $2 \cdot T$ neighboring frames. Formally, for each $l$-th frame, frame-wise AdaLN and frame-wise gating are computed by
\begin{align}
    \gamma_i^{l} \times \text{LN}(X_t^{l}) + \beta_i^{l} \in \real^h \quad \text{and} \quad \alpha_i^{l} \times X_{t}^{l} \in \real^h,
\end{align}
respectively, where $i$ $\in$ $\{1, 2\}$, $h$ is the hidden dimension, $\text{LN}(\cdot)$ denotes layer norm \cite{layernorm}, and $X_t^{l}$ is the $l$-th input for each operation at flow time $t$ $\in$ $[0,1]$. The coefficients $\alpha_i^{l}, \beta_i^{l}, \gamma_i^{l} \in \real^{h}$ are computed from the condition $\mathbf{c}_t^l$ $\in$ $\real^{h}$ through a linear layer, \textit{ToScaleShift}, as depicted in \cref{fig:cfmt}.

\noindent \textbf{Speech-driven Emotion Enhancement~}~ \textit{How can we make talking motions more expressive and natural?} During talking, humans naturally reflect their emotions through their voices, and these emotions influence talking motions. For instance, a person who speaks sadly may be more likely to shake the head and avoid eye contact. This non-verbal motion derived from emotions crucially impacts the naturalness of a talking portrait.
\\\indent Existing works \cite{mead, eamm, vasa_1} use image-emotion paired data or image-driven emotion predictor \cite{hsemotion} to generate the emotion-aware motion. In contrast, we incorporate speech-driven emotions, a more intuitive way of controlling emotion for audio-driven talking portrait. Specifically, we utilize a pre-trained speech emotion predictor \cite{speech2emotion} that produces softmax probabilities of seven distinct emotions: \textit{angry, disgust, fear, happy, neutral, sad, and surprise}, which we then input into the vector field predictor.
\\\indent However, as people do not always speak with a single, clear emotion, determining emotions solely from audio is often ambiguous \cite{eamm}. Naive introduction of speech-driven emotion can make emotion-aware motion generation more challenging. To address this issue, we inject the emotions together with other driving conditions at training phase and modify them at inference phase.

\noindent \textbf{Driving Conditions~}~ We concatenate the audio representation $a^{1:L}$ $\in$ $\real^{L \times d_a}$ of a pre-trained Wav2Vec2.0 \cite{wav2vec2}, the speech emotion label $w_e$ $\in$ $\real^7$, and the source motion latent $w_{r\to S}$ $\in$ $\real^{d}$. Next, we add the flow time step embedding $\text{Emb}(t)$ $\in$ $\real^{h}$ to these conditions, producing $\mathbf{c}_t$ $\in$ $\real^{L \times h}$ via a linear layer, \textit{ToCondition}, as depicted in \cref{fig:overview}, where $\text{Emb}(t)$ is computed using the sinusoidal position embedding \cite{transformer}. \\
\noindent \textbf{Training~}~ We train FLOAT by reconstructing a target vector field computed from driving frames using the corresponding audio segments and a source motion latent. We choose a pair of driving motions and corresponding audio ($w_{r \to D^{1:L}}$, $a^{1:L}$), and construct the target vector field $u_t(x|w_{r\to D^{1:L}}) = w_{r\to D^{1:L}} - x_0$ $\in$ $\real^{L \times d}$ with noisy input $\varphi_t(x_0) = (1-t) x_0 + t w_{r\to D^{1:L}}$ ($t$ $\sim$ $\mathcal{U}[0,1]$ and $x_0$ $\sim$ $\mathcal{N}(0^{1:L}, I)$).

For smooth transitions of sequences longer than the window length $L$, we incorporate last $L'$ audio features and motion latents $w_{r \to D^{-L':0}}$ from the preceding window as additional input. 

The flow matching objective $\loss_{\text{OT}}(\theta)$ is defined by
\begin{equation}
\begin{aligned}
    \loss_{\text{OT}}(\theta) & = \| v_t^{1:L}( x_t, \mathbf{c}_t; \theta) - u_t(x|w_{r\to D^{1:L}}) \|, \\
    & + \| v_t^{-L':0}(x_t, \mathbf{c}_t; \theta) - w_{r \to D^{-L':0}} \|,
\end{aligned}
\end{equation}
where $x_t$ $:=$ $[ w_{r \to D^{-L':0}}|~\varphi_{t}(x_0)]$ $\in$ $\real^{(-L' + L) \times d}$ is the concatenated input, $\mathbf{c}_t$ $\in$ $\real^{(-L'+L) \times h}$ is the driving condition consisting of $[t, w_{r \to S},  w_e, a^{1:L}, a^{-L':0}]$. Note that $w_e$ and $w_{r \to S}$ are shared across the $L'+L$ frames. We incorporate a velocity loss \cite{mdm} to supervise temporal consistency: 
\begin{align}
    \loss_{\text{vel}}(\theta) = \| \Delta v_t - \Delta u_t\|, \label{eq:vel}
\end{align}
where $\Delta v_t$ and $\Delta u_t$ are the one-frame difference along the time-axis for the prediction $v_t$ $\in$ $\real^{(-L'+L)\times d}$ and the target $[w_{r \to D^{-L':0}} |~ u_t]$ $\in$ $\real^{(-L'+L) \times d}$, respectively.

The total objective $\loss_{\text{total}}(\theta)$ is
\begin{align}
    \loss_{\text{total}}(\theta) = \lambda_{\text{OT}} \loss_{\text{OT}}(\theta) + \lambda_{\text{vel}} \loss_{\text{vel}}(\theta),
\end{align} 
where $\lambda_{\text{OT}}$ and $\lambda_{\text{vel}}$ are the balancing coefficients. During training, we apply dropout to $w_r$, $w_e$, and $a^{1:L}$ with a probability of $0.1$ for CFV. Additionally, we apply dropout to the preceding audio and motion latents with a probability $0.5$ for smooth transition in the initial window.

\begin{figure*}[t]
\begin{center}
    \captionof{table}{Quantitative comparison results with state-of-the-art methods on HDTF \cite{hdtf} / RAVDESS \cite{ravdess}. The best result for each metric is in \textbf{bold}, and the second-best result is \underline{underlined}. \hfill \textit{$^{\dagger}$: evaluated with raw $256\times256$ resolution outputs.}} \label{tab:quantitative_hdtf}
    \vspace*{-2mm}
    \resizebox{0.99\textwidth}{!}{
        \begin{tabular}{l | c c c c c | c c}
        \toprule
        \multicolumn{1}{c|}{\multirow{2}{*}{Method}} & \multicolumn{5}{c|}{Image \& Video Generation} & \multicolumn{2}{c}{Lip Synchronization} \\
        \cline{2-8}
        \multicolumn{1}{c|}{} & \textbf{FID} $\downarrow$ & \textbf{FVD} $\downarrow$ & \textbf{CSIM} $\uparrow$ & \textbf{E-FID} $\downarrow$ & \textbf{P-FID} $\downarrow$ & \textbf{LSE-D} $\downarrow$ & \textbf{LSE-C} $\uparrow$  \\
        \hline
        SadTalker$^{\dagger}$   \cite{sadtalker}    & 71.952 / 119.430 & 339.058 / 376.294 & 0.644 / 0.644 & 1.914 / 3.500 & 1.456 / 2.045 & 7.947 / \underline{7.273} & 7.305 / 4.748 \\
        EDTalk$^{\dagger}$ \cite{edtalk} & 50.078 / 75.020 & 211.284 / 304.933 & 0.626 / 0.676 & 1.579 / 3.468 & 0.054 / 0.090 & 8.123 / 7.682 & 7.623 / \underline{5.318} \\ 
        \hline        
        AniTalker$^{\dagger}$   \cite{anitalker}    & 39.512 / 70.430 & \underline{184.454} / \underline{265.341} & 0.643 / 0.725 & 1.830 / 2.330 & 0.092 / 0.126 & 7.907 / 8.176 & 7.288 / 4.555 \\
        Hallo                   \cite{hallo}        & \underline{25.363} / \underline{57.648} & 197.196 / 375.557 & \textbf{0.869} / \textbf{0.860} & \textbf{1.039} / 2.492 & 0.037 / 0.050 & \underline{7.792} / 7.613 & \underline{7.582} / 4.795 \\
        EchoMimic               \cite{echomimic}    & 33.552 / 81.839 & 296.757 / 320.220 & 0.823 / 0.805 & 1.234 / 3.201 & \textbf{0.023} / \underline{0.047} &  8.903 / 8.161 & 6.242 / 4.144 \\
        \hline
        \textbf{FLOAT (Ours)}                        & \textbf{21.100} / \textbf{31.681} & \textbf{162.052} / \textbf{166.359}  & \underline{0.843} / \underline{0.810} & \underline{1.229} / \textbf{1.367} & \underline{0.032} / \textbf{0.031} & \textbf{7.290} / \textbf{6.994} & \textbf{8.222} / \textbf{5.730} \\
        \bottomrule
        \end{tabular}
        }
\end{center}
\begin{center}
    \includegraphics[width=0.98\textwidth]{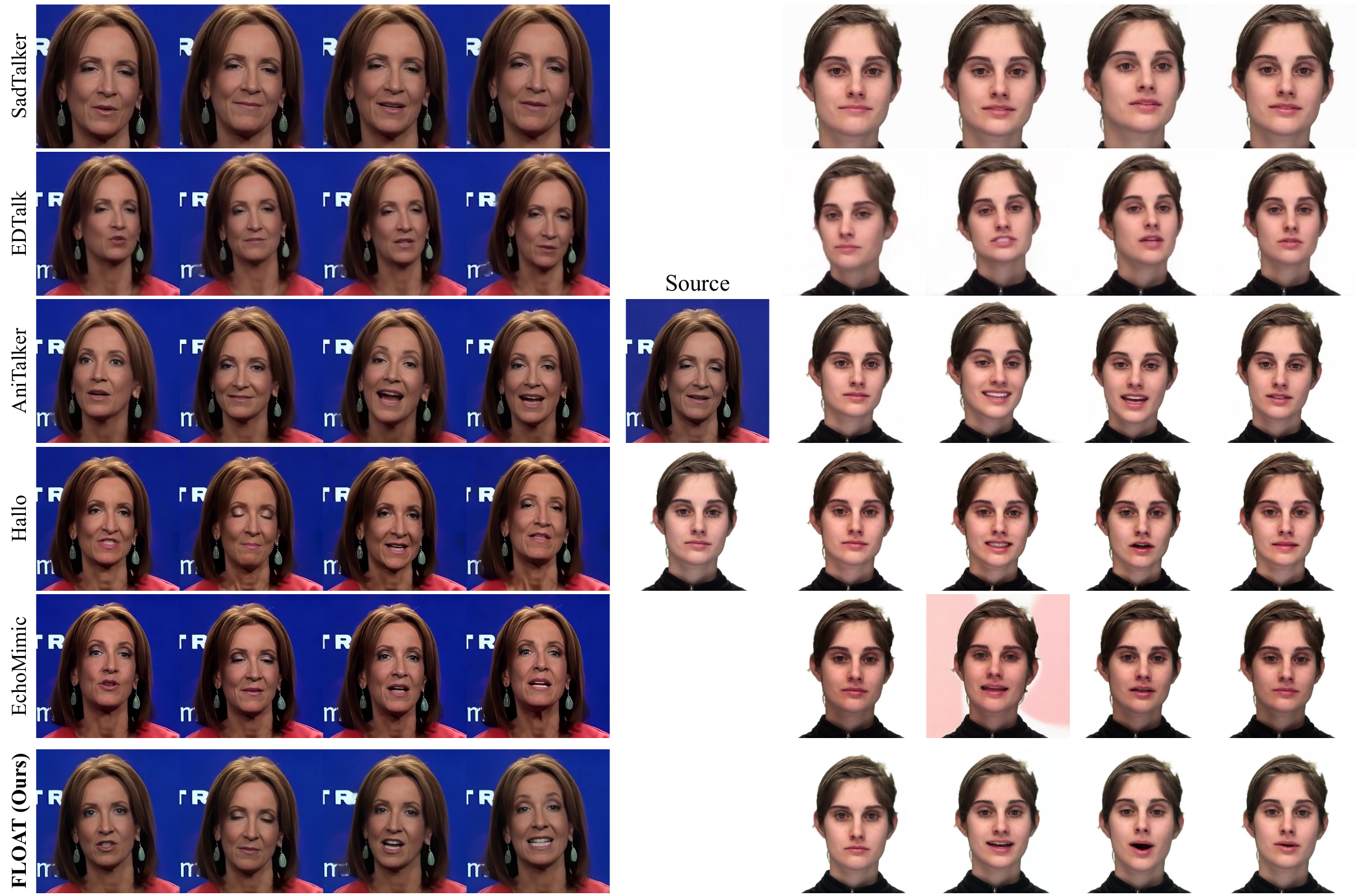}
    \vspace*{-2mm}
    \caption{Qualitative comparison results with state-of-the-art methods on HDTF \cite{hdtf} / RAVDESS \cite{ravdess}. Please refer to supplementary videos. Note that we additionally provide a video comparison with \textbf{EMO} \cite{emo} and \textbf{VASA-1} \cite{vasa_1} using their video demonstration.}
    \label{fig:sota_compare}
    \vspace*{-4mm}
\end{center}
\end{figure*}

\noindent \textbf{Inference~}~ During inference, we sample the generating vector field from noise $x_0$, using the driving conditions $w_{r \to S}$, $w_e$, and $a^{1:L}$, as well as the $L'$ frames of preceding audio and generated motion latents.

We extend the CFV \cite{lcfm} to an incremental CFV to separately adjust the audio and emotion, inspired by \cite{instructpix2pix}: 
\begin{align}
    \tilde{v}_t & \approx v_t(x_0, \mathbf{c}_t|_{\{a^{1:L}, w_e\}}) \nonumber \\
    & + \gamma_a \left[ v_t(x_0, \mathbf{c}_t |_{w_e} ) - v_t(x_0, \mathbf{c}_t|_{\{a^{1:L}, w_e\}}\right] \nonumber \\
    & + \gamma_e \left[ v_t(x_0, \mathbf{c}_t) - v_t(x_0, \mathbf{c}_t |_{w_e}) \right],
\end{align}
where $\gamma_a$ and $\gamma_e$ are the guidance scales for audio and emotion, respectively. $\mathbf{c}_t|_{\{x, y\}}$ denotes the driving condition without the condition $x$ and $y$. We set $\gamma_a=2$ and $\gamma_e=1$ based on the ablation studies on $\gamma_a$ and $\gamma_e$ provided in supplementary materials.

After sampling, ODE solver receives the estimated vector field to compute the motion latents through numerical integration. We empirically find that FLOAT can generate reasonable motion with around 10 number of function evaluations (NFE). Please refer to supplementary videos.

Lastly, we add the source identity latent to the generated motion latents and decode them into video frames using the motion latent decoder.

%% file: sec/4_experiments.tex
\section{Experiments}
\subsection{Dataset and Pre-processing} \label{sec:dataset}
For training the motion latent auto-encoder, we use three open-source datasets: \textbf{HDTF} \cite{hdtf}, \textbf{RAVDESS} \cite{ravdess}, and \textbf{VFHQ} \cite{vfhq}. When training FLOAT, we exclude VFHQ because it does not support the synchronized audio. HDTF \cite{hdtf} is for high-definition talking face generation, containing videos of over 300 unique identities. RAVDESS \cite{ravdess} includes more than 2,400 emotion-intensive videos of 24 different identities. VFHQ \cite{vfhq} is designed for high-resolution video super-resolution and includes a large number of unique identities, which compensates the limited number of identities of the preceding datasets. Following the strategy of \cite{fomm}, we first convert each video to 25 FPS and resample the audio into 16 kHz. Then, we crop and resize the facial region to $512^2$ resolution \cite{face_alignment}. After the pre-processing, for HDTF, we use a total of 11.3 hours of 240 videos featuring 230 different identities for training, and videos of 78 different identities, each 15 seconds long, for test. For RAVDESS, we use videos of 22 identities for training, and videos of the remaining 2 identities for test, with each 3-4 seconds long and representing 14 emotional intensities. Note that the identities in the training and test are disjoint in both datasets. 

\subsection{Implementation Details} \label{sec:implimentation}
The motion latent dimension is set to $d$ $=$ $512$ with $M$ $=$ $20$ distinct orthogonal directions. For the vector predictor, we use 8 attention heads, a hidden dimension $h$ $=$ $1024$, and an attention window length $T$ $=$ $2$. Considering the length of the training video clips, we set $L$ $=$ $50$ frames with preceding $L'$ $=$ $10$ frames at once, encompassing 2.4 seconds of video. We employ the Adam optimizer \cite{adam} with a batch size of 8 and  a learning late of $10^{-5}$. We use $L1$ distance for the norm $\|\cdot\|$ in the training objective. We set the balancing coefficients to $\lambda_{\text{OT}}$ $=$ $\lambda_{\text{vel}}$ $=$ $1$. The entire training takes about $2$ days for $2,000$k steps on a single NVIDIA A100 GPU. We use Euler method \cite{cfm} for the ODE solver. 

\subsection{Evaluation} \label{sec:evaluation}
\noindent \textbf{Metrics and Baselines~}~ For evaluating the image and video generation quality, we measure Fréchet Inecption Distance (\textbf{FID}) \cite{fid} and 16 frames Fréchet Video Distance (\textbf{FVD}) \cite{fvd}. For facial identity, expression and head motion, we measure Cosine Similarity of identity embedding (\textbf{CSIM}) \cite{arcface}, Expression FID (\textbf{E-FID}) \cite{emo} and Pose FID (\textbf{P-FID}), respectively. Lastly, we measure Lip-Sync Error Distance and Confidence (\textbf{LSE-D} and \textbf{LSE-C} \cite{wav2lip}) for audio-visual alignment.

We compare our method with state-of-the-art audio-driven talking portrait methods whose official implementations are publicly available. For non-diffusion methods, we compare with \textbf{SadTalker} \cite{sadtalker} and \textbf{EDTalk} \cite{edtalk}. For diffusion methods, we compare with \textbf{AniTalker} \cite{anitalker}, \textbf{Hallo} \cite{hallo}, and \textbf{EchoMimic} \cite{echomimic}.

\noindent \textbf{Comparison Results~}~ In \cref{tab:quantitative_hdtf} and \cref{fig:sota_compare}, we show the quantitative and qualitative comparison results, respectively. FLOAT outperforms other methods on most of the metrics and visual quality in both datasets.

Additionally, we provide video comparison results with \textbf{EMO} \cite{emo} and \textbf{VASA-1} \cite{vasa_1} in the supplementary materials, using their demonstration videos due to the infeasibility of direct implementation.

\begin{figure}[h]
\begin{center}
    \includegraphics[width=0.96\linewidth]{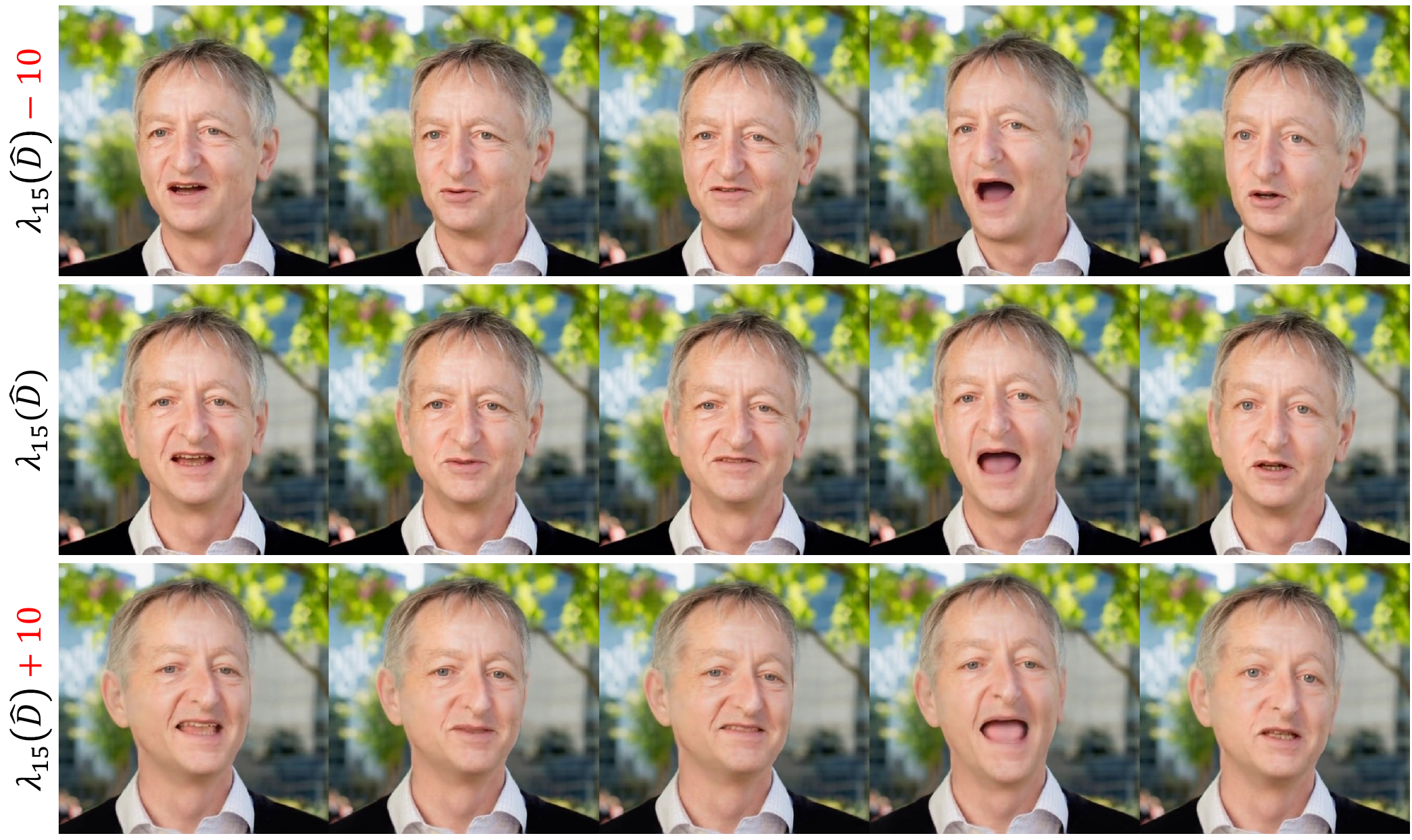}
    \vspace*{-3mm}
   \caption{Test-time pose editing using $\lambda$-control ($\lambda_{15}(\hat{D}) \pm 10$).} \label{fig:lambda_control}
\end{center}
    \vspace*{-6mm}
\end{figure}

 \subsection{Applications} \label{sec:further}
\textbf{Test-time Pose Editing via Orthonormal Basis $V$~}~ Since FLOAT learns the underlying motion latent structure, it is natural to assume that for any \textit{sampled} motion latent $w_{r\to\hat{D}}$, there exist motion coefficients $\{\lambda_m (\hat{D})\}_{m=1}^{M}$ satisfying the representation in \cref{eq:motion}: $w_{r \to \hat{D}}$ $=$ $\sum_{m=1}^{M} \lambda_m (\hat{D}) \cdot \mathbf{v}_m$.

We can always compute these coefficients in \textit{closed form} by taking inner products between the sampled motion $w_{r\to\hat{D}}$ and the learned orthonormal basis $V$:
\begin{equation}
     \langle w_{r\to\hat{D}}, ~\mathbf{v}_k \rangle = \langle \sum_{m=1}^{M} \lambda_m (\hat{D}) \cdot \mathbf{v}_m, ~\mathbf{v}_k \rangle = \lambda_k (\hat{D}), \label{eq:lambda}
\end{equation}
where $\langle \mathbf{v}_m, \mathbf{v}_k\rangle = \delta_{m,k}$ and $\delta$ is Kronecker delta. At this point, we can edit the sampled motions by editing the corresponding coefficients (e.g., via linear operation) and combining them back into the motion latent. As shown in \cref{fig:lambda_control}, it allows us to control head direction without interfering with other motions due to the orthogonality of the basis. We refer to this test-time editing technique as \textit{$\lambda$-control}.

\begin{figure*}[t]
\begin{center}
    \begin{subfigure}{0.455\linewidth}
        \includegraphics[width=\textwidth]{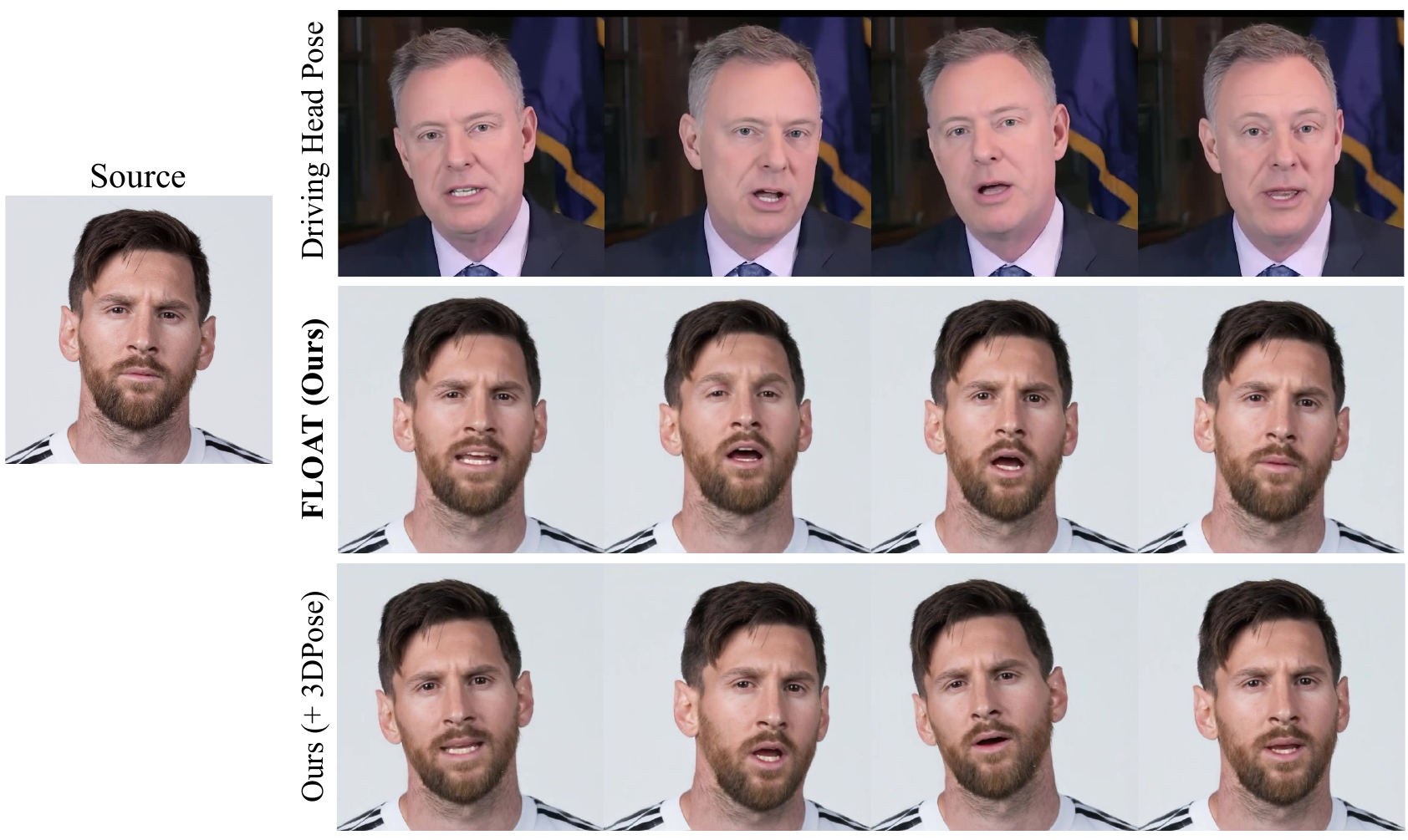}
    \end{subfigure}
    \begin{subfigure}{0.48\linewidth}
        \includegraphics[width=\textwidth]{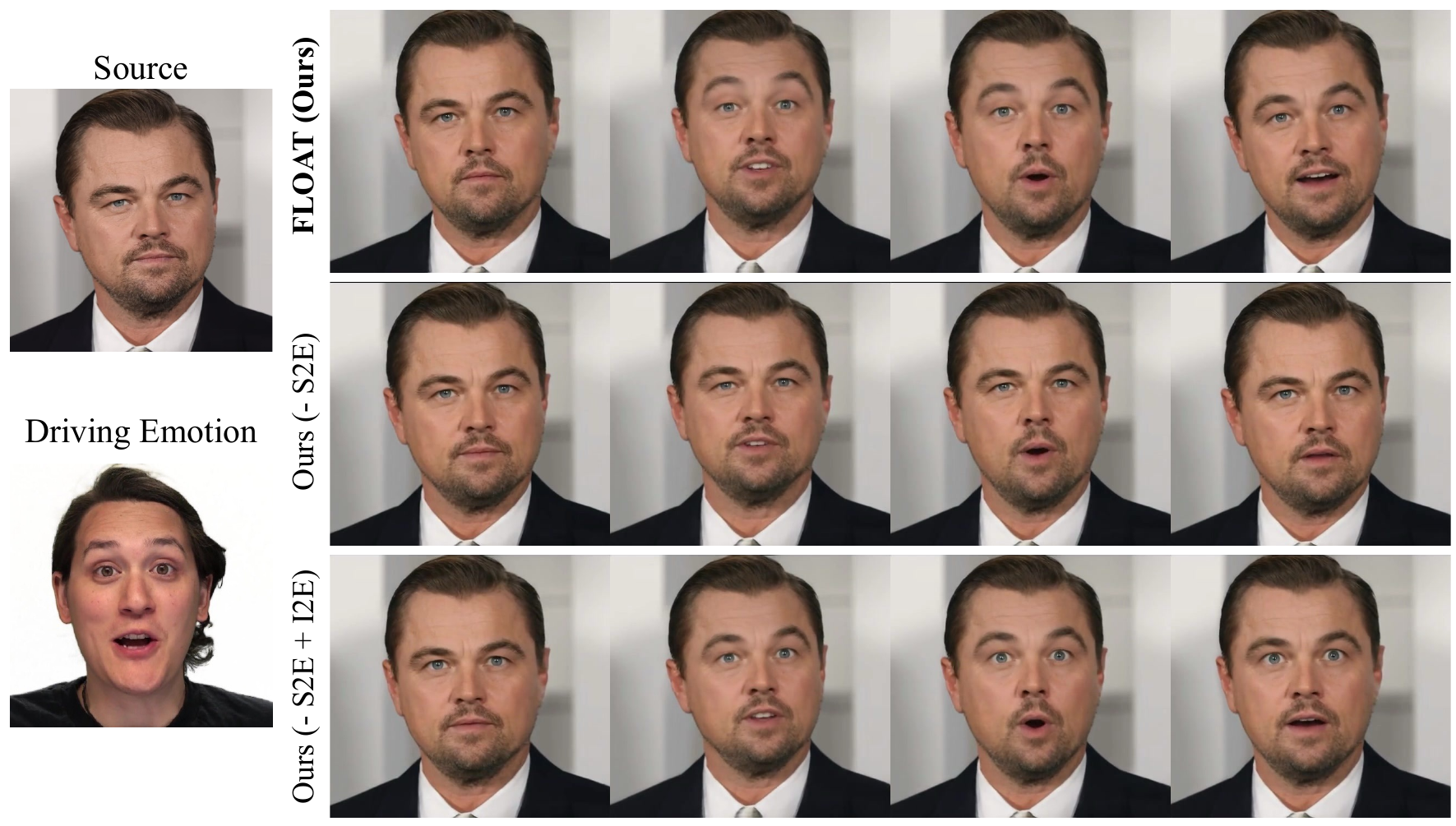}
    \end{subfigure}
    \vspace*{-1mm}
   \caption{Additional conditioning results of FLOAT. \textit{3DPose}, \textit{S2E}, and \textit{I2E} denote 3D head pose parameters \cite{bfm}, speech-to-emotion \cite{speech2emotion}, and image-to-emotion \cite{hsemotion}, respectively.} \label{fig:additional_condition}
\end{center}
    \vspace*{-8mm}
\end{figure*}

\noindent\textbf{Additional Driving Signals~}~ In \cref{fig:additional_condition} and \cref{tab:additional_condition}, we experiment with additional driving conditions, \textit{head poses} and \textit{image-driven emotion labels}, to explore additional controllability in our method. We employ 3DMM head pose parameters $p$ $\in$ $\real^6$ \cite{3dmm} extracted by \cite{bfm}. We concatenate a sequence of pose parameters $p^{1:L}$ $\in$ $\real^{L \times 6}$ with the other driving conditions, and then map them to $c_t^{1:L}$ $\in$ $\real^{L \times h}$. We also experiment on image-driven emotion \cite{hsemotion} for frame-wise emotion control rather than the long-term emotion enhancement. FLOAT can effectively accommodate these additional conditions, highlighting its flexibility across diverse control signals. 

\begin{figure}
    \centering
    \includegraphics[width=0.99\linewidth]{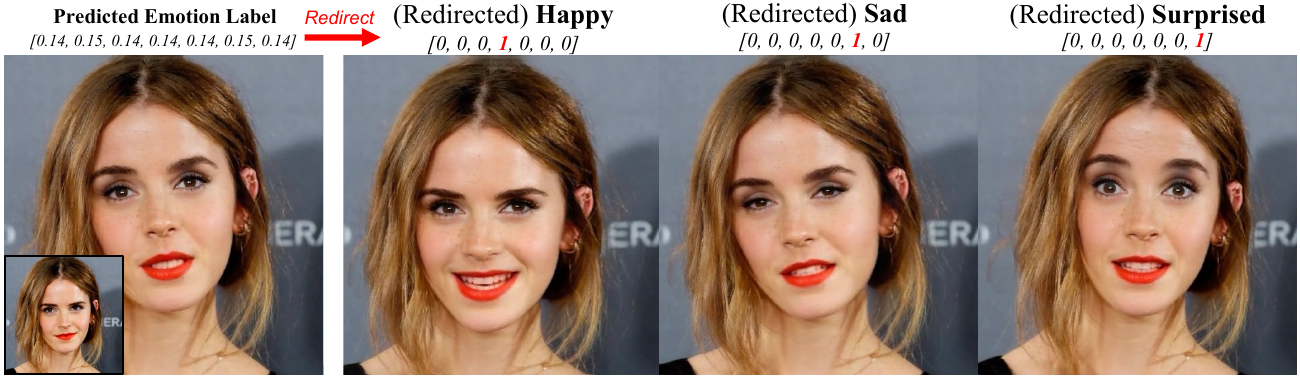}
    \vspace*{-1mm}
    \caption{Redirecting the unclear emotion prediction to a desirable one-hot encoding, which can be further intensified by the CFV.}
    \label{fig:emotion_redirection}
    \vspace*{-4mm}
\end{figure}

\noindent\textbf{Redirecting Speech-driven Emotion~} Since FLOAT learns diverse emotions in the emotion-intensive data distribution \cite{ravdess}, the generated emotion-aware motion can be modified by \textit{redirecting} the speech-driven emotion label toward a different emotion at inference time. As illustrated in \cref{fig:emotion_redirection}, this technique is particularly beneficial for manual redirection when the emotion predicted from speech is complex or ambiguous.

\begin{table}
    \centering
    \captionof{table}{Quantitative results of FLOAT with additional conditions (HDTF \cite{hdtf} / RAVDESS \cite{ravdess}). \textit{S2E}, \textit{I2E}, and \textit{3DPose} denote speech-to-emotion \cite{speech2emotion}, image-to-emotion \cite{hsemotion}, and 3DMM pose parameters \cite{bfm}, respectively.}  \label{tab:additional_condition}
    \vspace*{-2mm}
    \resizebox{0.99\linewidth}{!}{
        \begin{tabular}{l | l| c c c c c}
        \toprule
        \multicolumn{2}{l|}{Configurations} & \textbf{FID} $\downarrow$ & \textbf{FVD} $\downarrow$ & \textbf{E-FID} $\downarrow$ & \textbf{P-FID} $\downarrow$ & \textbf{LSE-D} $\downarrow$\\
        \cline{1-7} 
        A  & \textbf{FLOAT (Ours)} & 21.100 / 31.681 & 162.052 / 166.359 & 1.229 / 1.367 & 0.032 / 0.031 & 7.290 / 6.994\\
        \hline        
        B  & A + 3DPose & 19.721 / 29.721 & 126.663 / 112.894 & 0.926 / 1.152  & 0.012 / 0.016 & 7.516 / 7.047 \\
        C  & A - S2E & 21.235 / 32.035 & 155.032 / 166.866 & 1.254 / 1.502 & 0.031 / 0.025 & 7.264 / 7.222 \\
        D  & A - S2E + I2E & 21/528 / 31.609 & 158.577 / 162.369 & 1.158 / 1.305 & 0.034 / 0.022 & 7.183 / 7.150\\
        \bottomrule
        \end{tabular}
        }
\vspace*{-1mm}
\end{table}

\subsection{Ablation Studies} \label{sec:ablation}
\noindent \textbf{Ablation on Frame-wise AdaLN~}~ We compare frame-wise AdaLN (and gating) followed by masked self-attention to separate conditioning from attending, with a cross-attention that performs conditioning and attending simultaneously. As shown in \cref{tab:ablation}, both approaches achieve competitive image and video quality, while frame-wise AdaLN provides better expression generation and lip synchronization. We observe that frame-wise AdaLN can achieve more diverse head motions than the cross-attention. Please refer to supplementary videos.

\begin{figure}
    \centering
    \vspace*{-2mm}
    \includegraphics[width=0.95\linewidth]{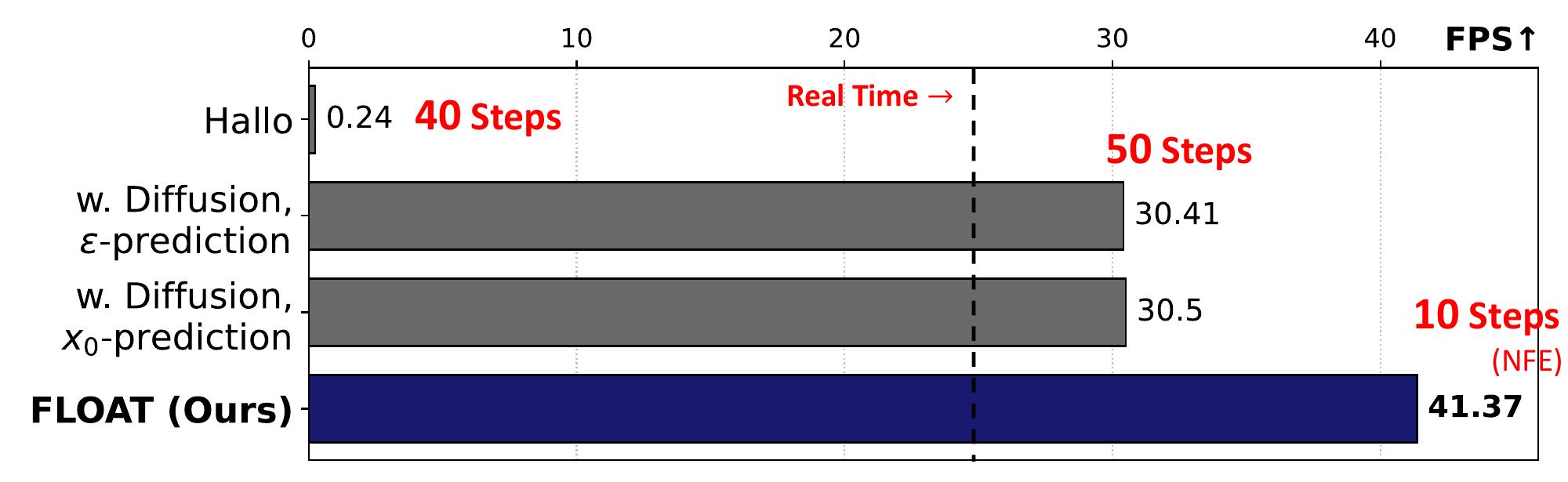}
    \vspace*{-3mm}
    \caption{Comparison of the forward pass efficiency. We compute FPS on a single NVIDIA V100 GPU.}\label{fig:fps}
    \vspace*{-5mm}
\end{figure}

\noindent \textbf{Ablation on Flow Matching~}~  We compare flow matching with two types of diffusion models: $\epsilon$-prediction (noise) and $x_0$-prediction (signal) \cite{dalle2, mdm}. In both cases, we adopt our vector predictor architecture as denoising networks. We adopt diffusion training settings of VASA-1 \cite{vasa_1} (500 diffusion steps with a cosine noise scheduler \cite{iddpm} and 50 DDIM denoising steps) for the indirect comparison with \cite{vasa_1}. Notably, diffusion and flow matching achieve competitive results on image quality while the latter achieves the better lip synchronization.
In \cref{fig:fps}, we compare the forward pass efficiency by measuring frames per second (FPS) of each model. Thanks to the compact motion latent representation and OT-based flow matching, FLOAT achieves the highest FPS, superior lip-sync performance, dynamic head motion, and the lowest NFEs.

\begin{table}[t]
    \centering
    \captionof{table}{Ablation studies of FLOAT on HDTF \cite{hdtf}. The best result for each metric is in \textbf{bold}, and the second-best result is \underline{underlined}.} \label{tab:ablation}
    \vspace*{-1mm}
    \resizebox{0.95\linewidth}{!}{
        \begin{tabular}{l | c c c c c}
        \toprule
        Method & \textbf{FID} $\downarrow$ & \textbf{FVD} $\downarrow$ & \textbf{E-FID} $\downarrow$ & \textbf{LSE-D} $\downarrow$ & \# \textbf{NFEs} $\downarrow$ \\
        \cline{2-5}
        \hline
        Ours (w. Cross-Attn.)                                       & 21.873             & 162.702              & 1.452             & \underline{7.757}   & \textbf{10} \\
        \hline        
        Ours (w. Diff., $\epsilon$-pred.)                      & \underline{21.190} & \textbf{161.666}     & \textbf{1.213}    & 9.922 & 50 \\
        Ours (w. Diff., $x_0$-pred.)                             & 21.697             & 162.847              & 1.278 & 9.048 & 50 \\
        \hline
        \textbf{FLOAT (Ours)}                                           & \textbf{21.100} & \underline{162.052}     & \underline{1.229}     & \textbf{7.290} & \textbf{10}\\
        \bottomrule
        \end{tabular}
        }
    \vspace*{-3mm}
\end{table}

%% file: sec/5_conclusion.tex
\section{Conclusion} \label{sec:conclusion}
We proposed FLOAT, a flow matching based audio-driven talking portrait generation model leveraging a learned motion latent space. We introduced a transformer-based vector field predictor, enabling temporally consistent motion generation. Additionally, we incorporated speech-driven emotion labels into the motion sampling process to improve the naturalness of the audio-driven talking motions. FLOAT addresses current core limitations of diffusion-based talking portrait video generation methods by reducing the sampling time through flow matching while achieving the remarkable sample quality. Extensive experiments verified that FLOAT achieves state-of-the-art performance in terms of visual quality, motion fidelity, and efficiency.

\noindent \textbf{Discussion~}~ We leave further discussion considering \textit{limitations}, \textit{future work}, and \textit{ethical considerations} in the supplementary materials.

%% file: sec/6_suppl.tex
\clearpage

\appendix 

\noindent In this supplement, we first provide more details on motion latent auto-encoder in \cref{sec:phase1_supp}, regarding the model itself (\cref{sec:phase1_model_itself_supp}), methods for improving the fidelity of facial components (\cref{sec:improving_component_supp}), the training objective (\cref{sec:training_objective_phase1_supp}), and implementation details (\cref{sec:implementation_details_phase1_supp}).

In \cref{sec:phase2_supp}, we provide more details on FLOAT, regarding details on evaluation metrics (\cref{sec:metric_supp}), baselines (\cref{sec:baseline_supp}), and ablation studies (\cref{sec:phase2_experiment_supp}). 

In \cref{sec:phase2_additional_result_supp}, we provide additional results, including comparison results (\cref{sec:additional_comparison_supp}), out-of-distribution results (\cref{sec:ood}), and user study (\cref{sec:user}). 

Finally, we discuss ethical considerations, limitations, and future work in \cref{sec:discussion}.


\section{More on Motion Latent Auto-encoder} \label{sec:phase1_supp}
In this section, we provide more details on our motion latent auto-encoder, including its model architecture, dataset, and training strategy.

\subsection{Model} \label{sec:phase1_model_itself_supp}
We provide a detailed model architecture of our motion latent auto-encoder in \cref{fig:phase1_architecture}.

\begin{figure*}[t]
  \begin{center}
        \begin{subfigure}{0.16\linewidth}
    \includegraphics[width=\textwidth]{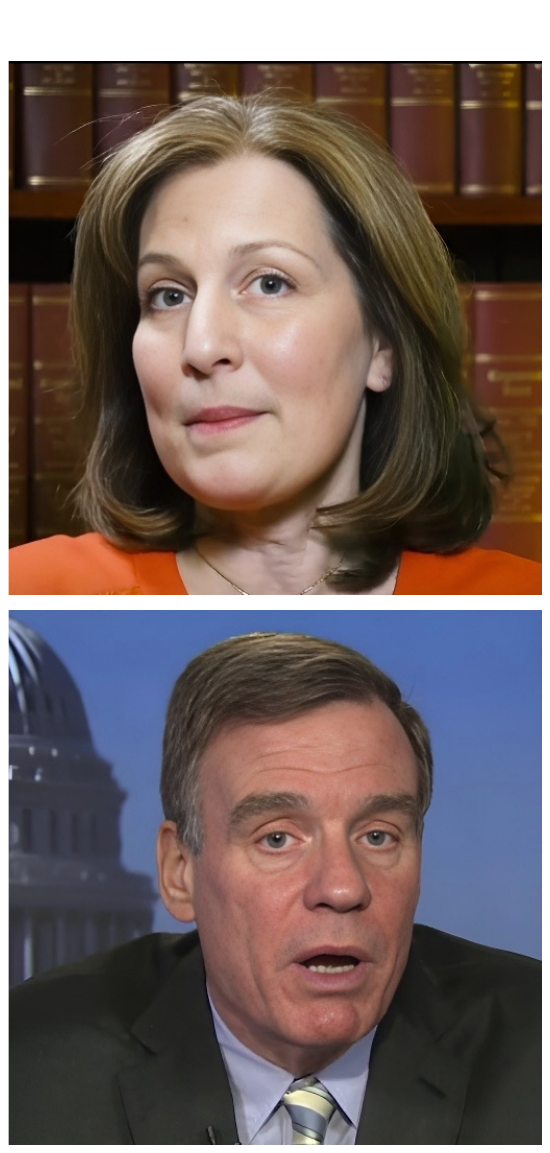}
    \caption{Source, $S$}
    \label{fig:short-1a}
  \end{subfigure}
  \begin{subfigure}{0.16\linewidth}
    \includegraphics[width=\textwidth]{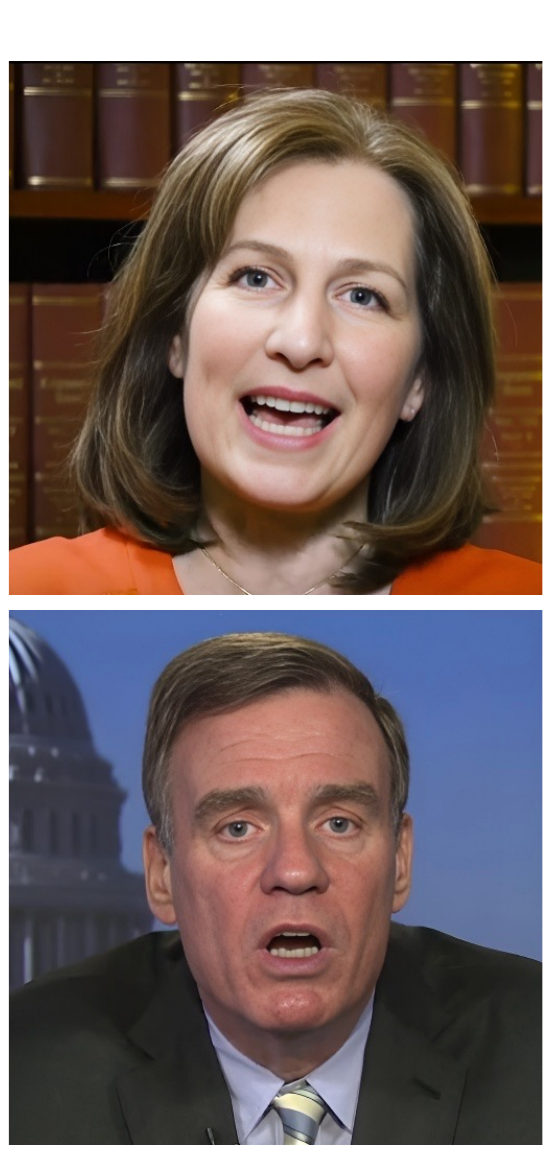}
    \caption{Driving, $D$}
    \label{fig:short-1b}
  \end{subfigure}
    \begin{subfigure}{0.16\linewidth}
    \includegraphics[width=\textwidth]{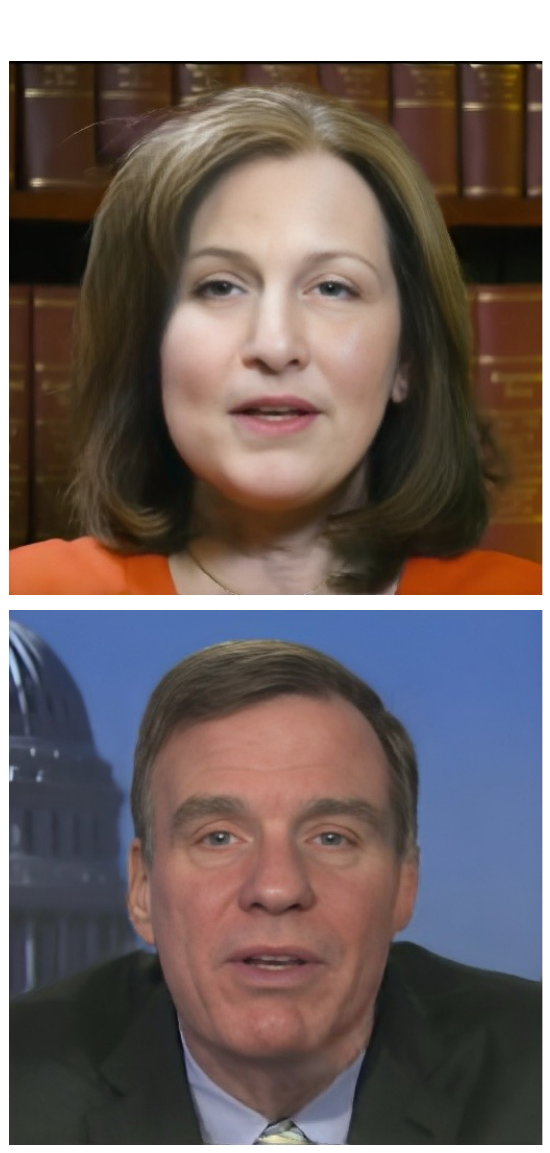}
    \caption{Identity, $w_{S\to r}$}
    \label{fig:short-1c}
  \end{subfigure}
    \begin{subfigure}{0.159\linewidth}
    \includegraphics[width=\textwidth]{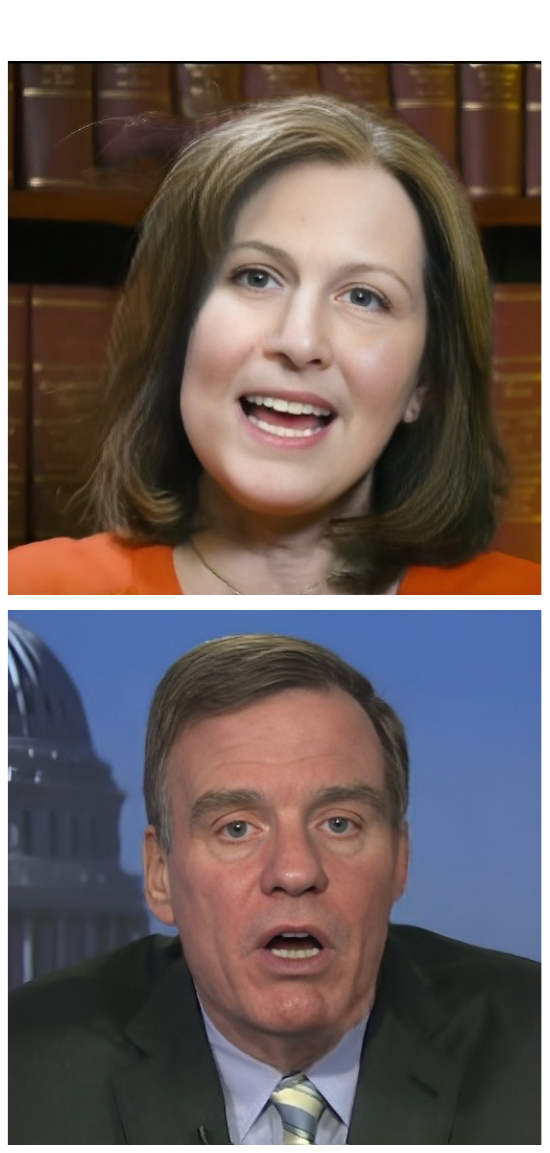}
    \caption{Reconstruction, $\hat{D}$}
    \label{fig:short-1d}
  \end{subfigure}
  \begin{subfigure}{0.16\linewidth}
    \includegraphics[width=\textwidth]{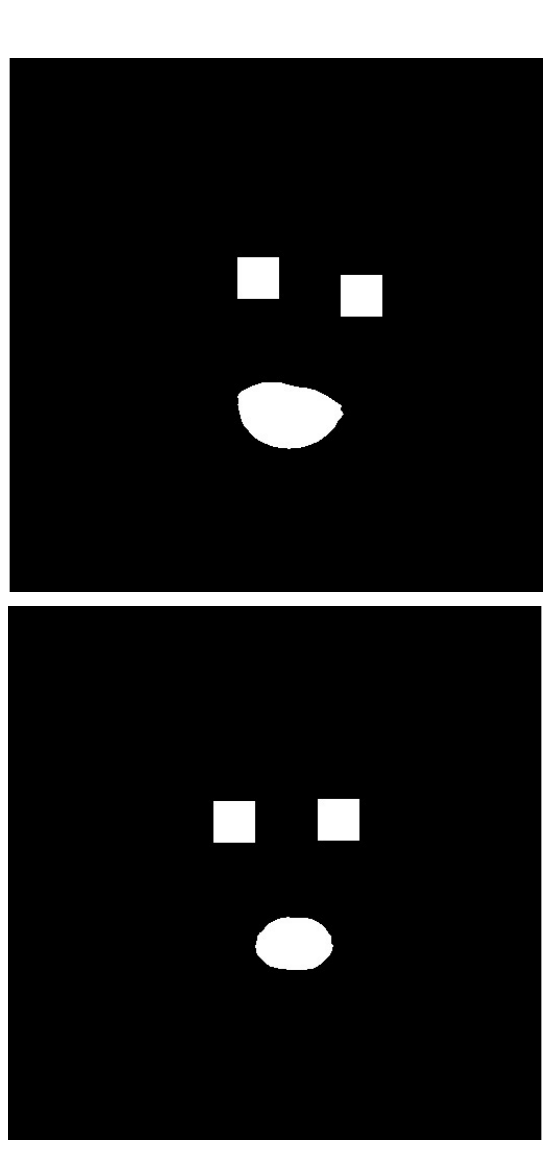}
    \caption{Component mask}
    \label{fig:short-b}
  \end{subfigure}
  \begin{subfigure}{0.16\linewidth}
    \includegraphics[width=\textwidth]{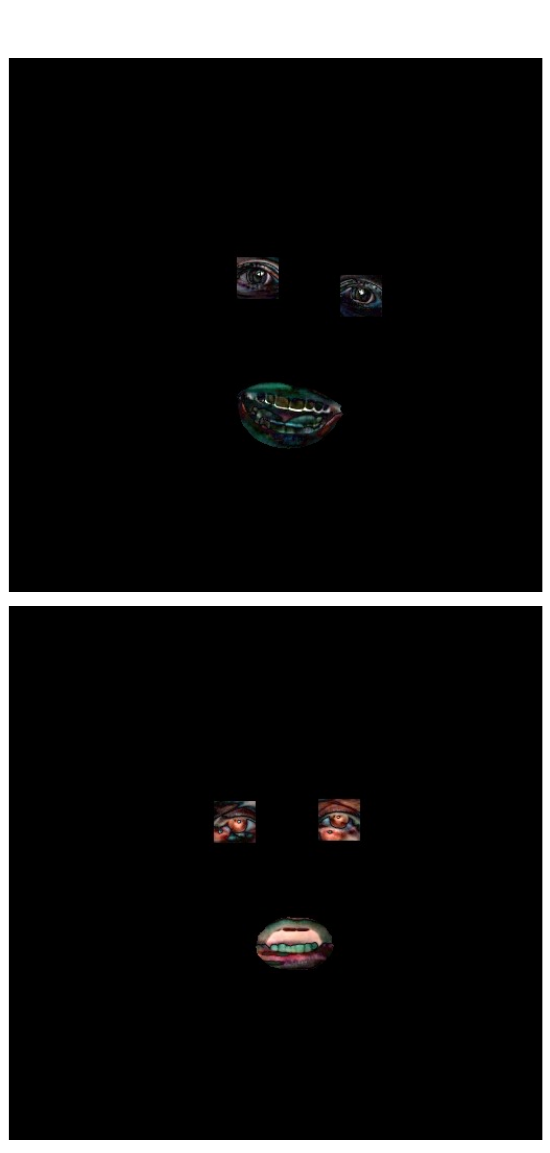}
    \caption{Component diff}
    \label{fig:short-c}
  \end{subfigure}
    \end{center}
    \vspace*{-5mm}
  \caption{Visualization results of the motion latent auto-encoder.}
   \label{fig:phase1_visualization}
\end{figure*}

In \cref{fig:short-1a}, \cref{fig:short-1b}, \cref{fig:short-1c}, and \cref{fig:short-1d}, we present visualization results of the latent decomposition \begin{align}
    w_S = w_{S \to r} + w_{r \to S} \in \real^{d}
\end{align}
of a source image $S$, following the approach of \cite{lia}. Notably, the identity latent $w_{r \to S}$ is decoded into image featuring the average head pose, expression, and field of view in pixel space.

\begin{figure}[h]
    \begin{center}
    \includegraphics[width=0.95\linewidth]{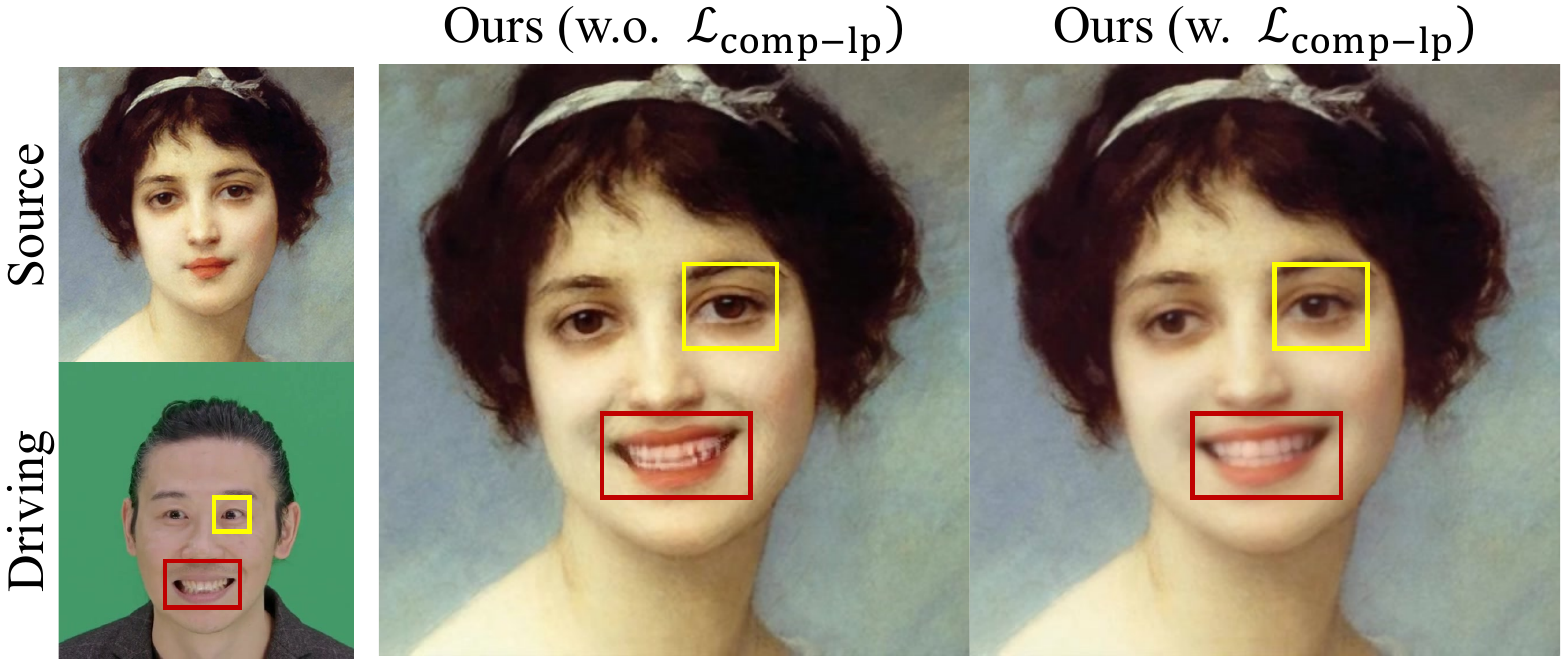}
    \vspace*{-2mm}
    \caption{Ablation study on Facial Component Loss $\loss_{\text{comp-lp}}$. It significantly improves the image fidelity of facial component (\eg, teeth, highlighted in red box) and fined-grained motion (eyeball movement, highlighted in yellow box).}
    \label{fig:wo_comp_loss}
    \end{center}
    \vspace*{-7mm}
\end{figure}

\begin{table*}[t]
\begin{center}
    \captionof{table}{Quantitative comparison result (Same-identity) of motion latent auto-encoders on HDTF \cite{hdtf} / RAVDESS \cite{ravdess} / VFHQ \cite{vfhq}. The best result for each metric is in \textbf{bold}. \hfill \textit{$^{\dagger}$: Results generated by official implementation ($256 \times 256$)}} \label{tab:phase1_quantitative}
    \vspace*{-2mm}
    \resizebox{0.98\textwidth}{!}{
        \begin{tabular}{l | c c c c c}
        \toprule
        \multicolumn{1}{c|}{Method} & \textbf{FID} $\downarrow$ & \textbf{FVD} $\downarrow$ & \textbf{LPIPS} $\downarrow$ & \textbf{E-FID} $\downarrow$ & \textbf{P-FID} $\downarrow$   \\
        \hline
        LIA$^{\dagger}$   \cite{lia}    & 47.481 / 67.541 / 89.209 & 172.195 / 130.836 / 342.964 & 0.184 / 0.122 / 0.245 & 1.279 / 1.153 / 1.106 & 0.120 / 0.005 / 0.013\\
        Ours (w.o. $\loss_{comp-lp}$)   & 21.061 / 28.866 / 46.950  & 150.340 / 103.145 / 299.757 & 0.110 / 0.072 / 0.165 & 1.369 / 1.157 / \textbf{0.872} & 0.011 / 0.010 / 0.014 \\
        \hline
        \textbf{Ours}                   & \textbf{19.803} / \textbf{23.350} / \textbf{43.992} & \textbf{147.089} / \textbf{100.345} / \textbf{291.560} & \textbf{0.108} / \textbf{0.062} / \textbf{0.161} & \textbf{1.334} / \textbf{1.053} / 1.006 & \textbf{0.010} / \textbf{0.008} / \textbf{0.012} \\
        \bottomrule
        \end{tabular}
        }
\end{center}
    \vspace*{-5mm}
\end{table*}

\subsection{Improving Fidelity of Facial Components} \label{sec:improving_component_supp}
\noindent \textbf{Facial Components: Texture vs. Structure~}~ As highlighted in face restoration work \cite{gfpgan}, facial components such as eyeballs and teeth play a important role in the perceptual quality of generated images. It treats the issue as a lack of \textit{texture} (lying in high frequencies) and mitigate it by introducing facial component discriminators with the gram matrix statistics matching. This approach is appropriate in face restoration, where training objective is to reconstruct a clear image from a degraded one that maintains the same spatial structure, ensuring that the low-frequency structure preserved.

However, in the context of training a motion auto-encoder, spatial mismatches are inevitably involved. Therefore, naively applying such discriminators proves ineffective. Instead, achieving high-fidelity facial components in a motion auto-encoder is more closely related to structural problems (lying in low frequencies) than to texture issues as shown in \cref{fig:short-c}.

\noindent {\textbf{Facial Component Perceptual Loss} $\loss_{\text{comp-lp}}$~}~ We introduce a simple yet effective \textit{facial component perceptual loss}, which leverages the standard perceptual loss $\loss_{\text{lp}}$ \cite{lpips} known for its ability to capture structural features lying in low frequencies. Formally, the facial component perceptual loss is defined by
\begin{align}
    \sum_{i=1}^{N} \frac{1}{|M_i|}\| M_i \otimes \phi_i(\hat{D}) - M_i \otimes \phi_i (D) \|_1,
\end{align}
where $D$ is the driving, $\hat{D}$ is the generated image, $N$ is the number of feature pyramid scales, $\phi_i(X)$ is the $i$-th feature of the input image $X$ computed by VGG-19 \cite{vgg, lpips}, $M_i$ is the binary mask of the facial components that has same size with $\phi_i(X)$, and $|M_i|$ is the sum of all values in the binary mask $M_i$. We adopt a single perceptual loss with $N=4$ scales of VGG-19 feature pyramids. It is worth noting that we mask all the multi-resolution features (not only the image).

To compute the facial component mask $M_i$, we utilize an off-the-shelf face segmentation model \cite{face_segmentation} for tight mouth regions and face landmark detector \cite{face_alignment} for the bounding box regions of the eyes as illustrated in \cref{fig:short-b}. 

In \cref{tab:phase1_quantitative}, we conduct ablation studies on motion latent auto-encoders. Notably, $\loss_{\text{comp-lp}}$ is consistently improves the image fidelity over three datasets. As illustrated in \cref{fig:wo_comp_loss}, an additional advantage of $\loss_{\text{comp-lp}}$ is its ability to directly supervise fine-grained motion (often neglected due to large head motion) such as eyeball movement without any external driving conditions such as eye-gazing direction \cite{emoportraits}. 

\subsection{Training Objective} \label{sec:training_objective_phase1_supp}
We train our motion latent auto-encoder by reconstructing a driving image $D$ from a source image $S$, both sampled from the same video clip.

The total loss function $\loss_{\text{total}}$ for the motion latent auto-encoder is defined as
\begin{align}
    \loss_{\text{total}} & = \loss_{L1} + \lambda_{\text{lp}}\loss_{\text{lp}} + \lambda_{\text{comp-lp}}\loss_{\text{comp-lp}} \nonumber \\
    & + \lambda_{\text{full-adv}} \loss_{\text{full-adv}} \nonumber \\
    & + \lambda_{\text{eye-adv}} \loss_{\text{eye-adv}} + \lambda_{\text{eye-FSM}} \loss_{\text{eye-FSM}} \nonumber \\
    & + \lambda_{\text{lip-adv}} \loss_{\text{lip-adv}} + \lambda_{\text{lip-FSM}} \loss_{\text{lip-FSM}},
\end{align}
where $\lambda_{\text{lp}}$, $\lambda_{\text{comp-lp}}$, $\lambda_{\text{eye-adv}}$, $\lambda_{\text{eye-FSM}}$, $\lambda_{\text{lip-adv}}$, $\lambda_{\text{lip-FSM}}$, and $\lambda_{\text{full-adv}}$ are the balancing coefficients. Here, $\loss_{L1}$ is the L1 loss, and $\loss_{\text{lp}}$ is the VGG-19 \cite{vgg} based multi-scale perceptual loss \cite{lpips} similar to $\loss_{\text{comp-lp}}$. We incorporate 2-scale discriminator $\loss_{\text{full-adv}}$ with the non-saturating loss:
\begin{align}
    \loss_{\text{full-adv}} = -\log [\text{Disc}_{\text{full}} (\hat{D})],
\end{align}
where $\text{Disc}$ denotes a discriminator adopted from \cite{stylegan2}. To improve the fidelity of the facial components, we also incorporate the facial component discriminators with the feature style matching (FSM) \cite{gfpgan}, 
\begin{align}
    \loss_{x\text{-adv}} & = -\log [\text{Disc}_{x} (\hat{D}_x)], \\
    \loss_{x\text{-FSM}} & = \| \text{Gram} (\psi(D_x)) - \text{Gram} (\psi(\hat{D}_x)) \|_1,
\end{align}
where $x \in \{\text{eye}, \text{lip}\}$. $D_x$ and $\hat{D}_x$ represent the region of interest (RoI) for the component $x$ in the driving $D$ and reconstruction $\hat{D}$, respectively. $\text{Gram}$ is a gram matrix calculation \cite{gram_matrix} and $\psi$ is the multi-resolution features extracted by the learned component discriminators.

\begin{figure*}[t]
    \begin{center}
    \includegraphics[width=0.99\linewidth]{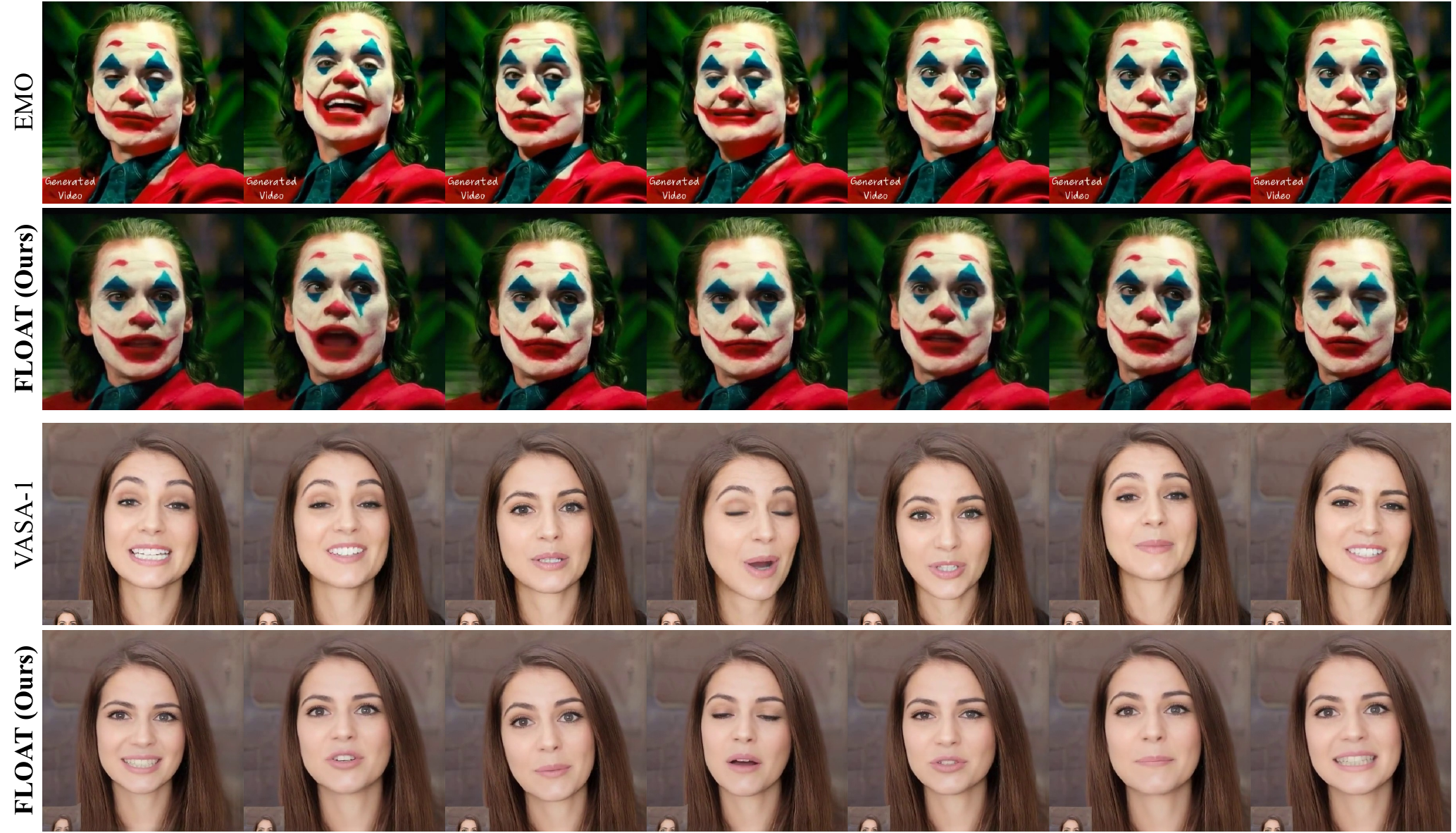}
    \vspace*{-2mm}
    \caption{Comparison results with \textbf{EMO} \cite{emo} and \textbf{VASA-1} \cite{vasa_1} based on their demonstration videos. Please note that their implementation are unavailable.}
    \label{fig:emo_vasa_1}
    \end{center}
    \vspace*{-7mm}
\end{figure*}

\subsection{Implementation Details} \label{sec:implementation_details_phase1_supp}
We set the balancing coefficients $\lambda_{\text{lp}}=10$, $\lambda_{\text{comp-lp}}=100$, $\lambda_{\text{eye-adv}}= 1$, $\lambda_{\text{eye-FSM}}=100$, $\lambda_{\text{lip-adv}} = 1$, $\lambda_{\text{lip-FSM}} = 100$, and $\lambda_{\text{full-adv}} = 1$. We employ Adam optimizer \cite{adam} with a batch size of 8 and a learning rate of $2\cdot 10^{-4}$. Entire training takes about 9 days for 460k steps on a single NVIDIA A100 GPU.

For training our motion latent auto-encoder, we use VFHQ \cite{vfhq} to supplement the limited number of identities provided by HDTF \cite{hdtf} and RAVDESS \cite{ravdess}. After the same pre-processing, remaining 14,362 video clips are used for training, and 49 video clips are used for test, respectively.

\section{More on FLOAT} \label{sec:phase2_supp}
In this section, we provide more details on FLOAT, including model, experiments, and further results.

In \cref{fig:phase2_architecture}, we provide a detailed model architecture for the driving conditions $\mathbf{c}_t$.

\subsection{Evaluation Metrics} \label{sec:metric_supp}
We provide further details of following metrics.
\begin{itemize}
    \item \textbf{LPIPS} \cite{lpips} is used to measure the perceptual similarity between reconstructed image and real image based on the pre-trained AlexNet features \cite{alexnet}.
    \item \textbf{FID} \cite{fid} aims to measure the distance between the feature distributions of real and generated datasets. It is computed as:
    \begin{align}
        \| \mu_r - \mu_g\|_2^2 + \text{Tr}(\Sigma_r + \Sigma_g - 2 (\Sigma_r \Sigma_g)^{\frac{1}{2}}) \label{eq:fid},
    \end{align}
    where $\mu_r$, $\Sigma_r$ and $\mu_g$, $\Sigma_g$ are the means and covariances of the pre-trained InceptionNet \cite{inception} features from the real and generated datasets, respectively.
    \item \textbf{FVD} \cite{fvd} is a variant of FID \cite{fid}, which is used to measure the spatio-temporal consistency between the real and generated datasets by leveraging the features of pre-trained video model \cite{i3d}. We compute this using 16 frames with a sliding window manner for each video.
    \item \textbf{CSIM} \cite{arcface} measures face similarity between the two face images by computing the cosine similarity between the pre-trained ArcFace features \cite{arcface} of two images.
    \item \textbf{E-FID} \cite{emo} aims to measure expression similarity by computing the FID score (\cref{eq:fid}) of 3DMM expression parameters (64-dim) \cite{bfm} of generated videos and real videos. 
    \item \textbf{P-FID} aims to measure the head pose similarity by computing the FID score (\cref{eq:fid}) of 3DMM pose parameters (6-dim) \cite{bfm} of generated videos and real videos.
    \item \textbf{LSE-D} and \textbf{LSE-C} \cite{wav2lip} measure lip synchronization using the pre-trained SynNet \cite{syncnet}. LSE-D computes the distance between the predicted audio embedding and the predicted video embedding, while LSE-C represents the confidence of synchronization.
    
\end{itemize}

\subsection{Baselines} \label{sec:baseline_supp}
For non-diffusion-based methods, we compare with SadTalker \cite{sadtalker} and EDTalk \cite{edtalk}. For diffusion-based methods, we compare with AniTalker \cite{anitalker}, Hallo \cite{hallo}, and EchoMimic \cite{echomimic}.
\begin{itemize}
    \item \textbf{SadTalker} \cite{sadtalker} employs an audio-conditional variational auto-encoder (VAE) to synthesize the head motion and eye blink in a probabilistic way. 

    \item \textbf{EDTalk} \cite{edtalk} uses normalizing for audio-driven head motion generation and can separately control the lip and head motion.

    \item \textbf{AniTalker} \cite{anitalker} introduces a diffusion model to the learned motion latent space (similar to FLOAT) along with a variance adapter to improve the motion diversity. We use HuBERT audio feature-based implementation \cite{hubert} for improved lip synchronization and apply default guidance scales and denoising steps of the official implementation.

    \item \textbf{Hallo} \cite{hallo} uilizes the pre-trained StableDiffusion \cite{ldm} as its image generator, incorporating a hierarchical audio attention module to separately control lip synchronization, expression, and head pose. We use default guidance scales and denoising steps provided in the official implementation.

    \item \textbf{EchoMimic} \cite{echomimic} is also StableDiffusion-based method, which leverages facial skeleton as additional driving signals. We use the default guidance scales and denoising steps provided in the official implementation.

    \item It is worth noting that we compare with two superior works \textbf{EMO} \cite{emo} and \textbf{VASA-1} \cite{vasa_1} based on their demonstration videos due to their unavailable implementation. We highly recommend referring to \textit{`01$\_$EMO$\_$VASA-1$\_$Comparison$/$xxxx.mp4'}.
\end{itemize}

\subsection{More on Experiments} \label{sec:phase2_experiment_supp}
For evaluating our method, we use the first frame of each video clip as the source image. We use the first-order Euler method \cite{cfm} as our ODE solver. We experimentally find that other ODE solvers, such as mid-point and Dopri5, do not lead to significant performance improvements.

\begin{table}[!h]
    \centering
    \captionof{table}{Ablation studies of the different NFE of ODE on HDTF \cite{hdtf}. FPS is computed on a single NVIDIA V100 GPU.} \label{tab:ablation_nfe}
    \vspace*{-2mm}
    \resizebox{0.99\linewidth}{!}{
        \begin{tabular}{l | c c c c | c}
        \toprule
        Ours-NFE & \textbf{FID} $\downarrow$ & \textbf{FVD} $\downarrow$ & \textbf{E-FID} $\downarrow$ & \textbf{LSE-D} $\downarrow$ & \textbf{FPS} $\uparrow$\\
        \cline{2-6}
        \hline
        Ours-2                              & 21.785 & 178.831 & 1.542 & 7.559 & 45.22 \\
        Ours-5                              & 21.440 & 164.463 & 1.331 & 7.155 & 44.74 \\
        Ours-10 \textbf{(default)}          & 21.100 & 162.052 & 1.229 & 7.290 & 41.37 \\
        Ours-20                             & 21.158 & 164.392 & 1.293 & 7.343 & 38.20 \\
        \bottomrule
        \end{tabular}
        }
\end{table}

\noindent \textbf{Ablation on NFE~}~ In general, increasing the number of function evaluation (NFE) reduces the solution error of ODEs. As shown in \cref{tab:ablation_nfe}, even with small NFE $=2$, FLOAT can achieve competitive image quality (FID) and lip synchronization (LSE-D). However, it struggles to capture consistent and expressive motions (FVD and E-FID), resulting in shaky head motion and a static expression. This is because FLOAT generates the motion in the latent space, while image fidelity is determined by the auto-encoder. We provide supplementary videos, illustrating the impact of different NFE (Number of Function Evaluations). Notably, with a small NFE of 2, the generated images exhibit good quality, but the head movements appear temporally unstable, and emotions may be exaggerated. Please refer to supplementary videos for temporal jitters of low NFE.

\begin{table}[!h]
    \centering
    \captionof{table}{Ablation studies of the audio guidance scale $\gamma_a$ and the emotion guidance scale $\gamma_e$ on RAVDESS \cite{ravdess}.} \label{tab:ablation_guidances}
    \vspace*{-1mm}
    \resizebox{0.99\linewidth}{!}{
        \begin{tabular}{l | c c c c}
        \toprule
        Guidance scales  & \textbf{FID} $\downarrow$ & \textbf{FVD} $\downarrow$ & \textbf{E-FID} $\downarrow$ & \textbf{LSE-D} $\downarrow$\\
        \cline{2-5}
        \hline
        $\gamma_a$=1, ~~$\gamma_e$=1                       & 33.066 & 171.047 & 1.555 & 7.049 \\
        $\gamma_a$=1, ~~$\gamma_e$=2                       & 31.844 & 166.041 & 1.334 & 7.212 \\
        $\gamma_a$=2, ~~$\gamma_e$=1 ~\textbf{(default)}    & 31.681 & 166.359 & 1.367 & 6.994 \\
        $\gamma_a$=2, ~~$\gamma_e$=2                       & 32.253 & 162.658 & 1.351 & 6.994 \\
        \bottomrule
        \end{tabular}
        }
\end{table}

\noindent \textbf{Ablation on Guidance Scales~}~ In \cref{tab:ablation_guidances}, we conduct ablation studies on guidance scales: $\gamma_a$ and $\gamma_e$, with the emotion intensive dataset RAVDESS \cite{ravdess}. Note that increasing $\gamma_a$ leads to better temporal consistency (FVD) and lip synchronization quality (LSE-D). Moreover, increasing $\gamma_e$ improves video consistency (FVD) and expressiveness (E-FID). This enables balanced control over emotional audio-driven talking portrait generation. 

In \cref{fig:ecfg_scale_supp}, we visualize the effect of different emotion guidance scale $\gamma_e$. For this experiments, the predicted speech-to-emotion label is \textit{disgust} with $99\%$ probability. Notably, as increasing $\gamma_e$ from 0 to 2, we can observe that emotion-related expressions and motions are enhanced.

\begin{figure}[!h]
    \centering
    \includegraphics[width=0.99\linewidth]{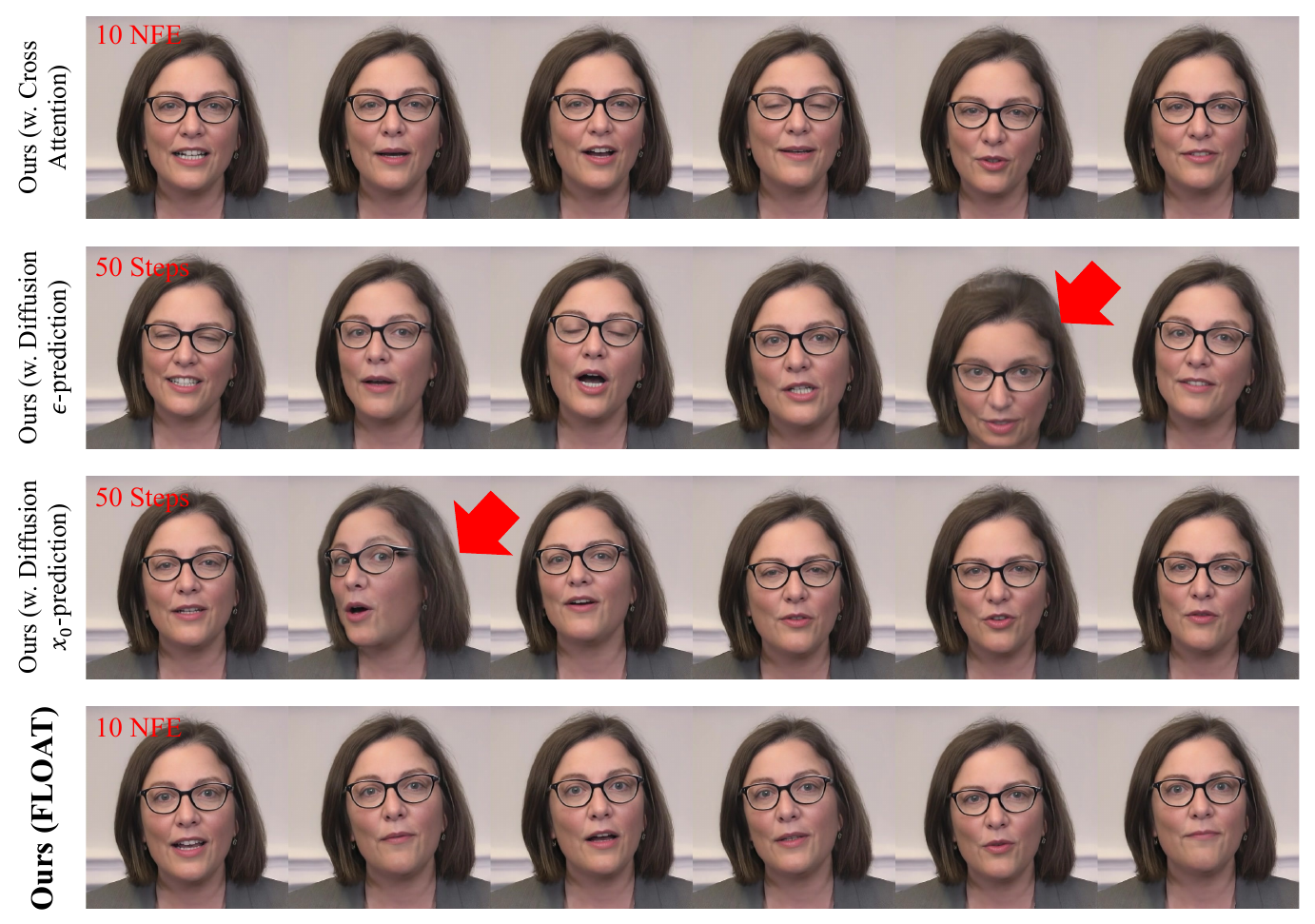}
    \vspace*{-2mm}
    \caption{Ablation results on frame-wise AdaLN and flow matching. Please refer to supplementary video for notable differences.}
    \label{fig:ablation_diffusion_supp2}
\end{figure}

\noindent \textbf{Ablation on AdaLN and Flow Matching}~ We conduct ablation study on frame-wise AdaLN by comparing it with a cross-attention. We adopt the stand cross-attention mechanism described in \cite{faceformer, diffposetalk}, using transformer encoder architecture for non-autoregressive sequence modeling. We use the same attention mask used in the frame-wise AdaLN, which attends to additional $2T$ adjacent frames for the $l$-th input latent: $[l-2, l-1, l, l+1, l+2]$. 

To compare against flow matching, we implement two diffusion models with distinct parameterizations: $\epsilon$-prediction and $x_0$-prediction. For $\epsilon$-prediction, we directly predict Gaussian noise by the noise predictor $s( \cdot ; \theta)$ parameterized by $\theta$ with the following simple loss:
\begin{align}
    \loss_{\text{simple, noise}}(\theta) = \| s(x_t, \mathbf{c}_t ; \theta) - \epsilon \|_2^2,
\end{align}
where $t$ $\sim$ $\mathcal{U}[0, 1]$, $\epsilon$ $\sim$ $\mathcal{N} (0^{-L':L}, I)$, and the noise input $x_t$ $\in$ $\real^{(L' + L) \times d}$ is sampled from a forward diffusion process $q(x_t |x_{t-1}) = \mathcal{N}(x_t; \sqrt{1-\beta_t} x_{t-1}, \beta_t I)$ \cite{ddpm}. In our case, $x_t$ is noisy motion latents at diffusion time step $t$, starting from $t=0$ with $x_0$ $=$ $w_{r \to D^{1:L}}$ $\in$ $\real^{(-L'+L) \times d}$.

For $x_0$-prediction, we predict a clean sample $x_0$, instead of noise \cite{dalle2}, by the predictor $s(\cdot; \theta)$ with the following simple loss:
\begin{align}
    \loss_{\text{simple}, x_0}(\theta) = \| s(x_t, \mathbf{c}_t ; \theta) - x_0 \|_2^2.
\end{align}
We also incorporate a velocity loss \cite{mdm}:
\begin{align}
    \loss_{\text{vel}, x_0}(\theta) = \| \Delta s - \Delta x_0 \|_2^2,
\end{align}
where $\Delta s$ and $\Delta x_0$ are the one-frame difference along the time-axis for $s$ and $x_0$, respectively. The total loss $\loss_{\text{total}, x_0}(\theta)$ is
\begin{align}
    \loss_{\text{total}, x_0}(\theta) = \loss_{\text{simple}, x_0}(\theta) + \loss_{\text{vel}, x_0}(\theta).
\end{align}
For reverse process, we use the DDIM \cite{ddim} sampler with 50 denoising steps.

In our implementation, both $\epsilon$-prediction and $x_0$-prediction achieve the best results with guidance scales $\gamma_a = \gamma_e = 1$ (default). In \cref{fig:ablation_diffusion_supp2}, \cref{fig:ablation_diffusion_supp1} and \cref{fig:ablation_diffusion_supp3}, we provide qualitative comparisons between these approaches and FLOAT. Notably, the cross-attention exhibits less diverse head motions compared to FLOAT, while diffusion-based approaches struggle to generate temporally stable lip and head motion, often resulting in out-of-sync movements or motion artifacts. 

\section{Additional Results} \label{sec:phase2_additional_result_supp}

\subsection{Additional Comparison Results} \label{sec:additional_comparison_supp}
We provide additional comparison results with baselines in \cref{fig:sota_compare_supp1}, \cref{fig:sota_compare_supp2}, and \cref{fig:sota_compare_supp3}. 

\subsection{Out-of-distribution (OOD) Results} \label{sec:ood}
In \cref{fig:ood_supp} and \cref{fig:ecfg_scale_supp}, we present additional out-of-distribution results, including paintings, non-English speech, and singing.

\subsection{User Study} \label{sec:user}
\begin{table}[h]
    \centering
    \captionof{table}{Mean opinion score (MOS) study results with $95\%$ confidence interval. The score ranges in 1 to 5. The best result for each metric is in \textbf{bold}.} \label{tab:user_study}
    \vspace*{-2mm}
    \resizebox{0.99\linewidth}{!}{
        \begin{tabular}{l | c c c c c}
        \toprule
        Method & \makecell{\textbf{Lip Sync} \\ \textbf{Accuracy}} & \makecell{\textbf{Natural} \\ \textbf{Head Motion}} & \makecell{\textbf{Teeth} \\ \textbf{Clarity}} & \makecell{\textbf{Natural} \\ \textbf{Emotion}} & \makecell{ \textbf{Overall} \\ \textbf{Visual Quality}} \\
        \cline{2-6}
        \hline        
        SadTalker \cite{sadtalker} & 2.20 $\pm$ 0.35 & 2.03 $\pm$ 0.26 & 1.53 $\pm$ 0.19 & 1.80 $\pm$ 0.28 & 1.97 $\pm$ 0.23 \\
        EdTalk \cite{edtalk} & 2.50 $\pm$ 0.34 & 2.60 $\pm$ 0.28 & 1.17 $\pm$ 0.17 & 2.07 $\pm$ 0.36 & 1.83 $\pm$ 0.27 \\
        \hline
        AniTalker \cite{anitalker} & 2.70 $\pm$ 0.31 & 3.00 $\pm$ 0.30 & 2.13 $\pm$ 0.27 & 3.17 $\pm$ 0.27 & 2.63 $\pm$ 0.26 \\
        Hallo \cite{hallo} & 3.30 $\pm$ 0.32 & 2.73 $\pm$ 0.35 & 2.23 $\pm$ 0.27 & 2.67 $\pm$ 0.35 & 2.27 $\pm$ 0.33 \\
        EchoMimic \cite{echomimic} & 2.67 $\pm$ 0.37 & 3.07 $\pm$ 0.30 & 2.20 $\pm$ 0.34 & 2.50 $\pm$ 0.37 & 2.70 $\pm$ 0.36 \\
        \hline
        \textbf{FLOAT (Ours)}  & \textbf{3.93 $\pm$ 0.21} & \textbf{3.57 $\pm$ 0.33} & \textbf{4.13 $\pm$ 0.27} & \textbf{3.77 $\pm$ 0.30} & \textbf{3.87 $\pm$ 0.30} \\
        \bottomrule
        \end{tabular}
        }
\end{table}

In \cref{tab:user_study}, we conduct a mean opinion score (MOS) based user study to compare the perceptual quality of each method (\eg, teeth clarity and naturalness of emotion). We generate 6 videos by using the baselines and FLOAT, and ask 15 participants to evaluate each generated video with five evaluation factors in the range of 1 to 5. As shown in \cref{tab:user_study}, FLOAT outperforms the baselines.

\begin{figure}[!h]
    \centering
    \includegraphics[width=\linewidth]{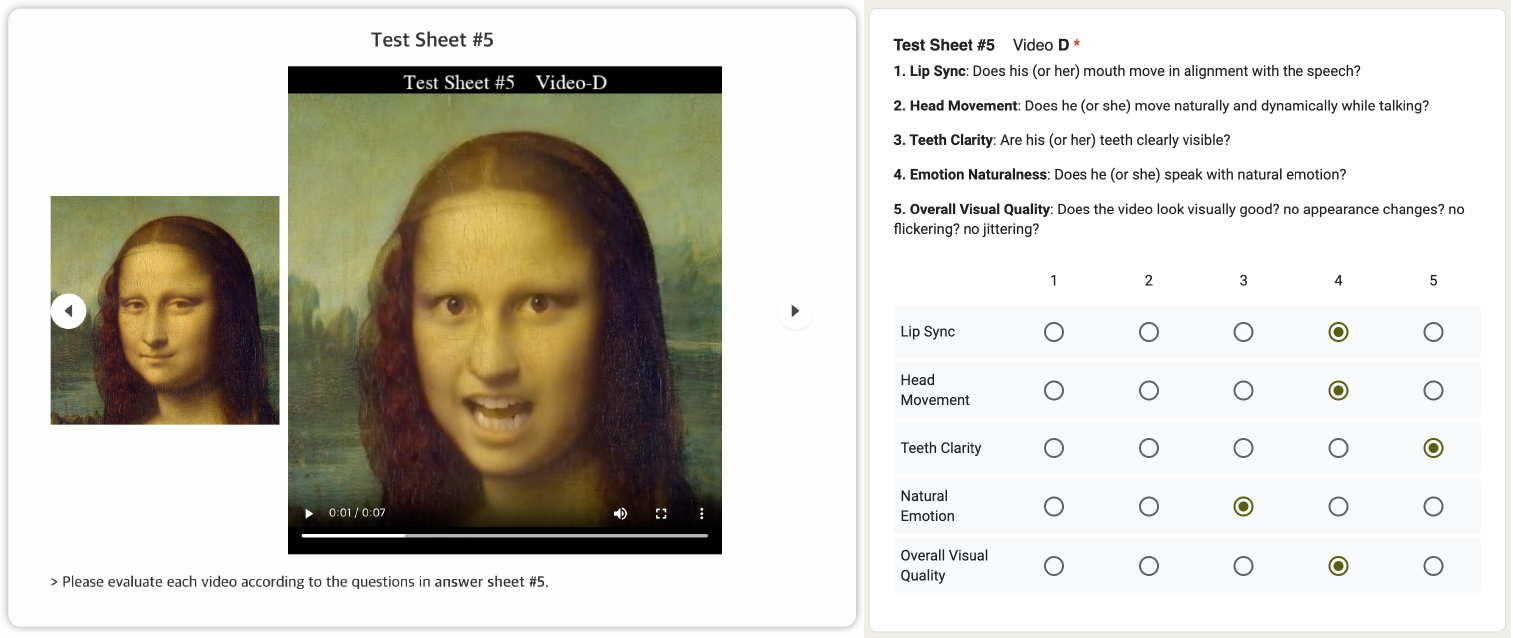}
    \vspace*{-4mm}
    \caption{Example of user study interface. (Left) Test Sheet; (Right) Answer Sheet. Participants were asked to evaluate 5 questions for each video (total 180 videos).} \label{fig:user_study}

\end{figure}

In \cref{fig:user_study}, we provide an example of test and answer sheet used of the user study. We asked 15 participants to evaluate five questions for each generated video produced by the baselines and FLOAT. Consequently, each participant scores total 180 questions, with responses ranged from 1 to 5. Additionally, we include the supplementary videos used in the user study.

\subsection{Video Results}
We include video results to further illustrate the performance of our method, including emotion redirection, additional driving conditions, and OOD results. Please refer to provided videos.

\section{Discussion} \label{sec:discussion}
\noindent \textbf{Ethical Consideration~}~ This work aims to advance virtual avatar generation. However, as it can generate realistic talking portrait only from a single image and audio, we considerably recognize the potential for misuse, such as deepfake creation. Attaching watermarks to generated videos and carefully restricted license can mitigate this issues. Additionally, we encourage researchers in deepfake detection to use our results as data to improve detection tools.

\noindent \textbf{Limitation and Further Work~}~ While our method can generate realistic talking portrait video from a single source image and a driving audio, it has several limitations.

First, our method cannot generate more vivid and naunced emotional talking motion. This is because the speech-driven emotion labels are restricted to seven basic emotions, making it challenging to capture more nuanced emotions like \textit{shyness}. We believe this limitation can be addressed by incorporating textual cues (\eg, \textit{``gazing forward with a shyness"}), an idea we plan to explore in future work. Moreover, any other approaches to enhance the naturalness of talking motion are key directions for our future work. 

\begin{figure}[t]
    \centering
    \includegraphics[width=0.999\linewidth]{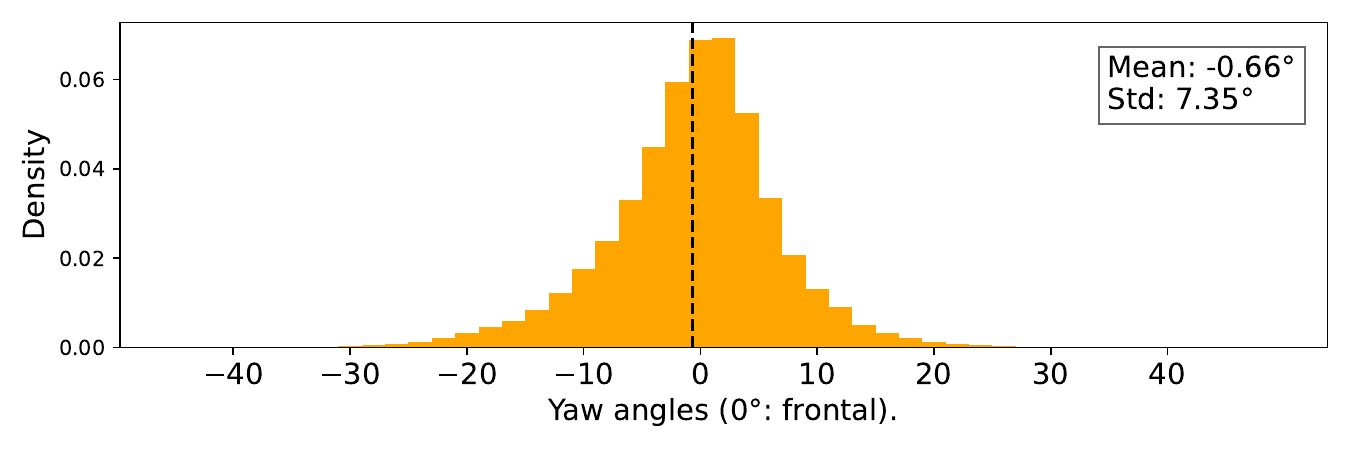}
    \vspace*{-9mm}
    \caption{Distribution yaw angles in training dataset \cite{hdtf, ravdess} for FLOAT. }\label{fig:yaw}
\end{figure}

\begin{figure}[t]
    \centering
    \vspace*{-1mm}
    \includegraphics[width=0.9\linewidth]{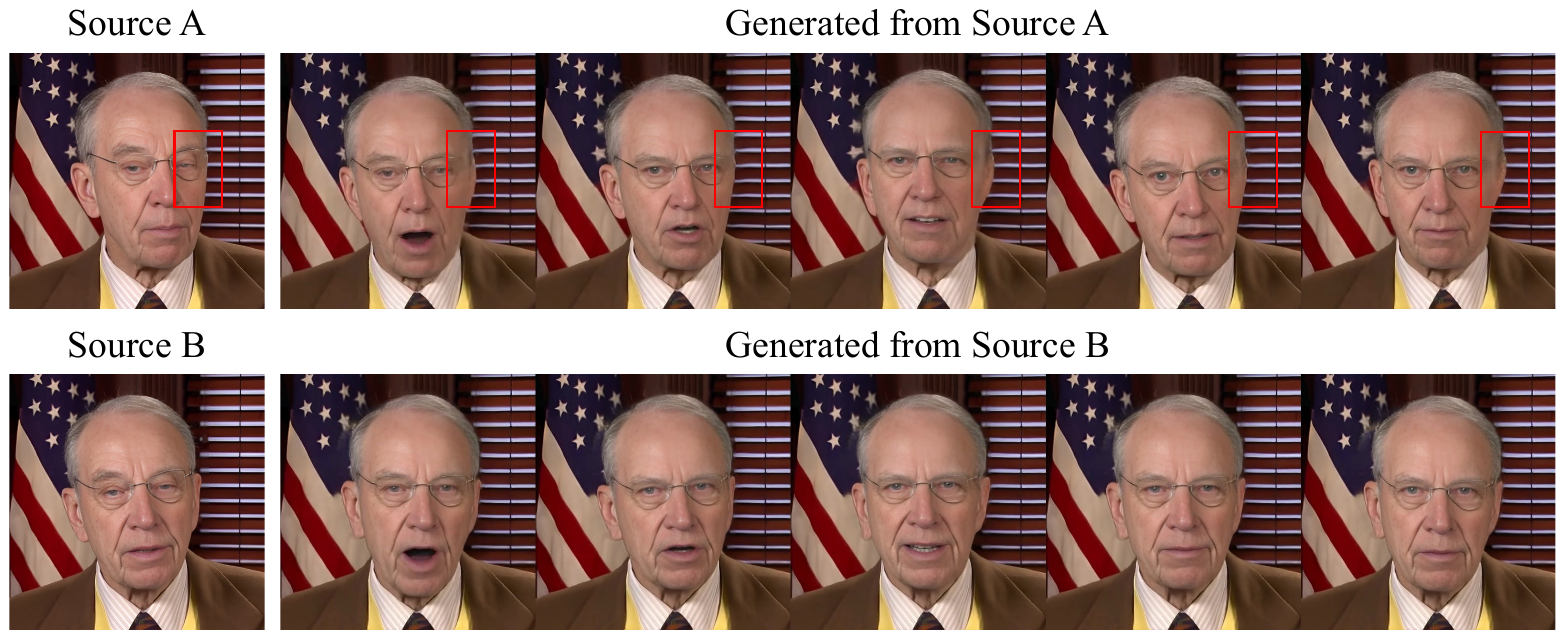}
    \vspace*{-2mm}
    \caption{Failure case of FLOAT. It often struggles to handle non-frontal faces and accessories, such as glasses. Please refer to supplementary video.}
    \label{fig:failure_case_supp1}
\end{figure}

Second, we aim to build our method solely upon high-definition open-source datasets. Since the training datasets are biased toward frontal head angles \cite{hdtf, ravdess}, the generated results also exhibit a similar bias, often producing suboptimal results for non-frontal (\eg, $ |\text{yaw angle}| \geq \ang{20}$) source images or images with notable accessories. This is partially because the head pose distribution of our training data as shown in \cref{fig:yaw}. Although we investigated other existing high-definite face video datasets, such as MEAD \cite{mead} and CelebV-Text \cite{celebv_text}, we found limitations in their suitability. MEAD \cite{mead} contains minimal head motion and a limited number of identities, while CelebV-Text \cite{celebv_text} is not organized for audio-driven talking portrait, containing out-of-sync audio and significant background inconsistencies.
 
 This limitations can be mitigated by introducing carefully curated external data, as demonstrated by other concurrent methods \cite{emo, hallo, loopy, vasa_1, gaia}, or by incorporating multi-view supervision \cite{live3dportrait} when training our motion latent auto-encoder. We provide examples of failure case in \cref{fig:failure_case_supp1} and supplementary video.

\noindent \textbf{Acknowledgment~}~ The source images and audio used in this paper are taken from other talking portrait generation methods \cite{emo, vasa_1, hallo, echomimic, sadtalker}. We sincerely thank the authors of these works for their valuable contributions. Note that the individuals depicted in our source images and the speech generated in our experiments are not associated with the actual persons they represent.



\clearpage

%% file: main.bbl
\begin{thebibliography}{101}
\providecommand{\natexlab}[1]{#1}
\providecommand{\url}[1]{\texttt{#1}}
\expandafter\ifx\csname urlstyle\endcsname\relax
  \providecommand{\doi}[1]{doi: #1}\else
  \providecommand{\doi}{doi: \begingroup \urlstyle{rm}\Url}\fi

\bibitem[Baevski et~al.(2020)Baevski, Zhou, Mohamed, and Auli]{wav2vec2}
Alexei Baevski, Yuhao Zhou, Abdelrahman Mohamed, and Michael Auli.
\newblock wav2vec 2.0: A framework for self-supervised learning of speech representations.
\newblock \emph{Advances in neural information processing systems}, 33:\penalty0 12449--12460, 2020.

\bibitem[Blanz and Vetter(1999)]{3dmm}
Volker Blanz and Thomas Vetter.
\newblock A morphable model for the synthesis of 3d faces.
\newblock In \emph{Proceedings of the 26th annual conference on Computer graphics and interactive techniques}, pages 187--194, 1999.

\bibitem[Brooks et~al.(2023)Brooks, Holynski, and Efros]{instructpix2pix}
Tim Brooks, Aleksander Holynski, and Alexei~A Efros.
\newblock Instructpix2pix: Learning to follow image editing instructions.
\newblock In \emph{Proceedings of the IEEE/CVF Conference on Computer Vision and Pattern Recognition (CVPR)}, pages 18392--18402, 2023.

\bibitem[Brooks et~al.(2024)Brooks, Peebles, Holmes, DePue, Guo, Jing, Schnurr, Taylor, Luhman, Luhman, et~al.]{sora}
Tim Brooks, Bill Peebles, Connor Holmes, Will DePue, Yufei Guo, Li Jing, David Schnurr, Joe Taylor, Troy Luhman, Eric Luhman, et~al.
\newblock Video generation models as world simulators, 2024.

\bibitem[Bulat and Tzimiropoulos(2017)]{face_alignment}
Adrian Bulat and Georgios Tzimiropoulos.
\newblock How far are we from solving the 2d \& 3d face alignment problem? (and a dataset of 230,000 3d facial landmarks).
\newblock In \emph{International Conference on Computer Vision}, 2017.

\bibitem[Carreira and Zisserman(2017)]{i3d}
Joao Carreira and Andrew Zisserman.
\newblock Quo vadis, action recognition? a new model and the kinetics dataset.
\newblock In \emph{proceedings of the IEEE Conference on Computer Vision and Pattern Recognition}, pages 6299--6308, 2017.

\bibitem[Chen et~al.(2018)Chen, Rubanova, Bettencourt, and Duvenaud]{node}
Ricky~TQ Chen, Yulia Rubanova, Jesse Bettencourt, and David~K Duvenaud.
\newblock Neural ordinary differential equations.
\newblock \emph{Advances in neural information processing systems}, 31, 2018.

\bibitem[Chen et~al.(2024)Chen, Cao, Chen, Li, and Ma]{echomimic}
Zhiyuan Chen, Jiajiong Cao, Zhiquan Chen, Yuming Li, and Chenguang Ma.
\newblock Echomimic: Lifelike audio-driven portrait animations through editable landmark conditions.
\newblock \emph{arXiv preprint arXiv:2407.08136}, 2024.

\bibitem[Cheng et~al.(2022)Cheng, Cun, Zhang, Xia, Yin, Zhu, Wang, Wang, and Wang]{videoretalking}
Kun Cheng, Xiaodong Cun, Yong Zhang, Menghan Xia, Fei Yin, Mingrui Zhu, Xuan Wang, Jue Wang, and Nannan Wang.
\newblock Videoretalking: Audio-based lip synchronization for talking head video editing in the wild.
\newblock In \emph{SIGGRAPH Asia 2022 Conference Papers}, pages 1--9, 2022.

\bibitem[Chung and Zisserman(2016)]{syncnet}
Joon~Son Chung and Andrew Zisserman.
\newblock Out of time: automated lip sync in the wild.
\newblock In \emph{Asian Conference on Computer Vision}, pages 251--263, 2016.

\bibitem[Dao et~al.(2023)Dao, Phung, Nguyen, and Tran]{lcfm}
Quan Dao, Hao Phung, Binh Nguyen, and Anh Tran.
\newblock Flow matching in latent space.
\newblock \emph{arXiv preprint arXiv:2307.08698}, 2023.

\bibitem[Deng et~al.(2019{\natexlab{a}})Deng, Guo, Xue, and Zafeiriou]{arcface}
Jiankang Deng, Jia Guo, Niannan Xue, and Stefanos Zafeiriou.
\newblock Arcface: Additive angular margin loss for deep face recognition.
\newblock In \emph{Proceedings of the IEEE/CVF Conference on Computer Vision and Pattern Recognition (CVPR)}, pages 4690--4699, 2019{\natexlab{a}}.

\bibitem[Deng et~al.(2019{\natexlab{b}})Deng, Yang, Xu, Chen, Jia, and Tong]{bfm}
Yu Deng, Jiaolong Yang, Sicheng Xu, Dong Chen, Yunde Jia, and Xin Tong.
\newblock Accurate 3d face reconstruction with weakly-supervised learning: From single image to image set.
\newblock In \emph{Proceedings of the IEEE/CVF conference on computer vision and pattern recognition workshops}, pages 0--0, 2019{\natexlab{b}}.

\bibitem[Dhariwal and Nichol(2021)]{beats}
Prafulla Dhariwal and Alexander Nichol.
\newblock Diffusion models beat gans on image synthesis.
\newblock \emph{Advances in neural information processing systems}, 34:\penalty0 8780--8794, 2021.

\bibitem[Dinh et~al.(2016)Dinh, Sohl-Dickstein, and Bengio]{realnvp}
Laurent Dinh, Jascha Sohl-Dickstein, and Samy Bengio.
\newblock Density estimation using real nvp.
\newblock \emph{arXiv preprint arXiv:1605.08803}, 2016.

\bibitem[Drobyshev et~al.(2022)Drobyshev, Chelishev, Khakhulin, Ivakhnenko, Lempitsky, and Zakharov]{megaportrait}
Nikita Drobyshev, Jenya Chelishev, Taras Khakhulin, Aleksei Ivakhnenko, Victor Lempitsky, and Egor Zakharov.
\newblock Megaportraits: One-shot megapixel neural head avatars.
\newblock In \emph{Proceedings of the 30th ACM International Conference on Multimedia}, pages 2663--2671, 2022.

\bibitem[Drobyshev et~al.(2024)Drobyshev, Casademunt, Vougioukas, Landgraf, Petridis, and Pantic]{emoportraits}
Nikita Drobyshev, Antoni~Bigata Casademunt, Konstantinos Vougioukas, Zoe Landgraf, Stavros Petridis, and Maja Pantic.
\newblock Emoportraits: Emotion-enhanced multimodal one-shot head avatars.
\newblock In \emph{Proceedings of the IEEE/CVF Conference on Computer Vision and Pattern Recognition}, pages 8498--8507, 2024.

\bibitem[Esser et~al.(2024)Esser, Kulal, Blattmann, Entezari, M{\"u}ller, Saini, Levi, Lorenz, Sauer, Boesel, et~al.]{sd3}
Patrick Esser, Sumith Kulal, Andreas Blattmann, Rahim Entezari, Jonas M{\"u}ller, Harry Saini, Yam Levi, Dominik Lorenz, Axel Sauer, Frederic Boesel, et~al.
\newblock Scaling rectified flow transformers for high-resolution image synthesis.
\newblock In \emph{Forty-first International Conference on Machine Learning}, 2024.

\bibitem[Fan et~al.(2022)Fan, Lin, Saito, Wang, and Komura]{faceformer}
Yingruo Fan, Zhaojiang Lin, Jun Saito, Wenping Wang, and Taku Komura.
\newblock Faceformer: Speech-driven 3d facial animation with transformers.
\newblock In \emph{Proceedings of the IEEE/CVF Conference on Computer Vision and Pattern Recognition (CVPR)}, pages 18770--18780, 2022.

\bibitem[Fischer et~al.(2023)Fischer, Gui, Ma, Stracke, Baumann, and Ommer]{boosting_flow_matching}
Johannes~S Fischer, Ming Gui, Pingchuan Ma, Nick Stracke, Stefan~A Baumann, and Bj{\"o}rn Ommer.
\newblock Boosting latent diffusion with flow matching.
\newblock \emph{arXiv preprint arXiv:2312.07360}, 2023.

\bibitem[Gatys et~al.(2016)Gatys, Ecker, and Bethge]{gram_matrix}
Leon~A. Gatys, Alexander~S. Ecker, and Matthias Bethge.
\newblock Image style transfer using convolutional neural networks.
\newblock In \emph{2016 IEEE Conference on Computer Vision and Pattern Recognition (CVPR)}, pages 2414--2423, 2016.

\bibitem[Goodfellow et~al.(2014)Goodfellow, Pouget-Abadie, Mirza, Xu, Warde-Farley, Ozair, Courville, and Bengio]{gan}
Ian Goodfellow, Jean Pouget-Abadie, Mehdi Mirza, Bing Xu, David Warde-Farley, Sherjil Ozair, Aaron Courville, and Yoshua Bengio.
\newblock Generative adversarial nets.
\newblock \emph{Advances in neural information processing systems}, 27, 2014.

\bibitem[Guan et~al.(2023)Guan, Zhang, Zhou, Hu, Wang, He, Feng, Liu, Ding, Liu, et~al.]{stylesync}
Jiazhi Guan, Zhanwang Zhang, Hang Zhou, Tianshu Hu, Kaisiyuan Wang, Dongliang He, Haocheng Feng, Jingtuo Liu, Errui Ding, Ziwei Liu, et~al.
\newblock Stylesync: High-fidelity generalized and personalized lip sync in style-based generator.
\newblock In \emph{Proceedings of the IEEE/CVF Conference on Computer Vision and Pattern Recognition (CVPR)}, pages 1505--1515, 2023.

\bibitem[Guo et~al.(2023)Guo, Yang, Rao, Liang, Wang, Qiao, Agrawala, Lin, and Dai]{animatediff}
Yuwei Guo, Ceyuan Yang, Anyi Rao, Zhengyang Liang, Yaohui Wang, Yu Qiao, Maneesh Agrawala, Dahua Lin, and Bo Dai.
\newblock Animatediff: Animate your personalized text-to-image diffusion models without specific tuning.
\newblock \emph{arXiv preprint arXiv:2307.04725}, 2023.

\bibitem[He et~al.(2023)He, Guo, Yu, Wang, Zhu, An, Li, Tan, Wang, Hu, et~al.]{gaia}
Tianyu He, Junliang Guo, Runyi Yu, Yuchi Wang, Jialiang Zhu, Kaikai An, Leyi Li, Xu Tan, Chunyu Wang, Han Hu, et~al.
\newblock Gaia: Zero-shot talking avatar generation.
\newblock \emph{arXiv preprint arXiv:2311.15230}, 2023.

\bibitem[Hinton(2015)]{distillation}
Geoffrey Hinton.
\newblock Distilling the knowledge in a neural network.
\newblock \emph{arXiv preprint arXiv:1503.02531}, 2015.

\bibitem[Ho et~al.(2020)Ho, Jain, and Abbeel]{ddpm}
Jonathan Ho, Ajay Jain, and Pieter Abbeel.
\newblock Denoising diffusion probabilistic models.
\newblock \emph{Advances in neural information processing systems}, 33:\penalty0 6840--6851, 2020.

\bibitem[Hsu et~al.(2021)Hsu, Bolte, Tsai, Lakhotia, Salakhutdinov, and Mohamed]{hubert}
Wei-Ning Hsu, Benjamin Bolte, Yao-Hung~Hubert Tsai, Kushal Lakhotia, Ruslan Salakhutdinov, and Abdelrahman Mohamed.
\newblock Hubert: Self-supervised speech representation learning by masked prediction of hidden units.
\newblock \emph{IEEE/ACM transactions on audio, speech, and language processing}, 29:\penalty0 3451--3460, 2021.

\bibitem[Hu(2024)]{animate_anyone}
Li Hu.
\newblock Animate anyone: Consistent and controllable image-to-video synthesis for character animation.
\newblock In \emph{Proceedings of the IEEE/CVF Conference on Computer Vision and Pattern Recognition (CVPR)}, pages 8153--8163, 2024.

\bibitem[Ji et~al.(2022)Ji, Zhou, Wang, Wu, Wu, Xu, and Cao]{eamm}
Xinya Ji, Hang Zhou, Kaisiyuan Wang, Qianyi Wu, Wayne Wu, Feng Xu, and Xun Cao.
\newblock Eamm: One-shot emotional talking face via audio-based emotion-aware motion model.
\newblock In \emph{ACM SIGGRAPH 2022 Conference Proceedings}, pages 1--10, 2022.

\bibitem[Jiang et~al.(2024)Jiang, Liang, Yang, Lin, Zhong, and Zheng]{loopy}
Jianwen Jiang, Chao Liang, Jiaqi Yang, Gaojie Lin, Tianyun Zhong, and Yanbo Zheng.
\newblock Loopy: Taming audio-driven portrait avatar with long-term motion dependency.
\newblock \emph{arXiv preprint arXiv:2409.02634}, 2024.

\bibitem[Kang et~al.(2024)Kang, Zhang, Barnes, Paris, Kwak, Park, Shechtman, Zhu, and Park]{diffusion2gan}
Minguk Kang, Richard Zhang, Connelly Barnes, Sylvain Paris, Suha Kwak, Jaesik Park, Eli Shechtman, Jun-Yan Zhu, and Taesung Park.
\newblock Distilling diffusion models into conditional gans.
\newblock \emph{arXiv preprint arXiv:2405.05967}, 2024.

\bibitem[Karras et~al.(2020)Karras, Laine, Aittala, Hellsten, Lehtinen, and Aila]{stylegan2}
Tero Karras, Samuli Laine, Miika Aittala, Janne Hellsten, Jaakko Lehtinen, and Timo Aila.
\newblock Analyzing and improving the image quality of stylegan.
\newblock In \emph{Proceedings of the IEEE/CVF Conference on Computer Vision and Pattern Recognition (CVPR)}, pages 8110--8119, 2020.

\bibitem[Ki and Min(2023)]{stylelipsync}
Taekyung Ki and Dongchan Min.
\newblock Stylelipsync: Style-based personalized lip-sync video generation.
\newblock In \emph{Proceedings of the IEEE/CVF International Conference on Computer Vision}, pages 22841--22850, 2023.

\bibitem[Kingma(2013)]{vae}
Diederik~P Kingma.
\newblock Auto-encoding variational bayes.
\newblock \emph{arXiv preprint arXiv:1312.6114}, 2013.

\bibitem[Kingma(2014)]{adam}
Diederik~P Kingma.
\newblock Adam: A method for stochastic optimization.
\newblock \emph{arXiv preprint arXiv:1412.6980}, 2014.

\bibitem[Kirschstein et~al.(2024)Kirschstein, Giebenhain, and Nie{\ss}ner]{diffusionavatars}
Tobias Kirschstein, Simon Giebenhain, and Matthias Nie{\ss}ner.
\newblock Diffusionavatars: Deferred diffusion for high-fidelity 3d head avatars.
\newblock In \emph{Proceedings of the IEEE/CVF Conference on Computer Vision and Pattern Recognition (CVPR)}, pages 5481--5492, 2024.

\bibitem[Krizhevsky et~al.(2012)Krizhevsky, Sutskever, and Hinton]{alexnet}
Alex Krizhevsky, Ilya Sutskever, and Geoffrey~E Hinton.
\newblock Imagenet classification with deep convolutional neural networks.
\newblock \emph{Advances in neural information processing systems}, 25, 2012.

\bibitem[Le et~al.(2024)Le, Vyas, Shi, Karrer, Sari, Moritz, Williamson, Manohar, Adi, Mahadeokar, et~al.]{voicebox}
Matthew Le, Apoorv Vyas, Bowen Shi, Brian Karrer, Leda Sari, Rashel Moritz, Mary Williamson, Vimal Manohar, Yossi Adi, Jay Mahadeokar, et~al.
\newblock Voicebox: Text-guided multilingual universal speech generation at scale.
\newblock \emph{Advances in neural information processing systems}, 36, 2024.

\bibitem[Lei~Ba et~al.(2016)Lei~Ba, Kiros, and Hinton]{layernorm}
Jimmy Lei~Ba, Jamie~Ryan Kiros, and Geoffrey~E Hinton.
\newblock Layer normalization.
\newblock \emph{ArXiv e-prints}, pages arXiv--1607, 2016.

\bibitem[Li et~al.(2023)Li, Hu, Khan, Li, Yang, Wang, Cheng, and Yang]{fasterdiffusion}
Senmao Li, Taihang Hu, Fahad~Shahbaz Khan, Linxuan Li, Shiqi Yang, Yaxing Wang, Ming-Ming Cheng, and Jian Yang.
\newblock Faster diffusion: Rethinking the role of unet encoder in diffusion models.
\newblock \emph{arXiv preprint arXiv:2312.09608}, 2023.

\bibitem[Lipman et~al.(2022)Lipman, Chen, Ben-Hamu, Nickel, and Le]{cfm}
Yaron Lipman, Ricky~TQ Chen, Heli Ben-Hamu, Maximilian Nickel, and Matt Le.
\newblock Flow matching for generative modeling.
\newblock \emph{arXiv preprint arXiv:2210.02747}, 2022.

\bibitem[Liu et~al.(2024)Liu, Chen, Fan, Du, Chen, Chen, and Yu]{anitalker}
Tao Liu, Feilong Chen, Shuai Fan, Chenpeng Du, Qi Chen, Xie Chen, and Kai Yu.
\newblock Anitalker: Animate vivid and diverse talking faces through identity-decoupled facial motion encoding.
\newblock \emph{arXiv preprint arXiv:2405.03121}, 2024.

\bibitem[Liu et~al.(2022)Liu, Gong, and Liu]{rectflow}
Xingchao Liu, Chengyue Gong, and Qiang Liu.
\newblock Flow straight and fast: Learning to generate and transfer data with rectified flow.
\newblock \emph{arXiv preprint arXiv:2209.03003}, 2022.

\bibitem[Liu et~al.(2023)Liu, Zhang, Ma, Peng, et~al.]{instaflow}
Xingchao Liu, Xiwen Zhang, Jianzhu Ma, Jian Peng, et~al.
\newblock Instaflow: One step is enough for high-quality diffusion-based text-to-image generation.
\newblock In \emph{The Twelfth International Conference on Learning Representations}, 2023.

\bibitem[Livingstone and Russo(2018)]{ravdess}
Steven~R Livingstone and Frank~A Russo.
\newblock The ryerson audio-visual database of emotional speech and song (ravdess): A dynamic, multimodal set of facial and vocal expressions in north american english.
\newblock \emph{PloS one}, 13\penalty0 (5):\penalty0 e0196391, 2018.

\bibitem[Lu et~al.(2022{\natexlab{a}})Lu, Zhou, Bao, Chen, Li, and Zhu]{dpm_solver}
Cheng Lu, Yuhao Zhou, Fan Bao, Jianfei Chen, Chongxuan Li, and Jun Zhu.
\newblock Dpm-solver: A fast ode solver for diffusion probabilistic model sampling in around 10 steps.
\newblock \emph{Advances in Neural Information Processing Systems}, 35:\penalty0 5775--5787, 2022{\natexlab{a}}.

\bibitem[Lu et~al.(2022{\natexlab{b}})Lu, Zhou, Bao, Chen, Li, and Zhu]{dpm_solver_pp}
Cheng Lu, Yuhao Zhou, Fan Bao, Jianfei Chen, Chongxuan Li, and Jun Zhu.
\newblock Dpm-solver++: Fast solver for guided sampling of diffusion probabilistic models.
\newblock \emph{arXiv preprint arXiv:2211.01095}, 2022{\natexlab{b}}.

\bibitem[Luo et~al.(2023)Luo, Tan, Huang, Li, and Zhao]{latent_consistency_model}
Simian Luo, Yiqin Tan, Longbo Huang, Jian Li, and Hang Zhao.
\newblock Latent consistency models: Synthesizing high-resolution images with few-step inference.
\newblock \emph{arXiv preprint arXiv:2310.04378}, 2023.

\bibitem[Ma et~al.(2023{\natexlab{a}})Ma, Wang, Hu, Fan, Lv, Ding, Deng, and Yu]{styletalk}
Yifeng Ma, Suzhen Wang, Zhipeng Hu, Changjie Fan, Tangjie Lv, Yu Ding, Zhidong Deng, and Xin Yu.
\newblock Styletalk: One-shot talking head generation with controllable speaking styles.
\newblock In \emph{Proceedings of the AAAI Conference on Artificial Intelligence}, pages 1896--1904, 2023{\natexlab{a}}.

\bibitem[Ma et~al.(2023{\natexlab{b}})Ma, Zhang, Wang, Wang, Zhang, and Deng]{dreamtalk}
Yifeng Ma, Shiwei Zhang, Jiayu Wang, Xiang Wang, Yingya Zhang, and Zhidong Deng.
\newblock Dreamtalk: When expressive talking head generation meets diffusion probabilistic models.
\newblock \emph{arXiv preprint arXiv:2312.09767}, 2023{\natexlab{b}}.

\bibitem[Min et~al.(2022)Min, Song, Ko, and Hwang]{styletalker}
Dongchan Min, Minyoung Song, Eunji Ko, and Sung~Ju Hwang.
\newblock Styletalker: One-shot style-based audio-driven talking head video generation.
\newblock \emph{arXiv preprint arXiv:2208.10922}, 2022.

\bibitem[Nichol and Dhariwal(2021)]{iddpm}
Alexander~Quinn Nichol and Prafulla Dhariwal.
\newblock Improved denoising diffusion probabilistic models.
\newblock In \emph{International conference on machine learning}, pages 8162--8171. PMLR, 2021.

\bibitem[Park et~al.(2022)Park, Kim, Hong, Choi, and Ro]{synctalkface}
Se~Jin Park, Minsu Kim, Joanna Hong, Jeongsoo Choi, and Yong~Man Ro.
\newblock Synctalkface: Talking face generation with precise lip-syncing via audio-lip memory.
\newblock In \emph{Proceedings of the AAAI Conference on Artificial Intelligence}, pages 2062--2070, 2022.

\bibitem[Peebles and Xie(2023)]{dit}
William Peebles and Saining Xie.
\newblock Scalable diffusion models with transformers.
\newblock In \emph{Proceedings of the IEEE/CVF International Conference on Computer Vision}, pages 4195--4205, 2023.

\bibitem[Pepino et~al.(2021)Pepino, Riera, and Ferrer]{speech2emotion}
Leonardo Pepino, Pablo Riera, and Luciana Ferrer.
\newblock Emotion recognition from speech using wav2vec 2.0 embeddings.
\newblock \emph{arXiv preprint arXiv:2104.03502}, 2021.

\bibitem[Polyak et~al.(2024)Polyak, Zohar, Brown, Tjandra, Sinha, Lee, Vyas, Shi, Ma, Chuang, et~al.]{moviegen}
Adam Polyak, Amit Zohar, Andrew Brown, Andros Tjandra, Animesh Sinha, Ann Lee, Apoorv Vyas, Bowen Shi, Chih-Yao Ma, Ching-Yao Chuang, et~al.
\newblock Movie gen: A cast of media foundation models.
\newblock \emph{arXiv preprint arXiv:2410.13720}, 2024.

\bibitem[Prajwal et~al.(2020)Prajwal, Mukhopadhyay, Namboodiri, and Jawahar]{wav2lip}
KR Prajwal, Rudrabha Mukhopadhyay, Vinay~P Namboodiri, and CV Jawahar.
\newblock A lip sync expert is all you need for speech to lip generation in the wild.
\newblock In \emph{Proceedings of the 28th ACM International Conference on Multimedia}, pages 484--492, 2020.

\bibitem[Ramesh et~al.(2022)Ramesh, Dhariwal, Nichol, Chu, and Chen]{dalle2}
Aditya Ramesh, Prafulla Dhariwal, Alex Nichol, Casey Chu, and Mark Chen.
\newblock Hierarchical text-conditional image generation with clip latents.
\newblock \emph{arXiv preprint arXiv:2204.06125}, 1\penalty0 (2):\penalty0 3, 2022.

\bibitem[Rezende and Mohamed(2015)]{normalizing_flow}
Danilo Rezende and Shakir Mohamed.
\newblock Variational inference with normalizing flows.
\newblock In \emph{International conference on machine learning}, pages 1530--1538. PMLR, 2015.

\bibitem[Rombach et~al.(2022)Rombach, Blattmann, Lorenz, Esser, and Ommer]{ldm}
Robin Rombach, Andreas Blattmann, Dominik Lorenz, Patrick Esser, and Bj{\"o}rn Ommer.
\newblock High-resolution image synthesis with latent diffusion models.
\newblock In \emph{Proceedings of the IEEE/CVF Conference on Computer Vision and Pattern Recognition (CVPR)}, pages 10684--10695, 2022.

\bibitem[Saharia et~al.(2022)Saharia, Chan, Saxena, Li, Whang, Denton, Ghasemipour, Gontijo~Lopes, Karagol~Ayan, Salimans, et~al.]{imagen}
Chitwan Saharia, William Chan, Saurabh Saxena, Lala Li, Jay Whang, Emily~L Denton, Kamyar Ghasemipour, Raphael Gontijo~Lopes, Burcu Karagol~Ayan, Tim Salimans, et~al.
\newblock Photorealistic text-to-image diffusion models with deep language understanding.
\newblock \emph{Advances in neural information processing systems}, 35:\penalty0 36479--36494, 2022.

\bibitem[Savchenko(2022)]{hsemotion}
Andrey~V Savchenko.
\newblock Hsemotion: High-speed emotion recognition library.
\newblock \emph{Software Impacts}, 14:\penalty0 100433, 2022.

\bibitem[Seitzer(2020)]{fid}
Maximilian Seitzer.
\newblock {pytorch-fid: FID Score for PyTorch}.
\newblock \url{https://github.com/mseitzer/pytorch-fid}, 2020.
\newblock Version 0.3.0.

\bibitem[Siarohin et~al.(2019)Siarohin, Lathuili{\`e}re, Tulyakov, Ricci, and Sebe]{fomm}
Aliaksandr Siarohin, St{\'e}phane Lathuili{\`e}re, Sergey Tulyakov, Elisa Ricci, and Nicu Sebe.
\newblock First order motion model for image animation.
\newblock \emph{Advances in neural information processing systems}, 32, 2019.

\bibitem[Simonyan and Zisserman(2014)]{vgg}
Karen Simonyan and Andrew Zisserman.
\newblock Very deep convolutional networks for large-scale image recognition.
\newblock \emph{arXiv preprint arXiv:1409.1556}, 2014.

\bibitem[Song et~al.(2020{\natexlab{a}})Song, Meng, and Ermon]{ddim}
Jiaming Song, Chenlin Meng, and Stefano Ermon.
\newblock Denoising diffusion implicit models.
\newblock \emph{arXiv preprint arXiv:2010.02502}, 2020{\natexlab{a}}.

\bibitem[Song et~al.(2020{\natexlab{b}})Song, Sohl-Dickstein, Kingma, Kumar, Ermon, and Poole]{sde}
Yang Song, Jascha Sohl-Dickstein, Diederik~P Kingma, Abhishek Kumar, Stefano Ermon, and Ben Poole.
\newblock Score-based generative modeling through stochastic differential equations.
\newblock \emph{arXiv preprint arXiv:2011.13456}, 2020{\natexlab{b}}.

\bibitem[Song et~al.(2023)Song, Dhariwal, Chen, and Sutskever]{consistency_model}
Yang Song, Prafulla Dhariwal, Mark Chen, and Ilya Sutskever.
\newblock Consistency models.
\newblock \emph{arXiv preprint arXiv:2303.01469}, 2023.

\bibitem[Stypu{\l}kowski et~al.(2024)Stypu{\l}kowski, Vougioukas, He, Zi{\k{e}}ba, Petridis, and Pantic]{diffusedhead}
Micha{\l} Stypu{\l}kowski, Konstantinos Vougioukas, Sen He, Maciej Zi{\k{e}}ba, Stavros Petridis, and Maja Pantic.
\newblock Diffused heads: Diffusion models beat gans on talking-face generation.
\newblock In \emph{Proceedings of the IEEE/CVF Winter Conference on Applications of Computer Vision}, pages 5091--5100, 2024.

\bibitem[Sun et~al.(2024)Sun, Lv, Ye, Lin, Sheng, Wen, Yu, and Liu]{diffposetalk}
Zhiyao Sun, Tian Lv, Sheng Ye, Matthieu Lin, Jenny Sheng, Yu-Hui Wen, Minjing Yu, and Yong-jin Liu.
\newblock Diffposetalk: Speech-driven stylistic 3d facial animation and head pose generation via diffusion models.
\newblock \emph{ACM Transactions on Graphics (TOG)}, 43\penalty0 (4):\penalty0 1--9, 2024.

\bibitem[Szegedy et~al.(2015)Szegedy, Liu, Jia, Sermanet, Reed, Anguelov, Erhan, Vanhoucke, and Rabinovich]{inception}
Christian Szegedy, Wei Liu, Yangqing Jia, Pierre Sermanet, Scott Reed, Dragomir Anguelov, Dumitru Erhan, Vincent Vanhoucke, and Andrew Rabinovich.
\newblock Going deeper with convolutions.
\newblock In \emph{Proceedings of the IEEE conference on computer vision and pattern recognition}, pages 1--9, 2015.

\bibitem[Tan et~al.(2023)Tan, Ji, and Pan]{emmn}
Shuai Tan, Bin Ji, and Ye Pan.
\newblock Emmn: Emotional motion memory network for audio-driven emotional talking face generation.
\newblock In \emph{Proceedings of the IEEE/CVF International Conference on Computer Vision}, pages 22146--22156, 2023.

\bibitem[Tan et~al.(2025)Tan, Ji, Bi, and Pan]{edtalk}
Shuai Tan, Bin Ji, Mengxiao Bi, and Ye Pan.
\newblock Edtalk: Efficient disentanglement for emotional talking head synthesis.
\newblock In \emph{European Conference on Computer Vision}, pages 398--416. Springer, 2025.

\bibitem[Tevet et~al.(2023)Tevet, Raab, Gordon, Shafir, Cohen-or, and Bermano]{mdm}
Guy Tevet, Sigal Raab, Brian Gordon, Yoni Shafir, Daniel Cohen-or, and Amit~Haim Bermano.
\newblock Human motion diffusion model.
\newblock In \emph{The Eleventh International Conference on Learning Representations}, 2023.

\bibitem[Tian et~al.(2024)Tian, Wang, Zhang, and Bo]{emo}
Linrui Tian, Qi Wang, Bang Zhang, and Liefeng Bo.
\newblock Emo: Emote portrait alive-generating expressive portrait videos with audio2video diffusion model under weak conditions.
\newblock \emph{arXiv preprint arXiv:2402.17485}, 2024.

\bibitem[Trevithick et~al.(2023)Trevithick, Chan, Stengel, Chan, Liu, Yu, Khamis, Chandraker, Ramamoorthi, and Nagano]{live3dportrait}
Alex Trevithick, Matthew Chan, Michael Stengel, Eric~R. Chan, Chao Liu, Zhiding Yu, Sameh Khamis, Manmohan Chandraker, Ravi Ramamoorthi, and Koki Nagano.
\newblock Real-time radiance fields for single-image portrait view synthesis.
\newblock In \emph{ACM Transactions on Graphics (SIGGRAPH)}, 2023.

\bibitem[Unterthiner et~al.(2018)Unterthiner, Van~Steenkiste, Kurach, Marinier, Michalski, and Gelly]{fvd}
Thomas Unterthiner, Sjoerd Van~Steenkiste, Karol Kurach, Raphael Marinier, Marcin Michalski, and Sylvain Gelly.
\newblock Towards accurate generative models of video: A new metric \& challenges.
\newblock \emph{arXiv preprint arXiv:1812.01717}, 2018.

\bibitem[Vaswani(2017)]{transformer}
A Vaswani.
\newblock Attention is all you need.
\newblock \emph{Advances in Neural Information Processing Systems}, 2017.

\bibitem[Wang et~al.(2024)Wang, Tian, Zhang, Guan, Luo, Shen, Jiang, Gu, Han, and Yang]{v_express}
Cong Wang, Kuan Tian, Jun Zhang, Yonghang Guan, Feng Luo, Fei Shen, Zhiwei Jiang, Qing Gu, Xiao Han, and Wei Yang.
\newblock V-express: Conditional dropout for progressive training of portrait video generation.
\newblock \emph{arXiv preprint arXiv:2406.02511}, 2024.

\bibitem[Wang et~al.(2020)Wang, Wu, Song, Yang, Wu, Qian, He, Qiao, and Loy]{mead}
Kaisiyuan Wang, Qianyi Wu, Linsen Song, Zhuoqian Yang, Wayne Wu, Chen Qian, Ran He, Yu Qiao, and Chen~Change Loy.
\newblock Mead: A large-scale audio-visual dataset for emotional talking-face generation.
\newblock In \emph{European Conference on Computer Vision}, pages 700--717. Springer, 2020.

\bibitem[Wang et~al.(2021{\natexlab{a}})Wang, Li, Ding, Fan, and Yu]{audio2head}
Suzhen Wang, Lincheng Li, Yu Ding, Changjie Fan, and Xin Yu.
\newblock Audio2head: Audio-driven one-shot talking-head generation with natural head motion.
\newblock \emph{arXiv preprint arXiv:2107.09293}, 2021{\natexlab{a}}.

\bibitem[Wang et~al.(2021{\natexlab{b}})Wang, Mallya, and Liu]{osfv}
Ting-Chun Wang, Arun Mallya, and Ming-Yu Liu.
\newblock One-shot free-view neural talking-head synthesis for video conferencing.
\newblock In \emph{Proceedings of the IEEE/CVF Conference on Computer Vision and Pattern Recognition (CVPR)}, pages 10039--10049, 2021{\natexlab{b}}.

\bibitem[Wang et~al.(2021{\natexlab{c}})Wang, Li, Zhang, and Shan]{gfpgan}
Xintao Wang, Yu Li, Honglun Zhang, and Ying Shan.
\newblock Towards real-world blind face restoration with generative facial prior.
\newblock In \emph{Proceedings of the IEEE/CVF Conference on Computer Vision and Pattern Recognition (CVPR)}, pages 9168--9178, 2021{\natexlab{c}}.

\bibitem[Wang et~al.(2022)Wang, Yang, Bremond, and Dantcheva]{lia}
Yaohui Wang, Di Yang, Francois Bremond, and Antitza Dantcheva.
\newblock Latent image animator: Learning to animate images via latent space navigation.
\newblock \emph{arXiv preprint arXiv:2203.09043}, 2022.

\bibitem[Wei et~al.(2024)Wei, Yang, and Wang]{aniportrait}
Huawei Wei, Zejun Yang, and Zhisheng Wang.
\newblock Aniportrait: Audio-driven synthesis of photorealistic portrait animation.
\newblock \emph{arXiv preprint arXiv:2403.17694}, 2024.

\bibitem[Xia et~al.(2023)Xia, Wang, Deng, Luo, and Liu]{gmtalker}
Yibo Xia, Lizhen Wang, Xiang Deng, Xiaoyan Luo, and Yebin Liu.
\newblock Gmtalker: Gaussian mixture based emotional talking video portraits.
\newblock \emph{arXiv preprint arXiv:2312.07669}, 2023.

\bibitem[Xie et~al.(2022)Xie, Wang, Zhang, Dong, and Shan]{vfhq}
Liangbin Xie, Xintao Wang, Honglun Zhang, Chao Dong, and Ying Shan.
\newblock Vfhq: A high-quality dataset and benchmark for video face super-resolution.
\newblock In \emph{Proceedings of the IEEE/CVF Conference on Computer Vision and Pattern Recognition (CVPR)}, pages 657--666, 2022.

\bibitem[Xu et~al.(2024{\natexlab{a}})Xu, Li, Su, Shang, Zhang, Liu, Wang, Van~Gool, Yao, and Zhu]{hallo}
Mingwang Xu, Hui Li, Qingkun Su, Hanlin Shang, Liwei Zhang, Ce Liu, Jingdong Wang, Luc Van~Gool, Yao Yao, and Siyu Zhu.
\newblock Hallo: Hierarchical audio-driven visual synthesis for portrait image animation.
\newblock \emph{arXiv preprint arXiv:2406.08801}, 2024{\natexlab{a}}.

\bibitem[Xu et~al.(2024{\natexlab{b}})Xu, Chen, Guo, Yang, Li, Zang, Zhang, Tong, and Guo]{vasa_1}
Sicheng Xu, Guojun Chen, Yu-Xiao Guo, Jiaolong Yang, Chong Li, Zhenyu Zang, Yizhong Zhang, Xin Tong, and Baining Guo.
\newblock Vasa-1: Lifelike audio-driven talking faces generated in real time.
\newblock \emph{arXiv preprint arXiv:2404.10667}, 2024{\natexlab{b}}.

\bibitem[Yin et~al.(2022)Yin, Zhang, Cun, Cao, Fan, Wang, Bai, Wu, Wang, and Yang]{styleheat}
Fei Yin, Yong Zhang, Xiaodong Cun, Mingdeng Cao, Yanbo Fan, Xuan Wang, Qingyan Bai, Baoyuan Wu, Jue Wang, and Yujiu Yang.
\newblock Styleheat: One-shot high-resolution editable talking face generation via pre-trained stylegan.
\newblock In \emph{European conference on computer vision}, pages 85--101. Springer, 2022.

\bibitem[Yu et~al.(2018)Yu, Wang, Peng, Gao, Yu, and Sang]{face_segmentation}
Changqian Yu, Jingbo Wang, Chao Peng, Changxin Gao, Gang Yu, and Nong Sang.
\newblock Bisenet: Bilateral segmentation network for real-time semantic segmentation.
\newblock In \emph{Proceedings of the European conference on computer vision (ECCV)}, pages 325--341, 2018.

\bibitem[Yu et~al.(2023)Yu, Zhu, Jiang, Loy, Cai, and Wu]{celebv_text}
Jianhui Yu, Hao Zhu, Liming Jiang, Chen~Change Loy, Weidong Cai, and Wayne Wu.
\newblock Celebv-text: A large-scale facial text-video dataset.
\newblock In \emph{Proceedings of the IEEE/CVF Conference on Computer Vision and Pattern Recognition}, pages 14805--14814, 2023.

\bibitem[Zhang et~al.(2023{\natexlab{a}})Zhang, Rao, and Agrawala]{controlnet}
Lvmin Zhang, Anyi Rao, and Maneesh Agrawala.
\newblock Adding conditional control to text-to-image diffusion models.
\newblock In \emph{Proceedings of the IEEE/CVF International Conference on Computer Vision}, pages 3836--3847, 2023{\natexlab{a}}.

\bibitem[Zhang et~al.(2018)Zhang, Isola, Efros, Shechtman, and Wang]{lpips}
Richard Zhang, Phillip Isola, Alexei~A Efros, Eli Shechtman, and Oliver Wang.
\newblock The unreasonable effectiveness of deep features as a perceptual metric.
\newblock In \emph{Proceedings of the IEEE/CVF Conference on Computer Vision and Pattern Recognition (CVPR)}, pages 586--595, 2018.

\bibitem[Zhang et~al.(2023{\natexlab{b}})Zhang, Cun, Wang, Zhang, Shen, Guo, Shan, and Wang]{sadtalker}
Wenxuan Zhang, Xiaodong Cun, Xuan Wang, Yong Zhang, Xi Shen, Yu Guo, Ying Shan, and Fei Wang.
\newblock Sadtalker: Learning realistic 3d motion coefficients for stylized audio-driven single image talking face animation.
\newblock In \emph{Proceedings of the IEEE/CVF Conference on Computer Vision and Pattern Recognition (CVPR)}, pages 8652--8661, 2023{\natexlab{b}}.

\bibitem[Zhang et~al.(2021)Zhang, Li, Ding, and Fan]{hdtf}
Zhimeng Zhang, Lincheng Li, Yu Ding, and Changjie Fan.
\newblock Flow-guided one-shot talking face generation with a high-resolution audio-visual dataset.
\newblock In \emph{Proceedings of the IEEE/CVF Conference on Computer Vision and Pattern Recognition (CVPR)}, pages 3661--3670, 2021.

\bibitem[Zhang et~al.(2023{\natexlab{c}})Zhang, Hu, Deng, Fan, Lv, and Ding]{dinet}
Zhimeng Zhang, Zhipeng Hu, Wenjin Deng, Changjie Fan, Tangjie Lv, and Yu Ding.
\newblock Dinet: Deformation inpainting network for realistic face visually dubbing on high resolution video.
\newblock In \emph{Proceedings of the AAAI Conference on Artificial Intelligence}, pages 3543--3551, 2023{\natexlab{c}}.

\bibitem[Zhou et~al.(2021)Zhou, Sun, Wu, Loy, Wang, and Liu]{pcavs}
Hang Zhou, Yasheng Sun, Wayne Wu, Chen~Change Loy, Xiaogang Wang, and Ziwei Liu.
\newblock Pose-controllable talking face generation by implicitly modularized audio-visual representation.
\newblock In \emph{Proceedings of the IEEE/CVF Conference on Computer Vision and Pattern Recognition (CVPR)}, pages 4176--4186, 2021.

\bibitem[Zhou et~al.(2020)Zhou, Han, Shechtman, Echevarria, Kalogerakis, and Li]{makeittalk}
Yang Zhou, Xintong Han, Eli Shechtman, Jose Echevarria, Evangelos Kalogerakis, and Dingzeyu Li.
\newblock Makelttalk: speaker-aware talking-head animation.
\newblock \emph{ACM Transactions On Graphics (TOG)}, 39\penalty0 (6):\penalty0 1--15, 2020.

\bibitem[Zhu et~al.(2024)Zhu, Chen, Dai, Xu, Cao, Yao, Zhu, and Zhu]{champ}
Shenhao Zhu, Junming~Leo Chen, Zuozhuo Dai, Yinghui Xu, Xun Cao, Yao Yao, Hao Zhu, and Siyu Zhu.
\newblock Champ: Controllable and consistent human image animation with 3d parametric guidance.
\newblock \emph{arXiv preprint arXiv:2403.14781}, 2024.

\end{thebibliography}
